\documentclass[11pt]{article}
\usepackage[final]{acl}
\usepackage{times}
\usepackage{latexsym}

\usepackage[T1]{fontenc}

\usepackage[utf8]{inputenc}

\usepackage{microtype}

\usepackage{inconsolata}

\usepackage{graphicx}
\usepackage{subcaption}
\usepackage{booktabs}

\usepackage[accsupp]{axessibility}  

\usepackage{rotating}
\usepackage{graphicx} 
\newcommand{\method}[1]{\textsc{#1}}

\usepackage{booktabs,threeparttable,multirow,makecell,array,pifont}
\usepackage{amsmath}
\usepackage{graphicx}
\usepackage{multirow}
\usepackage[table,xcdraw]{xcolor}
\usepackage{lscape}
\usepackage{booktabs,colortbl,adjustbox}
\usepackage{siunitx}
\usepackage{array,xcolor}
\newcolumntype{d}{S[table-format=2.4]}
\newcolumntype{D}{>{\columncolor{pink!20}}S[table-format=2.4]}

\renewcommand{\cline}[1]{\cmidrule{#1}}
\let\cline\cmidrule

\newcolumntype{P}[1]{>{\raggedright\arraybackslash}p{#1}}
\newcolumntype{C}[1]{>{\centering\arraybackslash}p{#1}}
\usepackage{orcidlink}

\title{MVC-Bench: Benchmarking Calibration of Medical Vision-Language Models}

\author{
\textbf{Ashshak Sharifdeen\textsuperscript{1}},
\textbf{Shihab Aaqil Ahamed\textsuperscript{1}},
\textbf{Ufaq Khan\textsuperscript{1}},
\textbf{Muhammad Akhtar Munir\textsuperscript{1}} \\
\textbf{Sujair Ibrahim\textsuperscript{2}},
\textbf{Mohamed Rafeek Mareer Ahamed\textsuperscript{3}},
\textbf{Yutong Xie\textsuperscript{1}}, \\
\textbf{Imran Razzak\textsuperscript{1}},
\textbf{Muhammad Haris Khan\textsuperscript{1}} \\
\\
\textsuperscript{1}Mohamed bin Zayed University of AI \\
\textsuperscript{2}Sabaragamuwa University of Sri Lanka \\
\textsuperscript{3}Digital Platform Development, SLT PLC  \\
\texttt{\{ashshak.sharifdeen, muhammad.haris\}@mbzuai.ac.ae}
}

\makeatletter
\renewcommand\@biblabel[1]{#1}   
\makeatother
\begin{document}
\maketitle
\begin{abstract}
Reliable evaluation of vision-language models (VLMs) and medical vision-language models (Medical-VLMs) requires calibrated confidence, particularly under realistic clinical conditions. However, existing efforts mainly focused on improving accuracy, leaving calibration in the medical domain underexplored. To this end, we propose MVC-Bench, a calibration-centric benchmark for medical image classification with VLMs and Medical-VLMs. MVC-Bench assesses the calibration across three axes: (i) robustness to modality, backbone, and domain shift (ii) effectiveness of calibration strategies and prompt-tuning methods (iii) stability under prompt-template and random-seed variations. The benchmark covers eight different backbones, three medical modalities, including fundus imaging, histopathology, and chest X-ray under in-domain and domain shift settings. It compares post-hoc calibration, train-time calibration, and zero-shot inference methods, together with six prompt-tuning methods. 
Across more than 1638  controlled experiments, we report accuracy and Expected Calibration Error (ECE) as primary metrics, and further report results with complementary calibration measures, including Maximum Calibration Error (MCE) and Adaptive Calibration Error (ACE).  We further investigate the underlying causes of miscalibration in VLMs and Medical-VLMs and propose a simple train-time calibration method, Multi-Class Margin (MCM) regularization, which achieves lowest ECE on 10 out of 12 settings in in-domain and remains competitive under domain shifts. Collectively, MVC-Bench provides a structured evaluation framework and actionable guidance for improving calibration in safety-critical medical workflows. Our Code:\url{https://github.com/ashshaksharifdeen/MVC-Bench}.
\end{abstract}

\section{Introduction}
\label{sec:intro}

\begin{figure}[t!]
\centering
\includegraphics[width=0.8\linewidth]{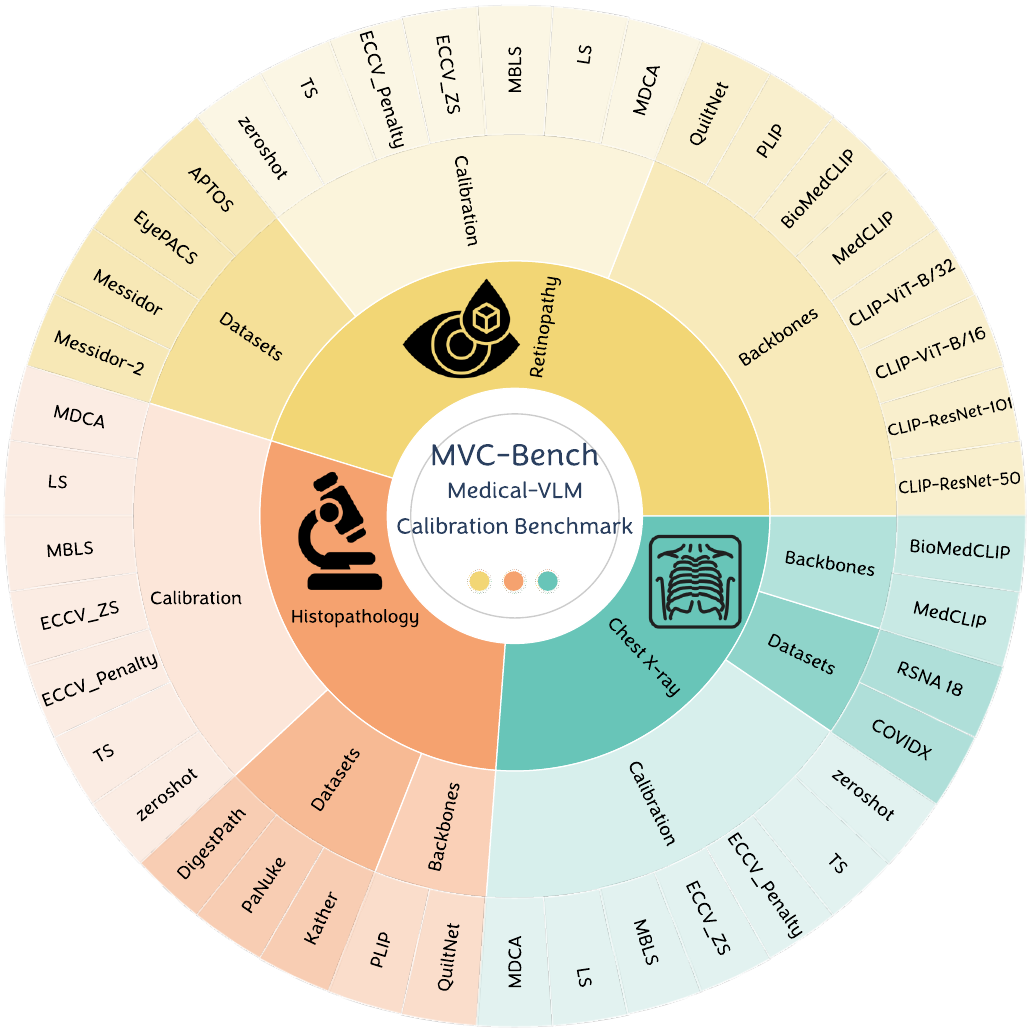}
\caption{Overview of MVC-Bench, which evaluates calibration of VLMs and Medical-VLMs across three axes: robustness to modality, backbone, and domain shift; effectiveness of calibration and prompt-tuning strategies; and stability under prompt-template and random-seed variations. The benchmark spans DR, histopathology, and chest X-ray datasets, covering relevant backbones and calibration settings across more than 1638  controlled experiments.}
\label{fig:teaser}
\vspace{-15pt}
\end{figure}

VLMs leverage large-scale language supervision to align visual and textual representations, enabling strong zero/few-shot transfer with limited task-specific labels~\cite{radford2021clip,zhai2022lit,li2022blip,li2023blip2}. Medical-VLMs extend this capability to clinical imaging, where annotations are scarce and costly. This capability has been explored across multiple medical modalities, including fundus imaging~\cite{gulshan2016development}, histopathology~\cite{campanella2019clinical}, and chest X-rays~\cite{rajaraman2020analyzing}
. By aligning medical images with disease-aware prompts, medical-VLMs support disease-pattern recognition and transferable evaluation across heterogeneous medical datasets~\cite{boecking2022making,wang2022medclip,zhang2023biomedclip,huang2023plip,ikezogwo2023quilt1m}.
Since full fine-tuning large VLMs is often costly, prompt tuning has become a practical parameter-efficient alternative, updating only a small set of prompt parameters while keeping the backbone frozen~\cite{liu2023pre}. This low-cost adaptation is particularly valuable in medical imaging, where full fine-tuning is often infeasible~\cite{hanif2024baple,hussein2024promptsmooth}.

Despite recent advances, prompt-tuning methods for VLMs mainly optimize accuracy while often overlooking confidence calibration~\cite{sharifdeen2026calibratingprompttuningvisionlanguage}. However, clinical deployment requires confidence scores that reflect the true likelihood of correctness, where a classifier is \textit{well-calibrated} if predictions made with confidence $p$ are correct approximately $p\%$ of the time~\cite{guo2017calibration}.
Miscalibration can cause \textit{overconfidence}, where incorrect predictions receive high confidence and may miss human review, or \textit{underconfidence}, where correct predictions receive unreliable confidence, undermining triage and trust~\cite{minderer2021revisiting,electronics13030503,zhu2023rethinkingconfidencecalibrationfailure,ao2023sidesmiscalibrationidentifyingunderconfidence}. 
In medical imaging, this problem is further amplified by class imbalance, heterogeneous acquisition, prevalence drift, and device or patient shifts~\cite{koh2021wilds}. Therefore, calibration of VLMs and Medical-VLMs under realistic factors, including domain shift, prompt tuning, hard-prompt choice, and seed variability, remains insufficiently explored.

In this paper, we address this crucial gap with \textbf{MVC-Bench}, a comprehensive calibration-centric benchmark for VLM-based medical image classification. \textbf{MVC-Bench} organizes calibration evaluation along three dimensions: \textit{robustness} across backbone, modality, and domain shift; \textit{effectiveness} of calibration strategies and prompt-tuning methods; and \textit{stability} under prompt-template and random-seed variations. This factorized design enables systematic analysis under clinically relevant conditions. To our knowledge, \textbf{MVC-Bench} is the first calibration-centric benchmark focused on prompt-tuned VLM and Medical-VLM image classification across these three medical modalities. We benchmark classical and VLM‑aware calibration techniques across eight popular backbones CLIP-ResNet/ViT~\cite{radford2021clip}, PLIP~\cite{huang2023visual}, QuiltNet~\cite{ikezogwo2023quilt1m}, MedCLIP~\cite{wang2022medclip}, and BioMedCLIP~\cite{zhang2023biomedclip} and systematically compare prompt‑tuning approaches by adopting CoOp as a widely used baseline and contrasting with more advanced prompt learners KgCoOp, MaPLe, PromptSRC, HiCroPL, and ProGrad~\cite{zhou2022coop,khattak2023maple,khattak2023promptsrc,yao2023visual,Zheng_2025_ICCV, zhu2023prompt}. Our datasets span three modalities: (i) \textit{fundus} (Aptos~\cite{aptos2019}, EyePACS~\cite{eyepacs2015}, Messidor~\cite{decenciere2014messidor}, Messidor-2~\cite{decenciere2014feedback}), (ii) \textit{histopathology} (Kather~\cite{kather2019predicting}, PaNuke~\cite{gamper2019pannuke}, DigestPath~\cite{da2022digestpath}), and (iii) \textit{X-ray} (COVIDX~\cite{rahman2021exploring}, RSNA 18~\cite{rsna2018pneumonia}). 

Fig.~\ref{fig:teaser} illustrates the coverage of our MVC-Bench. In total, we conduct over 1638  experiments organized around three dimensions: \textit{robustness}, \textit{effectiveness}, and \textit{stability}. These dimensions are guided by seven research questions: Under \textit{robustness}, we examine How domain shifts affect calibration, and Which backbones are better calibrated across medical modalities. Under \textit{effectiveness}, we analyze the accuracy-calibration trade-offs, which calibration methods are consistently reliable across backbones, and how prompt-tuning families affect calibration. Under \textit{stability}, we investigate How do different prompt initialization interact with Calibration, and How different seed initializations affect calibration performance?. We report accuracy and expected calibration error (ECE)~\cite{naeini2015bbq}, with additional metrics in the supplementary. 

Guided by these findings, we analyze miscalibration sources and observe that underconfidence often stems from weak inter-class separation, where the correct-class logit only slightly exceeds competing logits. We propose \textbf{Multi-Class Margin (MCM)} Regularization, a train-time regularizer that increases true-vs-rest margins while controlling margin variance to reduce underconfidence and spurious confidence spikes. MCM achieves the lowest ECE for most backbones and remains competitive under domain shifts. Overall, our contributions are a calibration-centric benchmark for VLMs and Medical-VLMs across robustness, effectiveness, and stability axes, a large-scale evaluation across modalities, backbones, calibration methods, and prompt-tuning strategies, and a simple Multi-Class Margin (MCM) regularizer that improves in-domain ECE while remaining competitive under domain shift.
\section{Related Work}
\label{sec:related_work}
\noindent\textbf{Medical VLMs and Prompt Tuning.}
Medical imaging has increasingly adopted VLMs to reduce annotation demands and improve transfer across institutions and modalities. In radiology, MedCLIP, BioViL/BioViL-T, BioMedCLIP, and CheXzero align medical images with free-text reports for zero-shot labeling and retrieval~\cite{wang2022medclip,boecking2022making,zhang2023biomedclip,tiu2022chexzero}, while pathology-focused models such as PLIP and Quilt/QuiltNet use language-supervised pretraining over slide tiles and pathology-specific corpora to improve recognition under data heterogeneity~\cite{huang2023visual,ikezogwo2023quilt}. 

Prompt-tuning methods, beginning with CoOp~\cite{zhou2022learning} and later extended by KgCoOp, MaPLe, HiCroPL, PromptSRC, and ProGrad~\cite{yao2023visual,khattak2023maple,Zheng_2025_ICCV,khattak2023promptsrc,zhu2023prompt}, adapt frozen VLMs through learnable prompts. However, although Medical-VLMs and prompt tuning often improve transfer or accuracy, their confidence calibration across modalities, institutions, prompts, and deployment shifts remains insufficiently understood, and prompt adaptation can even degrade calibration by shifting decision boundaries~\cite{sharifdeen2026calibratingprompttuningvisionlanguage}.
\\
\noindent\textbf{Calibration in Deep Neural Network and Medical Models.}
Calibration methods are commonly divided into post-hoc approaches, which adjust confidence without changing decision boundaries, and train-time approaches, which add calibration-aware objectives or regularizers. Representative post-hoc methods include Platt scaling, isotonic regression, binning-based calibration, temperature scaling, and multiclass calibration~\cite{platt1999probabilistic,zadrozny2002transforming,naeini2016binary,guo2017calibration,kull2019beyond}, while train-time methods include label smoothing, mixup, focal loss, class-balanced objectives, trainable calibration losses, and deep ensembles~\cite{muller2019does,thulasidasan2019mixup,lin2017focal,cui2019class,kumar2018trainable,lakshminarayanan2017simple}. 

In medical imaging, calibration is especially important for triage, escalation, and human-AI collaboration~\cite{gawlikowski2023survey}, but it is complicated by class imbalance, label noise, acquisition variation, and site shift~\cite{karimi2020improving,laves2020well,buda2018systematic,ovadia2019uncertainty,guo2017calibration,kull2017beta,naeini2015bbq,lakshminarayanan2017simple}. Most prior studies still focus on supervised CNN/ViT models with fixed label spaces~\cite{minderer2021revisiting}, leaving calibration in promptable, open-vocabulary Medical-VLMs under prompt tuning and domain shift relatively underexplored, despite the strong influence of prompts, class granularity, and template design on confidence reliability~\cite{radford2021clip,zhou2022coop,zhou2022cocoop,khattak2023maple,tiu2022chexzero,boecking2022making,zhang2023biomedclip}.
\\
\vspace{-1.5em}
\section{MVC-Bench}
\label{sec:benchmark}
Our \textbf{MVC-Bench} is formulated as a calibration stress-test protocol for VLMs and Medical-VLMs, rather than a single in-domain leaderboard. It organizes seven research questions into three axes: \textit{robustness} to modality, backbone choice, and domain shift; \textit{effectiveness} of calibration and prompt-tuning strategies; and \textit{stability} under prompt-template and random-seed variation. 

\noindent\textbf{Dataset characteristics}  
MVC-Bench covers fundus imaging (APTOS, EyePACS, Messidor, Messidor-2), histopathology (Kather, PaNuke, DigestPath), and CXR (COVIDX, RSNA18)~\cite{aptos2019,eyepacs2015,decenciere2014feedback,kather2019predicting,gamper2019pannuke,da2022digestpath,rahman2021exploring,Stein2018}. These datasets provide heterogeneous visual statistics, label spaces, and class priors for evaluating calibration under in-domain and shifted settings.

\noindent\textbf{Implementation details.} We report top-1 accuracy (Acc.) and 20-bin Expected Calibration Error (ECE) as the primary evaluation metrics, and provide complementary calibration analyses using Maximum Calibration Error (MCE) and Adaptive Calibration Error (ACE) in the appendix. Prompt tuning uses CoOp in a 16-shot setting with learning rate 0.005 and batch size 8, following official baseline implementations. All experiments are run over three seeds on an NVIDIA RTX A6000 GPU, with full hyperparameters in the \textit{supplementary material}.

\noindent\textbf{Vision-Language Models (VLMs).} We evaluate both generic VLMs and Medical-VLMs to study calibration across model families and modalities. The generic backbones include CLIP-ViT-B/16, CLIP-ViT-B/32, CLIP-ResNet-50, and CLIP-ResNet-101, while the medical backbones include PLIP, QuiltNet, MedCLIP, and BioMedCLIP. We further analyze parameter-efficient prompt-tuning methods, including CoOp, KgCoOp, MaPLe, PromptSRC, ProGrad, and HiCroPL, to assess their impact on calibrated confidence.

\noindent\textbf{Evaluated calibration baselines.} 
For our study, we selected prominent and well-referenced calibration methods covering both train-time and post-hoc approaches: MDCA~\cite{hebbalaguppe2022stitch}, LS~\cite{muller2019does}, MBLS~\cite{liu2022mbls}, ECCV\_ZS~\cite{murugesan2024robust}, ECCV\_penalty~\cite{murugesan2024robust}, and Temperature Scaling~\cite{guo2017calibration}. In addition, we report zero-shot inference for each frozen backbone as an uncalibrated reference baseline, without prompt tuning and post-hoc calibration.  
\section{Experiments and Discussion}
\subsection{Robustness: How do domain shifts affect calibration?}
\label{subsec:domainshift}
\begin{figure*}[h!]
\centering
\includegraphics[width=\textwidth,clip]{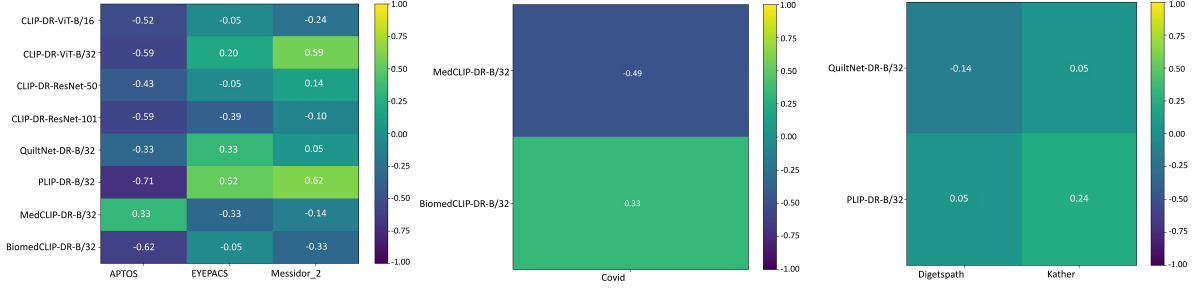} 
\vspace{-15pt}
\caption{Kendall rank correlation ($\tau$) between ECE-based method rankings on the In-domain(ID) source and each Domain shift(DS) dataset is shown for every backbone (rows = backbones, columns = DS sets). Cells are colored by $\tau$ (-$1$ to $1$) and annotated with values: higher/positive $\tau$ means rankings (and thus relative calibration quality) are preserved under shift, while near-zero/negative $\tau$ indicates rank reversals of the calibration under shift.}
\label{fig:dg}
\vspace{-10pt}
\end{figure*}
Domain shift is a central robustness challenge for clinical deployment of VLMs and Medical-VLMs, as changes in acquisition devices, patient populations, and clinical protocols can alter confidence reliability~\cite{koh2021wilds,Ghafoorian_2017}. These shifts manifest differently across modalities: fundus images vary across cameras and clinics~\cite{Galappaththige_2024_WACV}, histopathology is affected by staining variation~\cite{Tellez2019QuantifyingTE}, and chest X-rays reflect differences in disease prevalence and demographics~\cite{Zech2018VariableGP}. Since such shifts may preserve accuracy while degrading calibration, we examine whether ECE-based calibration performance learned in in-domain settings remains stable under domain-shifted evaluation across modalities.

\noindent\textbf{Experimental setup:} We evaluate whether in-domain (ID) calibration predicts domain shift (DS) performance using ECE across eight backbones and seven calibration methods. The analysis covers three domains: retinopathy (ID: Messidor, excluding class 4; DS: APTOS, EyePACS, Messidor-2), chest X-ray (ID: RSNA18; DS: COVIDX), and histopathology (ID: PaNuke; DS: DigestPath, Kather). For each backbone, we rank calibration methods by ID ECE and compare them with DS rankings using Kendall’s $\tau_b$ correlation. 
\\
\noindent\textbf{Discussion:} 
In Fig.~\ref{fig:dg}, most retinopathy backbones show negative or
near-zero $\tau_b$, indicating weak calibration-rank transfer under
external-domain evaluation. Although some backbones preserve rankings
for particular datasets, this behavior is not consistent across
shifts. Across all 30 ID-to-DS rank-transfer tests, the mean and median
Kendall correlations are $-0.0598$ and $-0.0476$, respectively
17/30 correlations are negative, 13/30 are positive, and none shows
significantly positive rank preservation at $p<0.05$
(Appendix~\ref{sec:ds_statistical_analysis}). Complementary ACE and MCE analyses
(Appendix~\ref{sec:supp_ds_metrics}) show that rankings can remain
stable for some closer shifts, but this stability is not universal
across modalities, backbones, or metrics. Overall, \textbf{calibration-method
rank transfer under external-domain shift is weak and metric-dependent.}
\subsection{Robustness: Which VLM-based backbones are better calibrated?}
\label{subsec:backbone}
Backbone selection is another key factor towards gauging  calibration robustness because generic VLMs and Medical-VLMs encode different visual and semantic priors. We therefore compare generic and domain-specific backbones across DR, histopathology, and X-ray tasks to identify which model families provide more reliable confidence beyond accuracy.

\begin{figure}[h!]
\centering
\includegraphics[width=1.0\linewidth]{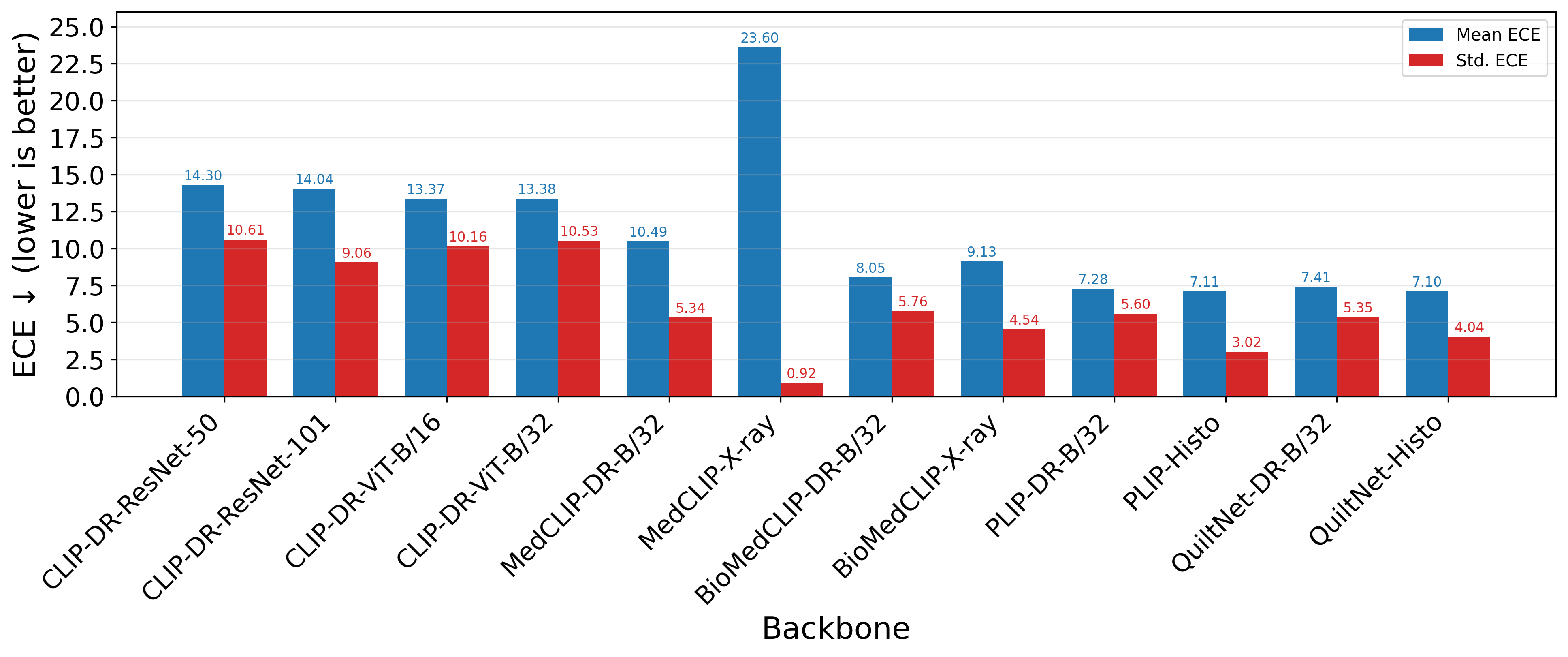}
\vspace{-10pt}
\caption{Comparison of mean and Std. of ECE for various VLM-based backbones across different medical modality inputs.}
\label{fig:bar_chart}
\vspace{-15pt}
\end{figure}

\noindent\textbf{Experimental setup:} We assess calibration across VLM and Medical-VLM backbones on DR grading, histopathology, and X-ray tasks using seven calibration methods, reporting the mean and standard deviation of ECE.
\\
\noindent\textbf{Discussion:} 
Fig.~\ref{fig:bar_chart} shows clear calibration disparities across VLM and Medical-VLM backbones. Domain-specific models generally achieve lower mean ECE than generic CLIP variants, with BioMedCLIP-X-ray outperforming MedCLIP-X-ray and
PLIP/QuiltNet variants showing strong calibration in their aligned
modalities. However, MedCLIP-X-ray is an important counterexample,
showing that medical-domain pretraining alone does not guarantee
calibrated confidence. Although MedCLIP and BioMedCLIP are both
medically pretrained, they differ substantially in pretraining scale
and data coverage. BioMedCLIP is trained on the much broader PMC-15M
biomedical image-text corpus\cite{zhang2023biomedclip}. Thus, calibration appears to depend not only on whether pretraining is domain-specific, but also on its scale, coverage, label-space alignment, and resulting confidence geometry.
ACE and MCE analyses lead to the same broader conclusion and likewise
identify MedCLIP-X-ray as a particularly poorly calibrated backbone
(Appendix~\ref{sec:supp_backbone_metrics}). Overall,
\textbf{medical/domain-aligned pretraining generally improves
calibration, but it is not sufficient by itself to guarantee reliable
confidence.} 
\subsection{Effectiveness: How does calibration vary with accuracy for different calibration methods?}
\vspace{-1pt}
Calibration effectiveness cannot be inferred from accuracy alone, because medical models may remain confidently wrong under class imbalance, heterogeneous acquisition, or prevalence shift. We therefore analyze the accuracy-calibration relationship across calibration methods and backbones using Pearson correlation between Acc. and ECE.

\begin{figure}[h!]
 \vspace{-4pt}
\centering
\includegraphics[width=1.0\linewidth]{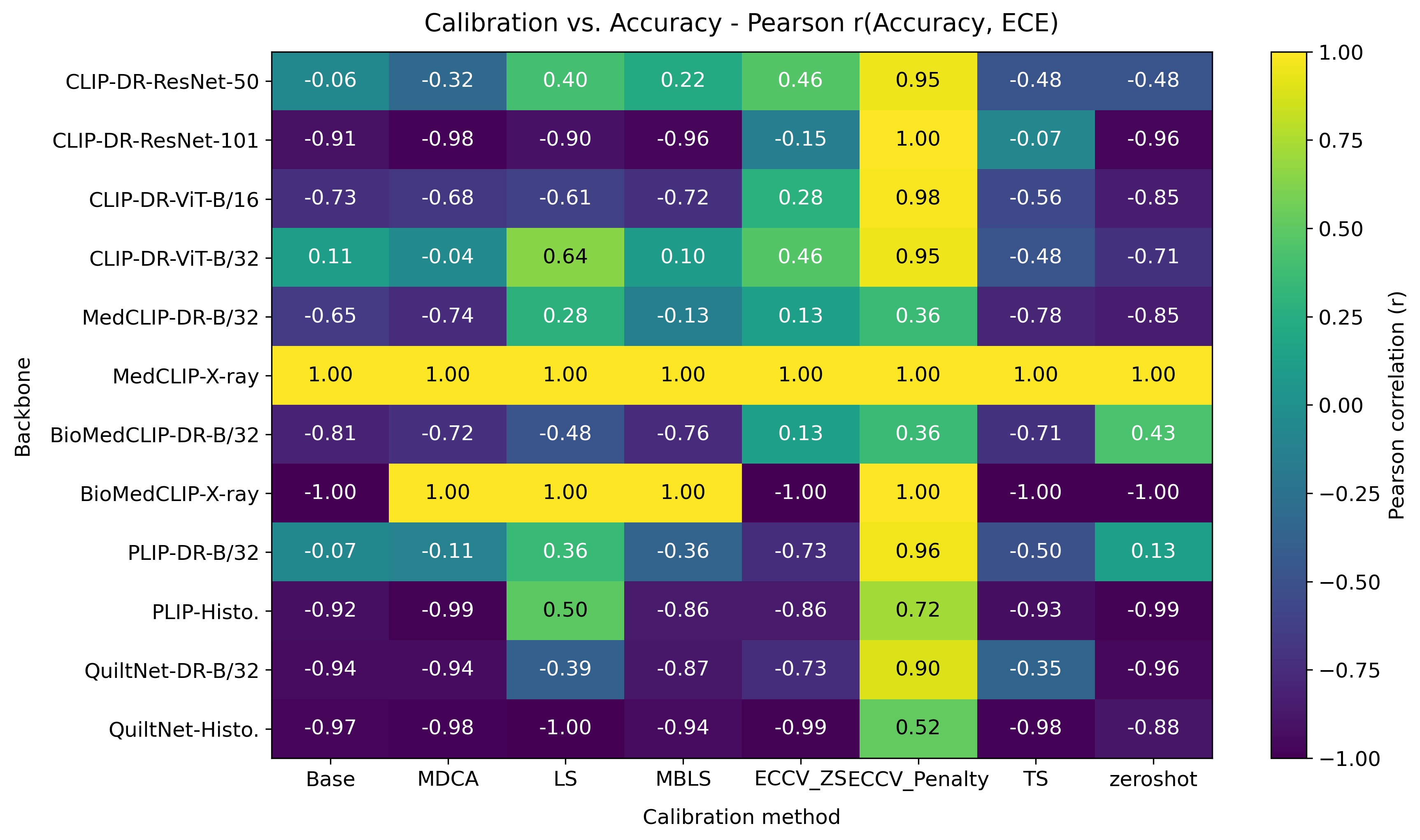}
\vspace{-10pt}
\caption{Pearson correlation matrix between accuracy and ECE evaluated across diverse calibration methods and backbones.}
\label{fig:pearson}
\vspace{-4pt}
\end{figure}

\noindent\textbf{Experimental setup:} We compute Pearson correlation $r(\text{Acc}, \text{ECE})$ across datasets for each calibration method and backbone to examine the accuracy--calibration relationship. Negative values indicate that higher accuracy corresponds to lower ECE, while positive values indicate that higher accuracy is associated with higher ECE.
\\
\noindent\textbf{Discussion:} 
Fig.~\ref{fig:pearson} shows that the accuracy-ECE relationship is highly method and backbone-dependent. TS, MDCA, and MBLS often show negative correlations, indicating that accuracy and calibration can improve together, whereas ECCV\_Penalty consistently shows positive correlations, suggesting an accuracy-calibration trade-off. CLIP backbones exhibit mixed behaviour, while X-ray backbones show extreme $\pm1$ values due to limited dataset pairs. Overall, \textbf{there is no universal accuracy-calibration trade-off, calibration effectiveness depends strongly on both the method and backbone.} Complementary ACE and MCE analyses show the same overall behavior:
the relationship between predictive accuracy and calibration remains
method and backbone dependent, confirming that higher accuracy does
not generally imply better calibration
(Appendix~\ref{sec:supp_accuracy_metrics}).
\subsection{Effectiveness: Which calibration methods show consistent calibration performance across various backbones?}
\label{subsec:calibmethod}
 A practical calibration method should remain reliable across backbones, modalities, and deployment shifts without requiring extensive re-tuning. We therefore evaluate whether each calibration strategy maintains low and consistent ECE across VLM and Medical-VLM backbones in both ID and DS settings.
\begin{figure*}[t!]
\centering
\includegraphics[width=\linewidth]{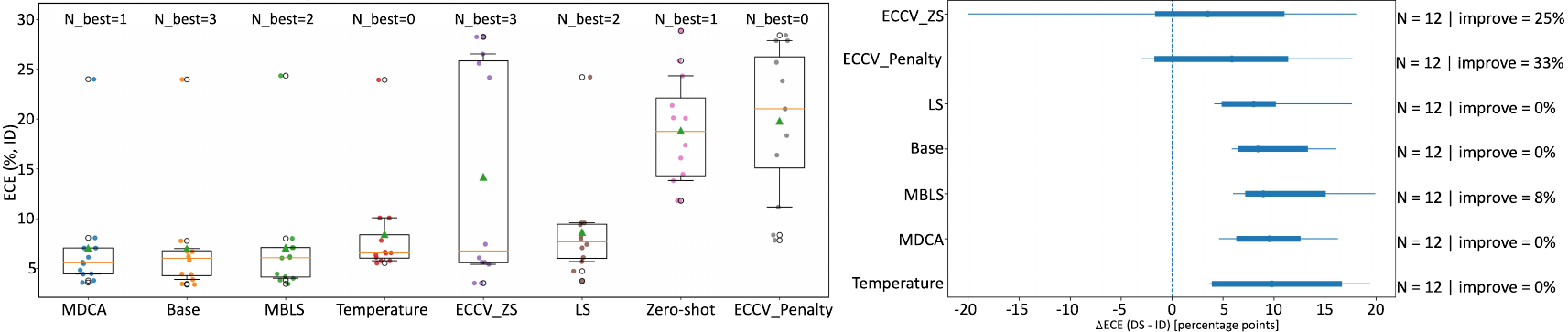}

\caption{Left (ID): Boxplots of ECE (\%) per method across backbones (IQR boxes, 10--90\% whiskers, $\blacktriangle$ mean, colored dots = backbone points, hollow circles = outliers). N$_{\text{best}}$ shows the number of times it performs best across backbones. Lower, tighter boxes = more consistent ID calibration.
Right (DS): Forest plot summarizing how each method’s ECE changes from ID to DS across backbones (dot = median, thick bar = IQR, thin line = 10--90\%, dashed line at zero change), right text reports N backbones and improve \% (share of backbones that get lower ECE in DS).}
\label{fig:cal-method}
\vspace{-5pt}
\end{figure*}

\noindent\textbf{Experimental setup:} In Fig.~\ref{fig:cal-method}, we evaluate seven calibration methods across 12 backbones and three medical modalities. The ID panel reports ECE for each backbone-method pair, while the DS panel reports the median backbone-level $\Delta\text{ECE}=\text{ECE}{\text{DS}}-\text{ECE}{\text{ID}}$ across the corresponding shifted datasets to assess robustness. 

\noindent\textbf{Discussion:} In ID settings, MBLS, Base, and MDCA show the most consistent calibration with low ECE and smaller spread, while LS and TS perform moderately, and ECCV\_ZS, ECCV\_Penalty, and Zero-shot show higher ECE variability. $N_{\text{best}}$ further favors Base as the most frequent lowest-ECE method. Under DS, $\Delta\text{ECE}$ is mostly positive, indicating degraded calibration. Only ECCV\_ZS and ECCV\_Penalty improve in a subset of cases, likely due to their use of pretrained knowledge for shift-aware calibration. Overall, \textbf{no method is consistently reliable across both ID and DS settings, highlighting the need for calibration methods specifically designed for robust medical deployment under domain shift.} This conclusion is consistent under ACE and MCE: calibration method rankings and dispersion remain backbone  and shift-dependent, with no existing baseline providing uniformly reliable calibration across ID and DS settings (Appendix~\ref{sec:supp_method_metrics}).
\subsection{Effectiveness: How do different prompt-tuning methods interact with Calibration?}
Recent work has studied how prompt tuning effectively affects calibration and global logit distributions in natural image benchmarks~\cite{liu2023pre}. However, in medical imaging, this interaction is more complex due to device heterogeneity, prevalence shifts, ambiguous class boundaries, context length, text-image alignment, and backbone pretraining. These factors can significantly alter confidence reliability even when accuracy remains high.

\noindent\textbf{Experimental setup:} We compare six prompt-tuning families CoOp, KgCoOp, MaPLe, PromptSRC, HiCroPL, and ProGrad on DR classification across APTOS, EyePACS, Messidor, and Messidor-2. Using the same backbone, data splits, and no additional calibration method, only the prompt-tuning strategy is varied. The figure reports mean Accuracy and ECE across datasets with standard deviation error bars, where higher Accuracy and lower ECE indicate better performance.  

\begin{figure}[h!]
\vspace{-8pt}
\centering
\includegraphics[width=1.0\linewidth]{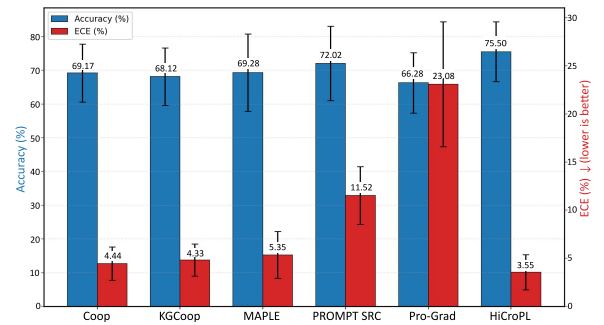}
\vspace{-10pt}
\caption{Comparison of Acc and ECE for different prompt-tuning methods.} 
\label{fig:prompt-tuning}
\vspace{-10pt}
\end{figure}

\noindent\textbf{Discussion:} Fig.~\ref{fig:prompt-tuning} shows a clear accuracy--calibration trade-off. PromptSRC achieves the highest accuracy ($\sim$72\%) but has poor calibration (ECE $\approx$11.5\%) and high variability, while ProGrad performs worst in both accuracy ($\sim$66\%) and ECE ($\sim$23\%). In contrast, CoOp, HiCroPL, and KgCoOp provide the best balance, maintaining competitive accuracy ($\sim$68--75\%) with low ECE ($\sim$3.5--4.8\%) and smaller variance. MaPLe shows intermediate performance. Overall, \textbf{high accuracy alone does not guarantee reliable confidence, highlighting the need for prompt-tuning methods that explicitly preserve calibration.} ACE and MCE exhibit the same broad prompt-family trend, with HiCroPL,
CoOp, and KgCoOp providing comparatively stronger calibration than
several higher-error prompt learners (Appendix~\ref{sec:supp_prompt_metrics}). Thus, the observed accuracy calibration mismatch is not specific to ECE.

\subsection{Stability: How does hard-prompt initialization affect calibration?} Prompt-template selection is an important stability factor because Generic prompts may improve calibration in natural-image settings~\cite{yoon2024c,sharifdeen2025tpt,ahamed2026tpt}, but medical tasks require careful phrasing due to fine-grained ontologies and variable terminology. Since subtle wording changes can alter confidence without changing accuracy, we benchmark natural-image-inspired, noisy, and clinically tailored prompts to study their effect on medical VLM calibration. 

\noindent\textbf{Experimental setup:} We study the effect of hard-prompt initialization on calibration using MaPLe-ViT-B/16~\cite{khattak2023maple} for diabetic retinopathy classification. We evaluate Accuracy and ECE under seven hard-prompt templates (P1-P7) with different phrasing, context, and domain specificity. Prompt-template details are provided in the supplementary material.

\begin{figure*}[t]
\centering

\begin{subfigure}[t]{0.44\linewidth}
  \centering
  \includegraphics[width=\linewidth]{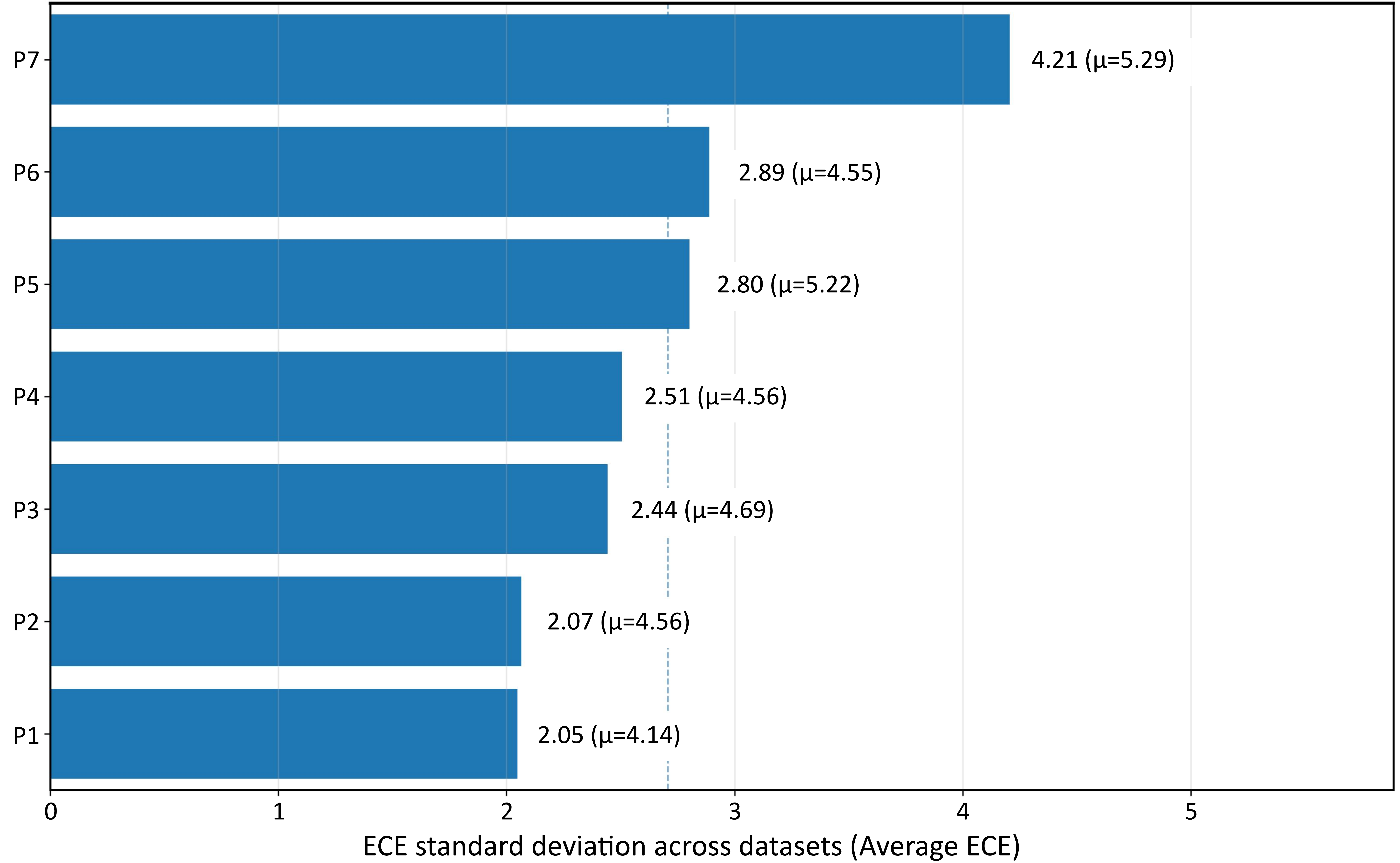}
  \vspace{-6pt}
  \caption{Comparison of Std. and mean of ECE ($\mu$) across datasets for seven different hard-prompt templates (P1-P7). Templates P1-P4 have low and stable ECE across datasets, while P7 has high variance.}
  \label{fig:prompt}
\end{subfigure}
\hfill
\begin{subfigure}[t]{0.48\linewidth}
  \centering
  \includegraphics[width=\linewidth]{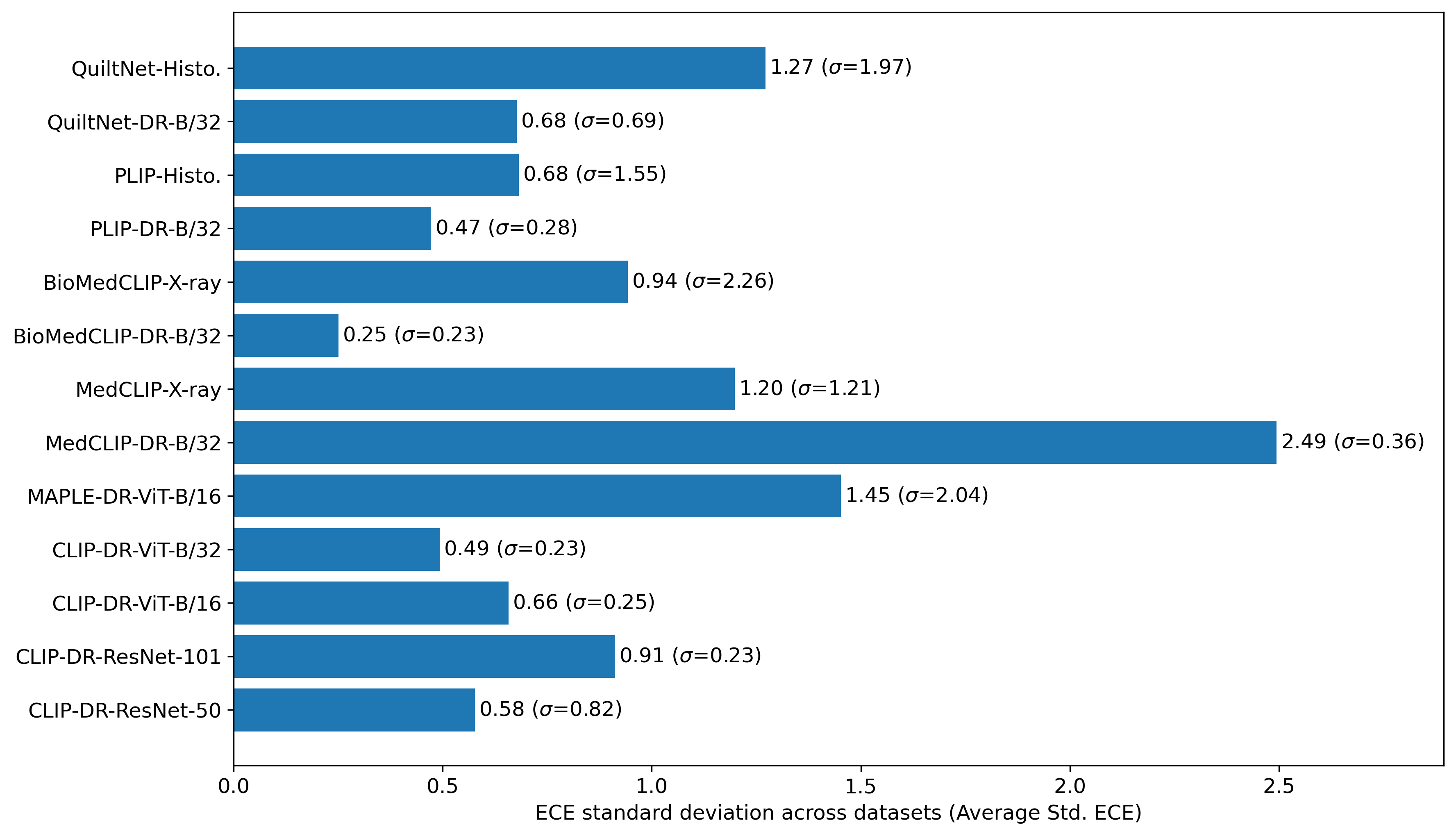}
  \vspace{-6pt}
  \caption{Comparison of seed variability, i.e., Std. of ECE and Acc ($\sigma$), across different backbones and three random seeds. Some backbones show stable calibration, while others are seed-sensitive.}
  \label{fig:random_seed}
\end{subfigure}

\vspace{-2pt}
\caption{Prompt-template sensitivity and random-seed sensitivity of calibration.}
\label{fig:prompt_seed_parallel}
\vspace{-0.3pt}
\end{figure*}

\noindent\textbf{Discussion:} Fig.~\ref{fig:prompt} reports the ECE standard deviation across datasets for each hard-prompt template, with mean ECE $\mu$ shown as a side label. Templates P1--P4 achieve low and stable ECE, with \textit{``a photo of a''} giving the most stable calibration. In contrast, the noisy P7 prompt shows high variance, indicating that calibration is highly sensitive to hard-prompt initialization. \textbf{This highlights the need for prompt-tuning strategies that remain robust to prompt wording and initialization choices.} The same sensitivity is observed with ACE and MCE, where early
templates such as P1-P2 remain comparatively stable while P6-P7
show substantially larger variability(Appendix~\ref{sec:supp_template_metrics}). 


\subsection{Stability: How do different seed initializations affect calibration performance?}

Random-seed variation is another important source of calibration instability in medical imaging, where limited data, long-tailed prevalence, label noise, and subtle findings can make confidence estimates sensitive to initialization. This instability can be further amplified under prevalence and acquisition shifts~\cite{guo2017calibration}. To assess reproducibility, we run each setup across three seeds and report the standard deviation of ECE and accuracy, allowing us to identify backbones with stable calibration estimates and those that are more seed-sensitive.
\\
\noindent\textbf{Experimental setup:} We assess random-seed sensitivity by running each backbone with the base calibration method across three seeds, $s \in {1,2,3}$, while keeping data splits, hyperparameters, and preprocessing fixed. Seed variability is measured as the standard deviation of ECE across seeds, with lower Std. ECE indicating more stable calibration.
\\
\noindent\textbf{Discussion:} Fig.~\ref{fig:random_seed} shows that calibration stability varies across seeds, with lower Std. ECE indicating more stable backbones. BioMedCLIP-DR-B/32, PLIP-DR-B/32, and CLIP-DR-ViT-B/32 show low seed variability, with Std. ECE values of 0.25, 0.47, and 0.49, respectively, indicating consistent calibration. In contrast, MedCLIP-DR-B/32 and MaPLe-DR-ViT/16 show higher variability, with Std. ECE values of 2.49 and 1.45, making them more seed-sensitive. Overall, \textbf{calibration performance and reproducibility depend on both seed initialization and backbone choice.} ACE and MCE likewise confirm that seed sensitivity is strongly backbone-dependent, with MedCLIP variants showing particularly large variation compared with several other backbones (Appendix~\ref{sec:supp_seed_metrics}).

\section{Multi-Class Margin Regularization (MCM)}
\begin{figure}[h!]
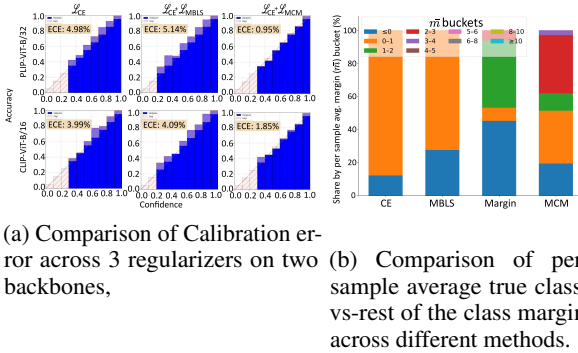

\vspace{-5pt}
\centering
\begin{subfigure}[t]{0.54\linewidth}
\centering
\includegraphics[width=\linewidth]{fig/LCE_LCM_LMBLS_V3.pdf}
\vspace{-15pt}
\captionsetup{font=footnotesize}
\caption{Comparison of Calibration error across 3 regularizers on two backbones, }
\label{fig:reliab}
\vspace{-10pt}
\end{subfigure}\hfill
\begin{subfigure}[t]{0.44\linewidth}
\centering
\includegraphics[width=\linewidth]{fig/buckets.pdf}
\vspace{-5pt}
\captionsetup{font=footnotesize}
\caption{Comparison of per-sample average true class-vs-rest of the class margin across different methods.}
\label{fig:mcm}
\vspace{-3pt}
\end{subfigure}
\caption{Miscalibration analyses.}
\vspace{-8pt}
\end{figure}

\noindent\textbf{Motivation:} Findings from research questions~\ref{subsec:domainshift}, \ref{subsec:backbone}, and \ref{subsec:calibmethod} show that ID calibration often fails to transfer to DS settings, increasing ECE, while backbone and calibration-method performance remain inconsistent. To understand this, we analyze DR reliability diagrams in Fig.~\ref{fig:reliab} and observe underconfidence across two backbones, both with and without MBLS~\cite{liu2022mbls}. Fig.~\ref{fig:mcm} further shows that CE and CE+MBLS concentrate most samples in the low-margin region $(\bar{m}\leq1)$, where the correct-class logit only slightly exceeds competing logits. This weak inter-class separation produces uncertain confidence and underconfidence. MBLS reduces extreme margins to limit overconfidence but does not sufficiently lift small margins, leaving logits concentrated in the low-margin region.
\vspace{-1pt}

\noindent\textbf{Multi-Class Margin Regularization:}
Our proposed \textbf{MCM} introduces a true-class-vs-rest margin objective that reshapes the inter-class logit distribution. It shifts samples from low-margin regions $(\bar{m}\leq1)$ toward moderate margins $(\approx2$--$4)$, improving class separation while controlling margin dispersion to produce reliable confidence. For logits $\mathbf{z}_i\in\mathbb{R}^{C}$ and label $y_i$, we define the true-vs-rest margin as $m_{i,k}=z_{i,y_i}-z_{i,k}$ for $k\neq y_i$, the per-sample average margin as $\bar m_i=\frac{1}{C-1}\sum_{k\neq y_i}m_{i,k}$, and the batch mean as $\mu=\frac{1}{B}\sum_{i=1}^{B}\bar m_i$. The all-pairs margin variance is then computed as $\sigma^2_{\text{all-pairs}}=\frac{1}{B(C-1)}\sum_{i=1}^{B}\sum_{k\neq y_i}(m_{i,k}-\mu)^2$, and the MCM loss is defined as:
\begin{align}
\mathcal{L}_{\mathrm{MCM}}
&=
-\,\alpha\,\mu
+
\beta\,\sigma^2_{\text{all-pairs}},
\qquad \alpha,\beta>0,
\\
\mathcal{L}_{\mathrm{total}}
&=
\mathcal{L}_{\mathrm{CE}}
+
\mathcal{L}_{\mathrm{MCM}}.
\end{align}
The $\mu$ term, weighted by $\alpha$, increases the margin between the ground-truth logit and competing class logits, directly reducing underconfidence. The $\sigma^2_{\text{all-pairs}}$ term, weighted by $\beta$, controls margin dispersion and prevents spurious confidence spikes by stabilizing the margin distribution. As shown in Fig.~\ref{fig:reliab}, MCM substantially reduces ECE across both backbones compared with existing methods. Unlike the concurrent \method{Margin} loss~\cite{sharifdeen2026calibratingprompttuningvisionlanguage}, which only considers the ground-truth versus second-highest logit margin, MCM uses all true-vs-rest margins. Fig.~\ref{fig:mcm} shows that \method{Margin} reduces samples in the $(0$--$1)$ bucket but remains unstable across other margin regions, including the $\leq0$ bucket where incorrect classes dominate. In contrast, MCM maintains a more stable inter-class margin distribution across all buckets.
 
\noindent\textbf{Results:}
Tab.~\ref{tab:calibration_id} provides a compact comparison of MCM against Base and ECCV\_ZS, which were selected as representative baselines based on their stronger ID consistency and comparatively favorable DS behavior in Fig.~\ref{fig:cal-method}. Across the 12 modality-specific settings, MCM achieves the lowest ECE in 10/12 ID
configurations while remaining competitive under DS. Complete
per-setting comparisons against all applicable calibration baselines, including Base, MDCA, LS, MBLS, ECCV\_ZS, ECCV\_Penalty, and Temperature Scaling, are reported in Appendix Tabs.~\ref{tab:idallmet} and~\ref{tab:dsallmet}.
Comparisons across additional prompt-tuning methods are also provided in the supplementary material.
 \begin{figure*}[h!]
 \centering
 \includegraphics[width=\linewidth]{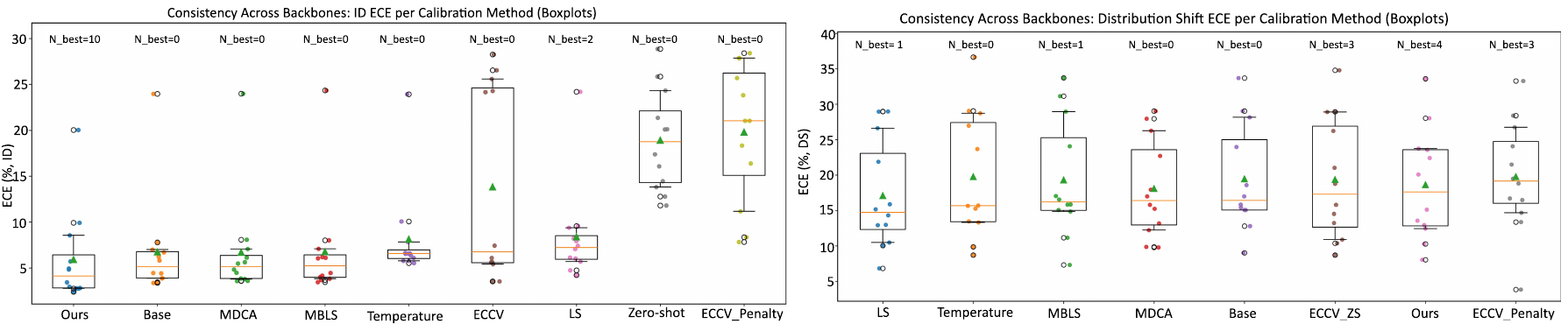}
 \caption{
 ECE (\%) across backbones for each calibration method under ID (top) and DS (bottom) settings. Boxplots show the interquartile range (IQR), whiskers indicate the 10--90\% range, and individual backbone results are overlaid as colored markers. N$_{\text{best}}$ denotes the number of backbones for which a method achieves the lowest ECE. Lower medians and tighter distributions indicate more consistent calibration across architectures.}
 \label{fig:ece-method}
 \vspace{-10pt}
 \end{figure*}
Fig.~\ref{fig:ece-method} extends the comparison to all evaluated calibration methods. MCM achieves the highest frequency of best calibration performance in ID, obtaining the lowest ECE in 10/12 settings. Under DS, MCM achieves the lowest ECE in 4/12 settings, while other methods, particularly ECCV\_ZS, remain competitive in individual configurations. The broader DS distributions further show
that strong ID calibration does not guarantee uniform robustness after distribution shift. Overall, the figure supports the conclusion that MCM provides strong and consistent ID calibration gains while its DS advantage is more setting-dependent.
\noindent\textbf{Statistical significance:}
A paired Wilcoxon analysis across the 12 matched configurations further supports the ID gains of MCM, with statistically significant improvements over ECCV\_ZS ($p=0.0212$) and ECCV\_Penalty ($p=0.0005$). Under DS, MCM remains competitive but is not uniformly superior. Full statistical results are provided in Appendix~\ref{sec:mcm_wilcoxon}.
\noindent\textbf{Hyperparameter transfer:}
The MCM hyperparameters are selected using validation data only and then fixed across all remaining backbones, modalities, and DS evaluations. A four-fold DR Leave One Dataset Out analysis independently selects the same $(\alpha,\beta)=(0.1,0.01)$ setting in every fold,
supporting its transferability without target-dataset tuning
(Appendix~\ref{sec:lodo_hyperparameter}).

\noindent\textbf{Failure cases:}
MCM primarily targets low-margin underconfidence and is therefore not uniformly effective across modalities. For PLIP-Histo. and QuiltNet-Histo., the modality-specific reliability patterns indicate a different calibration regime, with overconfidence more prominent than the low-margin underconfidence targeted by MCM (Appendix Fig.~\ref{fig:reliability_diagram}). Consequently, MCM
increases ECE in these cases.

\noindent\textbf{Ablation:} Tab.~\ref{tab:compana} shows that using either $-\alpha\mu$ or $+\beta\sigma^2_{\text{all-pairs}}$ alone worsens calibration, increasing ECE to 5.45\% and 6.37\%. Combining both terms reduces ECE to 3.43\% while preserving accuracy at 69.05\%, confirming that margin expansion and margin control are complementary for calibration.
\begin{table}[t!]
\centering

\begin{minipage}[t]{1.0\linewidth}
\centering
\resizebox{\linewidth}{!}{%
\begin{tabular}{@{}l|c|cc>{\columncolor{pink!20}}c|cc>{\columncolor{pink!20}}c@{}}
\toprule
\multirow{3}{*}{\textbf{Backbone}} & \multirow{3}{*}{\textbf{Metric}} & \multicolumn{3}{c|}{\textbf{ID}} & \multicolumn{3}{c}{\textbf{DS}} \\
\cline{3-8}
& & \multirow{2}{*}{\textbf{Base}}
& \textbf{ECCV\_} & \textbf{Ours} & \multirow{2}{*}{\textbf{Base}} & \textbf{ECCV\_} & \textbf{Ours} \\
& & & \textbf{ZS} & \textbf{(MCM)} & & \textbf{ZS} & \textbf{(MCM)} \\
\midrule

\multirow{2}{*}{CLIP-DR-ResNet-50}
& Acc. & 67.87 & 72.09 & 67.71 & 52.29 & 27.98 & 51.30 \\
& ECE & 4.40 & 28.26 & \textbf{2.96} & 15.07 & 18.77 & \textbf{10.26} \\
\midrule

\multirow{2}{*}{CLIP-DR-ResNet-101}
& Acc. & 68.29 & 67.69 & 68.19 & 41.91 & 59.63 & 51.73 \\
& ECE & 5.82 & 25.59 & \textbf{4.82} & 16.99 & 34.79 & \textbf{15.12} \\
\midrule

\multirow{2}{*}{CLIP-DR-ViT-B/16}
& Acc. & 69.18 & 68.56 & 69.05 & 32.90 & 38.51 & 33.35 \\
& ECE & 4.44 & 26.54 & \textbf{3.43} & 33.68 & \textbf{15.79} & 20.08 \\
\midrule

\multirow{2}{*}{CLIP-DR-ViT-B/32}
& Acc. & 68.13 & 72.09 & 67.79 & 43.58 & 29.03 & 46.59 \\
& ECE & 3.89 & 28.26 & \textbf{2.77} & 15.83 & 14.53 & \textbf{12.44} \\
\midrule

\multirow{2}{*}{MedCLIP-DR-B/32}
& Acc. & 59.82 & 59.85 & 59.74 & 52.69 & 59.91 & 59.83 \\
& ECE & 6.69 & 5.63 & \textbf{4.94} & 23.95 & 26.23 & \textbf{23.54} \\
\midrule

\multirow{2}{*}{MedCLIP-X-ray}
& Acc. & 65.77 & 65.96 & 65.83 & 79.15 & 79.02 & 79.15 \\
& ECE & 23.97 & 24.17 & \textbf{20.03} & 29.03 & 28.91 & \textbf{28.04} \\
\midrule

\multirow{2}{*}{BioMedCLIP-DR-B/32}
& Acc. & 67.47 & 59.85 & 66.80 & 59.45 & 59.11 & 59.26 \\
& ECE & 3.88 & 5.63 & \textbf{2.82} & 15.06 & 21.03 & \textbf{12.94} \\
\midrule

\multirow{2}{*}{BioMedCLIP-X-ray}
& Acc. & 73.56 & 72.94 & 73.70 & 74.56 & 73.63 & 78.84 \\
& ECE & 7.78 & 7.44 & \textbf{5.71} & 9.00 & 10.38 & \textbf{8.02} \\
\midrule

\multirow{2}{*}{PLIP-DR-B/32}
& Acc. & 68.05 & 67.79 & 67.97 & 55.40 & 54.27 & 47.16 \\
& ECE & 3.38 & 3.53 & \textbf{2.75} & 12.79 & \textbf{10.88} & 13.56 \\
\midrule

\multirow{2}{*}{PLIP-Histo.}
& Acc. & 85.43 & 86.27 & 85.30 & 52.83 & 65.35 & 53.88 \\
& ECE & 7.00 & \textbf{6.08} & 9.92 & 18.58 & \textbf{13.24} & 22.43 \\
\midrule

\multirow{2}{*}{QuiltNet-DR-B/32}
& Acc. & 66.54 & 67.79 & 66.49 & 40.08 & 50.50 & 27.92 \\
& ECE & 3.45 & 3.53 & \textbf{2.40} & 15.42 & \textbf{8.69} & 23.72 \\
\midrule

\multirow{2}{*}{QuiltNet-Histo.}
& Acc. & 86.73 & 87.87 & 86.87 & 46.46 & 44.93 & 46.76 \\
& ECE & 6.21 & \textbf{5.42} & 8.55 & \textbf{28.15} & 28.94 & 33.60 \\
\bottomrule
\end{tabular}%
}
\vspace{-2pt}
\caption{Overall average accuracy and ECE across 12 settings covering three modalities. ID: in-domain and DS: domain shift (out-domain).}
\label{tab:calibration_id}
\end{minipage}

\vspace{10pt}

\begin{minipage}[t]{1.0\linewidth}
\centering
\resizebox{\linewidth}{!}{%
\begin{tabular}{l|c|c|c|c|c|c} 
\toprule
\textbf{Loss Term} & \textbf{Metric} & \textbf{Aptos} & \textbf{eyepacs} & \textbf{messidor} & \textbf{messidor\_2} & \textbf{Avg}\\ 
\midrule
\multirow{2}{*}{CoOp~\cite{zhou2022coop}} & Acc. & 77.90 & 74.64 & 59.50 & 64.66 & 69.18\\ 
& ECE & 3.99 & 2.25 & 5.29 & 6.24 & 4.44 \\
\midrule
\multirow{2}{*}{MBLS~\cite{liu2022mbls}} & Acc. & 77.94 & 74.64 & 59.53 & 64.62 & 69.18\\ 
& ECE & 4.09 & 2.25 & 5.31 & 6.13 & 4.45 \\
\midrule
\multirow{2}{*}{$ -\,\alpha\,\mu$} & Acc. & 74.82 & 73.72 & 59.83 & 62.79 & 67.54\\ 
& ECE & 4.81 & 9.32 & 4.98 & 2.71 & 5.45 \\ 
\midrule
\multirow{2}{*}{$+\beta\,\sigma^2_{\text{all-pairs}}$} 
& Acc. & 78.26 & 74.49 & 58.42 & 64.62 & 68.94 \\ 
& ECE & 8.03 & 2.98 & 5.61 & 8.87 & 6.37 \\
\midrule
\rowcolor{pink!20}
& \textbf{Acc.} & 77.50 & 74.53 & 59.72 & 64.43 & 69.05 \\ 
\rowcolor{pink!20}
\multirow{-2}{*}{$-\,\alpha\,\mu + \beta\,\sigma^2_{\text{all-pairs}}$} & \textbf{ECE} & 1.85 & 3.25 & 5.09 & 3.53 & \textbf{3.43} \\ 
\bottomrule
\end{tabular}%
}

\caption{\footnotesize Loss Component analyses in our MCM.}
\label{tab:compana}
\vspace{-5pt}
\end{minipage}

\vspace{-5pt}
\end{table}
\section{Conclusion}
We introduced \textbf{MVC-Bench}, a large-scale benchmark for evaluating calibration in VLMs and Medical-VLMs under clinically relevant conditions. Our analysis shows that calibration is sensitive to DS, modality, and backbone choice. ID calibration does not reliably transfer, and medically aligned backbones are generally better calibrated than generic CLIP variants. It demonstrates that accuracy and calibration are not universally aligned, and no calibration or prompt-tuning method is consistently optimal across all settings. It shows that calibration varies with hard-prompt initialization and random seeds, highlighting the limitations of prompt initialization and the sensitivity of seed evaluation. Motivated by these findings, we proposed \textbf{Multi-Class Margin Regularization (MCM)}, which reduces ID calibration error and remains competitive under DS. Overall, MVC-Bench reveals that calibration should be treated as a first-class requirement for safe medical VLM deployment, evaluated through robustness, adaptation effectiveness, and stability under realistic variation.
\section{Limitations}
While MVC-Bench provides a large-scale calibration study of VLMs and
Medical-VLMs, its scope is limited to medical image classification.
We do not evaluate whether the same calibration behaviour extends to
segmentation, detection, report generation, retrieval, or medical VQA,
which require different uncertainty definitions and evaluation
protocols. Furthermore, MCM targets low-margin underconfidence and is therefore not a universal solution to all calibration failures. Its improvements are
less consistent in histopathology, particularly for PLIP-Histo. and
QuiltNet-Histo., where overconfidence appears more prominent. Although
we use a single hyperparameter setting across backbones and modalities,
further work is needed to understand modality-specific margin behavior
and calibration objectives under broader clinical shifts.




\clearpage
\bibliography{custom}
\setcounter{page}{1}
\appendix
\begin{center}
    {\Large\bfseries Appendices}
\end{center}

In this supplementary material, we provide the following:
\begin{enumerate}
    \item \textbf{Evaluation metrics and protocol} (Sec.~\ref{sec:evaluation})
    \item \textbf{Evaluated calibration baselines} (Sec.~\ref{sec:Evaluatedbaselines})
    \item \textbf{Loss component analyses} (Sec.~\ref{sec:Loss component analyses})
    \item \textbf{Prompt-learning Comparision} (Sec.~\ref{sec:Prompt-learning Comparision})
    \item \textbf{Prompt-template details} (Sec.~\ref{sec:Prompt-template details})
    \item \textbf{Reliability diagram for other backbones} (Sec.~\ref{sec:Reliability diagram for other backbones})
    \item \textbf{Hyperparameter sweep }(Sec.~\ref{sec:Hyperparameter sweep})
    \item \textbf{Statistical Analysis of Domain-Shift Rank Transfer }(Sec.~\ref{sec:ds_statistical_analysis})
    \item \textbf{Paired Statistical Comparison of MCM }(Sec.~\ref{sec:mcm_wilcoxon})
    \item \textbf{Scale of the Experimental Evaluation}(Sec.~\ref{sec:Other Metrics Comparison})
    \item \textbf{Other Metrics Analysis on Research Questions}(Sec.~\ref{sec:experiment_scale})
    \item \textbf{Brier Score Analyses}(Sec.~\ref{sec:Brier Score Analyses})
    \item \textbf{Weighted Subset ECE Analyses}(Sec.~\ref{sec:Weighted Subset ECE Analyses})
    \item \textbf{Results}(Sec.~\ref{sec:Results})
    \item \textbf{Declaration of LLM usage}(Sec.~\ref{sec:decllm})
\end{enumerate}
\setcounter{section}{0}
\setcounter{figure}{0}
\renewcommand{\thefigure}{A\arabic{figure}}
\setcounter{table}{0}
\renewcommand{\thetable}{A\arabic{table}}

\section{Evaluation metrics and protocol} 
\label{sec:evaluation}
The following metrics are used to evaluate the models’ performance: Accuracy (Acc.), Expected Calibration Error (ECE), Maximum Calibration Error (MCE), Adaptive Calibration Error (ACE), and reliability diagram.
\\
\textbf{Accuracy (Acc)} Proportion of correctly classified samples, indicating overall predictive performance.
\\
\textbf{Expected calibration error (ECE)~\cite{naeini2015obtaining}} The goal of calibration is to ensure the model's predicted probabilities accurately reflect the true likelihood of accuracy. If the outputs of the model are perfectly calibrated, then
\begin{equation}
\mathbb{P}\left(\hat{X}=X \mid \hat{P}=p\right) = p, \forall p~\in~[0, 1]
\end{equation}
where $X$ is the ground truth label, $\hat{X}$ is the predicted class label with its predicted confidence $\hat{P}=p$ that need to be calibrated. The ECE metric is computed as the weighted average
of the absolute difference between the accuracy and the mean confidence within each bin. Mathematically, ECE is formulated as follows:
\begin{equation}
\text{ECE} = \sum\limits_{z=1}^{Z}\frac{\left|C_z\right|}{M}\left|\text{acc}\left(C_z\right) - \text{conf}\left(C_z\right)\right|,
\end{equation} where $Z$ denotes the total number of bins. $C_z$ defines the image set with predicted confidence falling into the z-th bin. $|C_z|$ is the number of images within this bin, and M is the total number of predictions across all the probability bins. Furthermore, $\text{acc}\left(C_z\right)$, and $\text{conf}\left(C_z\right)$ represent the accuracy of the predictions and
average prediction confidence of all associated with bin $z$, respectively.
\\
\textbf{Maximum Calibration Error (MCE)~\cite{naeini2015obtaining}}
This metric captures the maximum discrepancy between confidence and accuracy across all bins, highlighting the worst-case deviation. It is defined as:
\[
\mathrm{MCE} = \max_{z \in \{1,\dots,Z\}} \left| \mathrm{acc}(C_z) - \mathrm{conf}(C_z) \right| ,
\]
where the terms are as previously defined.
\\
\textbf{Adaptive Calibration Error (ACE)~\cite{nixon2019measuring}}
This metric addresses some limitations of ECE by using an adaptive binning approach, ensuring that each bin has an equal number of samples. The ACE is calculated as:
\[
\mathrm{ACE} = \frac{1}{Z} \sum_{z=1}^{Z} \left| \mathrm{acc}(C_z) - \mathrm{conf}(C_z) \right| ,
\]
where \(Z\) is the number of bins, and other terms are as previously defined.
\\
\textbf{Reliability diagram} also called the calibration curve, provides a qualitative description of calibration. 
With better calibration, the points will lie closer to the main diagonal that extends from the bottom left to the top right of the reliability diagram. The points below the diagonal indicate that the model is overconfident, with predicted probabilities that are too large. Those above the diagonal indicate that the model is underconfident, and the predicted probabilities are too small. 

\section{Evaluated calibration baselines}\label{sec:Evaluatedbaselines} The following mix of post-hoc and training-time calibration baselines is used in this benchmarking study:

\textbf{Multi-class difference in
confidence and accuracy (MDCA)~\cite{hebbalaguppe2022stitch}} penalizes mismatches between the model's confidence and accuracy to capture class interactions. By adding this auxiliary loss, the model's predicted probabilities better align with its actual accuracy across multiple classes, thereby improving overall calibration. 
\\
\textbf{Label smoothing (LS)~\cite{muller2019does}} combats overfitting and overconfidence by replacing hard labels with a smoothed probability distribution.
This discourages the model from assigning full probability to a single class, leading to more moderate and calibrated predictions.
\\
\textbf{Margin-based label smoothing (MBLS)~\cite{liu2022devil}} is a class-balanced loss based on the number of samples to counter prevalence skew (e.g., the dominance of ``No DR''), mitigating calibration bias induced by class imbalance.
\\
\textbf{ECCV\_ZS~\cite{murugesan2024robust}} shifts the direction of the prior-weighted logit vector to match the correct category encoded in the one-hot label, and the logit norm normalizes logit values according to the ZS logit range.
\\
\textbf{ECCV\_penalty~\cite{murugesan2024robust}} when the constraint is not satisfied, i.e., there exist logit magnitudes outside the zero-shot logit range, the value of the penalty term increases, backpropagating gradients to modify the logit values according to the enforced constraint.
\\
\textbf{Temperature scaling~\cite{guo2017calibration}} adjusts the model’s logits by a scalar, learnable temperature parameter $\tau>0$ before the softmax layer, optimized on a held-out validation set by minimizing the negative log-likelihood (NLL).
A zero-shot variant can be applied in scenarios where a validation set is not available. 

\section{Loss component analyses}\label{sec:Loss component analyses}
\begin{figure*}[t!]
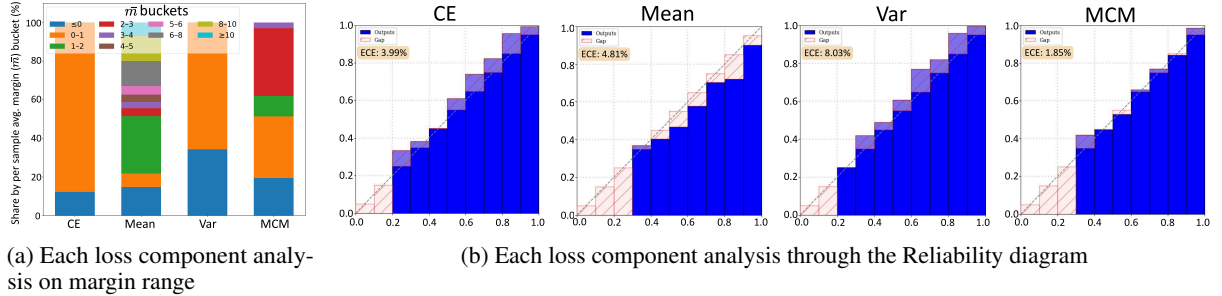

\vspace{-5pt}
\centering
\begin{subfigure}[t]{0.25\linewidth}
\centering
\includegraphics[width=\linewidth]{fig/Component.pdf}
\vspace{-15pt}
\captionsetup{font=footnotesize}
\caption{Each loss component analysis on margin range}
\label{fig:comp}
\vspace{-10pt}
\end{subfigure}\quad
\begin{subfigure}[t]{0.72\linewidth}
\centering
\includegraphics[width=\linewidth]{fig/Reliabilitywithname-1.pdf}
\vspace{-15pt}
\captionsetup{font=footnotesize}
\caption{Each loss component analysis through the Reliability diagram}
\label{fig:reliab}
\vspace{-10pt}
\end{subfigure}
\caption{Loss components analysis}
\vspace{-10pt}
\end{figure*}

As shown in Fig.~\ref{fig:comp}, the inclusion of the mean term ($-\alpha\mu$) with CE loss drives a larger inter-class margin range. However, this explicit expansion leads to the overconfidence issue observed in Fig.~\ref{fig:reliab}. In contrast, the variance term ($\beta\sigma^2_{\text{all-pairs}}$) restricts the margin range ($\approx 0\text{--}1$), resulting in underconfidence. Combining both terms (MCM) balances these effects, producing a stable margin field and lower ECE.

\section{Prompt-learning Comparison}\label{sec:Prompt-learning Comparision}

We compare five prompt-learning baselines (MaPLe, KgCoOp, PromptSRC, ProGrad, and HiCroPL)~\cite{zhou2022coop,khattak2023maple,khattak2023promptsrc,yao2023visual,zhu2023prompt,zheng2025hierarchicalcrossmodalpromptlearning} on diabetic retinopathy, histopathology, and chest X-ray benchmarks using Accuracy and Expected Calibration Error (ECE). We report each method and its calibrated variant (suffix “\_Ours”) in Table.~\ref{tab:dr_prompt_calib}.

Our calibration reduces \emph{average} ECE across each method, while retaining high accuracy. Per method, the average ECE drops from \(12.44\to 10.02\) for MaPLe, from \(10.57\to8.28\) for KgCoOp
from \(15.64\to12.62\) for PromptSRC, from \(18.59\to15.41\) for ProGrad, and from \(5.45\to3.74\) for HiCroPL.
Accuracy remains effectively unchanged: from \(59.21\to58.14\) for MaPLe, from \(60.08\to63.79\) for KgCoOp, from \(60.98\to62.15\) for PromptSRC,
from \(57.59\to57.43\) for ProGrad, and from \(72.09\to73.10\) for HiCroPL. (Table~\ref{tab:dr_prompt_calib}). 
Furthermore, Table.~\ref{tab:dr_prompt_calib_ds} shows that our method consistently reduces ECE in domain shift (DS) while retaining accuracy.

Our method provides substantial calibration gains without sacrificing accuracy cost, making it a practical drop-in for prompt-based VLM calibration.

\begin{table*}[t!]
\centering
\resizebox{\linewidth}{!}{%
\begin{tabular}{@{}l|c|*{9}{d}D@{}}
\toprule
\textbf{Method} & \textbf{Metric}
& \multicolumn{1}{c}{\rotatebox[origin=c]{45}{\parbox{15mm}{\centering \textbf{APTOS}}}}
& \multicolumn{1}{c}{\rotatebox[origin=c]{45}{\parbox{15mm}{\centering \textbf{EyePACS}}}}
& \multicolumn{1}{c}{\rotatebox[origin=c]{45}{\parbox{15mm}{\centering \textbf{Messidor}}}}
& \multicolumn{1}{c}{\rotatebox[origin=c]{45}{\parbox{15mm}{\centering \textbf{Messidor-2}}}}
& \multicolumn{1}{c}{\rotatebox[origin=c]{45}{\parbox{15mm}{\centering \textbf{Pannuke}}}}
& \multicolumn{1}{c}{\rotatebox[origin=c]{45}{\parbox{15mm}{\centering \textbf{Kather}}}}
& \multicolumn{1}{c}{\rotatebox[origin=c]{45}{\parbox{15mm}{\centering \textbf{DigestPath}}}}
& \multicolumn{1}{c}{\rotatebox[origin=c]{45}{\parbox{15mm}{\centering \textbf{RSNA18}}}}
& \multicolumn{1}{c}{\rotatebox[origin=c]{45}{\parbox{15mm}{\centering \textbf{COVIDX}}}}
& \multicolumn{1}{>{\columncolor{pink!20}}c@{}}{\rotatebox[origin=c]{45}{\parbox{15mm}{\centering \textbf{Average}}}} \\
\midrule
\multirow{2}{*}{MaPLe} & Acc. & 80.60 & 77.58 & 57.83 & 61.10 & 57.94 & 52.05 & 45.14 & 32.53 & 68.10 & 59.21 \\
& ECE & 3.97 & 3.02 & 5.86 & 8.56 & 10.77 & 16.21 & 32.39 & 11.55 & 19.60 & 12.44 \\
\midrule
\multirow{2}{*}{MaPLe\_Ours} & Acc. & 81.79 & 77.58 & 55.36 & 64.81 & 58.57 & 51.84 & 47.23 & 32.79 & 53.25 & 58.14 \\
& ECE & 3.09 & 2.92 & 5.46 & 7.66 & 10.58 & 14.45 & 27.70 & 7.08 & 11.26 & 10.02 \\
\midrule
\multirow{2}{*}{KgCoOp} & Acc. & 76.31 & 73.99 & 58.20 & 63.97 & 57.94 & 52.05 & 45.14 & 43.88 & 69.21 & 60.08 \\
& ECE & 3.85 & 3.45 & 4.82 & 7.18 & 10.77 & 16.21 & 32.39 & 10.07 & 6.38 & 10.57 \\
\midrule
\multirow{2}{*}{KgCoOp\_Ours} & Acc. & 75.99 & 74.30 & 57.58 & 63.91 & 61.79 & 58.03 & 68.11 & 44.26 & 70.11 & 63.79 \\
& ECE & 2.68 & 3.63 & 4.34 & 4.91 & 16.79 & 8.91 & 18.01 & 9.69 & 5.57 & 8.28 \\
\midrule
\multirow{2}{*}{PromptSRC} & Acc. & 84.42 & 77.85 & 60.61 & 65.19 & 53.48 & 79.59 & 12.31 & 40.78 & 74.55 & 60.98\\
& ECE & 15.99 & 10.14 & 9.63 & 10.32 & 17.46 & 9.18 & 52.73 & 1.11 & 14.19 & 15.64 \\
\midrule
\multirow{2}{*}{PromptSRC\_Ours} & Acc. & 84.71 & 78.04 & 63.47 & 69.99 & 53.36 & 79.45 & 14.65 & 40.36 & 75.30 & 62.15 \\
& ECE & 9.24 & 3.74 & 7.60 & 7.38 & 18.37 & 1.87 & 50.44 & 1.12 & 13.84 & 12.62 \\
\midrule
\multirow{2}{*}{ProGrad} & Acc. & 73.13 & 73.81 & 54.92 & 63.24 & 59.20 & 50.36 & 39.11 & 40.38 & 64.12 & 57.59 \\
& ECE & 27.67 & 27.16 & 13.72 & 23.77 & 19.15 & 4.75 & 30.86 & 9.04& 11.16 & 18.59 \\
\midrule
\multirow{2}{*}{ProGrad\_Ours} & Acc. & 74.15 & 73.82 & 56.17 & 63.95 & 61.01 & 56.76 & 24.04 & 40.64 & 66.31 & 57.43 \\
& ECE & 27.78 & 23.95 & 13.33 & 22.57 & 5.96 & 4.03 & 23.56 & 8.59 & 8.94 & 15.41 \\
\midrule
\multirow{2}{*}{HiCroPL} & Acc. & 86.78 & 77.57 & 66.20 & 71.45 & 73.16 & 85.29 & 72.57 & 40.44 & 75.32 & 72.09 \\
& ECE & 5.12 & 1.05 & 4.69 & 3.33 & 4.02 & 8.95 & 2.95 & 3.25 & 15.65 & 5.45 \\
\midrule
\multirow{2}{*}{HiCroPL\_Ours} & Acc. & 86.87 & 77.81 & 66.75 & 71.58 & 74.93 & 85.00 & 74.78 & 41.83 & 78.31 &  73.10\\
& ECE & 1.36 & 1.83 & 4.14 & 2.01 & 3.91 & 1.42 & 2.26 & 1.99  & 14.75 & 3.74  \\

\bottomrule
\end{tabular}%
}
\caption{Accuracy (Acc.) and calibration (ECE) on diabetic retinopathy, histopathology, and
chest X-ray benchmarks for different prompt-learning methods, with and without our calibration (suffix “\_Ours”). Our calibration consistently lowers ECE while maintaining accuracy.}
\label{tab:dr_prompt_calib}
\end{table*}
\begin{table*}[t!]
\centering
\resizebox{0.8\linewidth}{!}{%
\begin{tabular}{@{}l|c|*{6}{d}D@{}}
\toprule
\textbf{Method} & \textbf{Metric}
& \multicolumn{1}{c}{\rotatebox[origin=c]{45}{\parbox{15mm}{\centering \textbf{APTOS}}}}
& \multicolumn{1}{c}{\rotatebox[origin=c]{45}{\parbox{15mm}{\centering \textbf{EyePACS}}}}
& \multicolumn{1}{c}{\rotatebox[origin=c]{45}{\parbox{15mm}{\centering \textbf{Messidor-2}}}}
& \multicolumn{1}{c}{\rotatebox[origin=c]{45}{\parbox{15mm}{\centering \textbf{Kather}}}}
& \multicolumn{1}{c}{\rotatebox[origin=c]{45}{\parbox{15mm}{\centering \textbf{DigestPath}}}}
& \multicolumn{1}{c}{\rotatebox[origin=c]{45}{\parbox{15mm}{\centering \textbf{COVIDX}}}}
& \multicolumn{1}{>{\columncolor{pink!20}}c@{}}{\rotatebox[origin=c]{45}{\parbox{15mm}{\centering \textbf{Average}}}} \\
\midrule
\multirow{2}{*}{MaPLe} & Acc. & 46.64 & 67.61 & 57.09 & 20.61 & 36.95 & 67.01 & 49.32 \\
& ECE & 12.62 & 31.95 & 19.34 & 8.73 & 41.41 & 12.06 & 21.02 \\
\midrule
\multirow{2}{*}{MaPLe\_Ours} & Acc. & 46.11 & 68.55 & 57.19 & 20.58 & 37.22 & 65.31 & 49.16 \\
& ECE & 11.63 & 26.44 & 17.63 & 7.85 & 21.15 & 11.03 & 15.96 \\
\midrule
\multirow{2}{*}{KgCoOp} & Acc. & 7.88 & 3.66 & 10.07 & 23.86 & 33.59 & 47.82 & 21.15 \\
& ECE & 54.06 & 60.10 & 42.84 & 59.06 & 52.39 & 14.10 & 47.09 \\
\midrule
\multirow{2}{*}{KgCoOp\_Ours} & Acc. & 31.34 & 40.50 & 35.47 & 22.83 & 20.69 & 47.21 & 33.01 \\
& ECE & 26.51 & 13.43 & 10.72 & 52.74 & 50.47 & 13.33 & 27.87 \\
\midrule
\multirow{2}{*}{PromptSRC} & Acc. & 50.03 & 59.58 & 53.77 & 15.20 & 15.96 & 74.57 & 44.85 \\
& ECE & 33.73 & 23.82 & 13.54 & 4.91 & 53.02 & 14.03 & 23.84 \\
\midrule
\multirow{2}{*}{PromptSRC\_Ours} & Acc. & 49.23 & 58.88 & 54.05 & 15.90 & 15.94 & 75.29 & 44.88 \\
& ECE & 31.38 & 18.69 & 11.23 & 2.18 & 52.55 & 13.73 & 21.63 \\
\midrule
\multirow{2}{*}{ProGrad} & Acc. & 44.03 & 69.18 & 53.17 & 26.19 & 11.64 & 64.70 & 44.82 \\
& ECE & 12.13 & 23.02 & 17.85 & 50.03 & 72.84 & 7.21 & 30.51 \\
\midrule
\multirow{2}{*}{ProGrad\_Ours} & Acc. & 43.38 & 68.92 & 52.27 & 25.98 & 13.43 & 67.72 & 45.28 \\
& ECE & 13.63 & 22.22 & 11.96 & 46.56 & 70.05 & 6.88 & 28.55 \\
\midrule
\multirow{2}{*}{HiCroPL} & Acc. & 48.44 & 63.04 & 49.43 & 20.61 & 36.95 & 62.12 & 46.77 \\
& ECE & 24.04 & 6.96 & 3.31 & 8.73 & 41.41 & 3.53 & 14.66 \\
\midrule
\multirow{2}{*}{HiCroPL\_Ours} & Acc. & 47.57 & 62.96 & 49.77 & 26.03 & 42.65 & 62.19 & 48.53 \\
& ECE & 22.10 & 4.96 & 2.63 & 7.17 & 31.96 & 3.25 & 12.01 \\

\bottomrule
\end{tabular}%
}
\caption{Accuracy (Acc.) and calibration (ECE) under distribution shift. Messidor, Pannuke, and RSNA18 are the datasets used to train the model for the respective modalities' DS evaluation.}
\label{tab:dr_prompt_calib_ds}
\end{table*}

\section{Prompt-template details}\label{sec:Prompt-template details} 

We consider seven hard prompt templates (P1-P7) that differ in phrasing, context, and domain. We have used the prompt template as follows: \textit{``a photo of a'', ``a photo of a retina with'', ``a color fundus photograph showing diabetic retinopathy severity level as'', ``Classify diabetic retinopathy severity from this fundus photo of a'', ``You are classifying diabetic retinopathy on a fundus photo of a'', ``This picture illustrates diabetic retinopathy severity of a'', ``The DR level is''}.

\section{Reliability diagram for other backbones}\label{sec:Reliability diagram for other backbones}
\begin{figure*}[t]
    \centering
    \includegraphics[width=\linewidth]{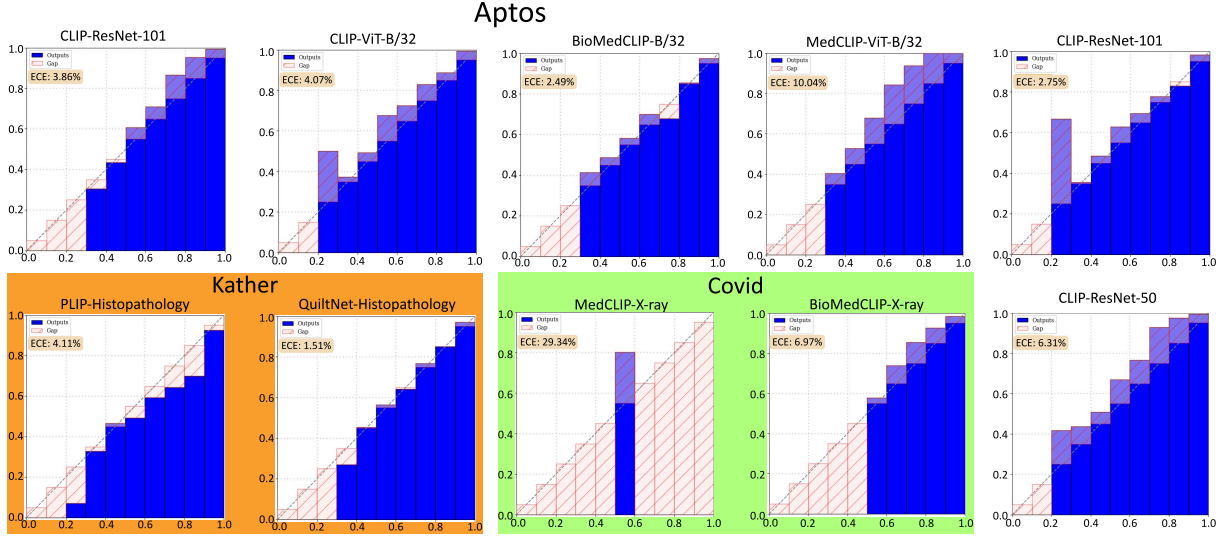}
    \caption{Reliability diagrams and Expected Calibration Error (ECE) analysis on medical datasets (Aptos, Kather, Covid). The diagrams present generic backbones (e.g., CLIP-ResNet-101, CLIP-ViT-B/32) and domain-specific models (e.g., PLIP, QuiltNet, MedCLIP, BioMedCLIP).}
    \label{fig:reliability_diagram}
\end{figure*}

Fig.\ref{fig:reliability_diagram} shows that most of the other backbones have underconfidence issues across the modalities.

\section{Hyperparameter sweep}\label{sec:Hyperparameter sweep}

We tuned MCM on CLIP-ViT-B/16 using a DR benchmark( CLIP-DR-ViT-B/16), averaging three seeds. We swept \(\alpha\in\{0.1,0.2,0.3\}\) and \(\beta\in\{0.01,0.05\}\) (Table~\ref{tab:hp}). The best calibration was at \((\alpha,\beta)\)=(0.1,0.01) with Acc. 69.05 and ECE 3.43. This setting remains optimal when averaging over four DR datasets.

At fixed \(\beta=0.01\), increasing \(\alpha\) raised accuracy but worsened calibration:
\((0.1,0.01)\rightarrow(0.3,0.01)\): Acc. (69.05\(\to\)70.49) alongside ECE (3.43\(\to\)6.79). At fixed \(\alpha\), moving from \(\beta=0.01\) to \(0.05\) generally increased ECE and reduced accuracy (e.g., \(\alpha=0.1\): Acc \(69.05\to68.29\), ECE \(3.43\to5.73\)); the exception was \(\alpha=0.3\), where both \(\beta\) settings remained poorly calibrated.

We select (0.1,0.01) as the default: it achieves the \emph{lowest} ECE without compromising accuracy, providing the best overall balance. To verify that this choice does not depend on the target evaluation dataset, we additionally conduct a Leave One Dataset Out (LODO) hyperparameter-transfer analysis in
Sec.~\ref{sec:lodo_hyperparameter}, where the same $(\alpha,\beta)=(0.1,0.01)$ setting is independently selected in all four DR folds.
\begin{table}[h!]
\centering
\resizebox{0.35\linewidth}{!}{%
\begin{tabular}{@{}lcr@{}}
\toprule
$(\alpha,\beta)$ & {Acc.} & {ECE} \\
\midrule
\rowcolor{pink!20}
(0.1, 0.01) & 69.05 & 3.43 \\
(0.2, 0.01) & 70.15 & 4.15 \\
(0.3, 0.01) & 70.49 & 6.79 \\
(0.1, 0.05) & 68.29 & 5.73 \\
(0.2, 0.05) & 69.12 & 5.90 \\
(0.3, 0.05) & 69.69 & 6.53 \\
\bottomrule
\end{tabular}%
}
\caption{\textbf{Hyperparameter settings} we used $\alpha$ was chosen from 0.1 to 0.3, and $\beta$ from 0.01 to 0.05. The backbone used was CLIP-ViT-B/16. Experiments were run across three seeds on a DR dataset. The average across 4 datasets indicated that the optimal hyperparameters were $(\alpha, \beta) = (0.1, 0.01)$.}
\label{tab:hp}
\vspace{-1.5em}
\end{table}
\subsection{Leave One Dataset Out Hyperparameter Transfer}
\label{sec:lodo_hyperparameter}

To verify that the selected MCM hyperparameters are not the result of
calibration-oriented tuning on the final evaluation datasets, we
perform a Leave One Dataset Out (LODO) analysis across the four DR
datasets. For each fold, the target dataset is completely excluded
from hyperparameter selection. We select $(\alpha,\beta)$ using only
the validation splits of the remaining three DR datasets and then
evaluate the selected setting on the held out target dataset. No final
test labels or shifted-domain test labels are used during
hyperparameter selection.
\begin{table*}[t]
\centering
\small
\resizebox{0.93\textwidth}{!}{%
\begin{tabular}{l l c c c c}
\toprule
\textbf{Held-out target}
& \textbf{Hyperparameter-selection data}
& \textbf{LODO-selected $(\alpha,\beta)$}
& \textbf{Base ECE}
& \textbf{Default MCM ECE}
& \textbf{LODO-MCM ECE}
\\
\midrule

APTOS
& EyePACS + Messidor + Messidor-2
& (0.1, 0.01)
& 3.99
& 1.85
& 1.85
\\

EyePACS
& APTOS + Messidor + Messidor-2
& (0.1, 0.01)
& 2.25
& 3.25
& 3.25
\\

Messidor
& APTOS + EyePACS + Messidor-2
& (0.1, 0.01)
& 5.29
& 5.09
& 5.09
\\

Messidor-2
& APTOS + EyePACS + Messidor
& (0.1, 0.01)
& 6.24
& 3.53
& 3.53
\\

\midrule

Average
& --
& --
& 4.44
& 3.43
& 3.43
\\

\bottomrule
\end{tabular}%
}
\caption{
Leave-One-Dataset-Out (LODO) analysis of MCM hyperparameter selection
across the four diabetic-retinopathy datasets. For each target dataset,
the target is excluded entirely from hyperparameter selection, and
$(\alpha,\beta)$ is selected using only the remaining three DR
validation splits. Every fold independently selects
$(\alpha,\beta)=(0.1,0.01)$, and LODO-MCM reproduces the ECE obtained
using the default MCM setting.
}
\label{tab:lodo_mcm}
\end{table*}
As shown in Table~\ref{tab:lodo_mcm}, all four LODO folds independently
select the same hyperparameter setting,
$(\alpha,\beta)=(0.1,0.01)$, as the default MCM configuration.
Consequently, LODO-MCM exactly reproduces the Default-MCM ECE on every
held-out dataset, with an average ECE of $3.43$ compared with $4.44$
for Base. This analysis shows that the selected hyperparameters are
stable across the DR datasets and that the reported MCM results are
not obtained by tuning $(\alpha,\beta)$ on the corresponding target
evaluation dataset. The same selected setting is subsequently fixed
across the remaining backbones, modalities, and domain-shift
experiments without per-backbone or per-modality retuning.

\section{Statistical Analysis of Domain-Shift Rank Transfer}
\label{sec:ds_statistical_analysis}

To further quantify the stability of calibration-method rankings under
domain shift, we aggregate the Kendall rank-correlation results across
all ID-to-DS evaluations. As summarized in
Table~\ref{tab:rank_transfer_statistics}, the analysis contains 30
ID-to-DS rank-transfer tests. Overall rank preservation is weak, with
a mean Kendall correlation of $-0.0598$ and a median correlation of
$-0.0476$. Among the 30 tests, 17 produce negative correlations,
whereas 13 produce positive correlations, and none show significantly
positive rank preservation at $p<0.05$.

These results should not be interpreted causally as identifying a
single source of distribution shift. The external datasets jointly
differ in acquisition conditions, patient populations, class
prevalence, dataset characteristics, and other distributional factors.
Rather, the analysis evaluates whether calibration methods selected
from in-domain validation retain their relative ranking when
transferred to realistic external-domain evaluation. The absence of
significantly positive rank preservation indicates that in-domain
calibration rankings are generally not reliable predictors of their
relative performance under such combined distributional changes.
\begin{table*}[t]
\centering
\small
\resizebox{0.95\textwidth}{!}{%
\begin{tabular}{c c c c c c c}
\toprule
\textbf{Rank-transfer tests}
& \textbf{Mean $\tau_b$}
& \textbf{Median $\tau_b$}
& \textbf{Non-positive $\tau_b$}
& \textbf{Negative $\tau_b$}
& \textbf{Positive $\tau_b$}
& \textbf{Significantly positive ($p<0.05$)}
\\
\midrule
30 & -0.0598 & -0.0476 & 17 & 17 & 13 & 0 \\
\bottomrule
\end{tabular}%
}
\caption{
Aggregate statistical summary of ID-to-DS calibration-method
rank transfer across 30 tests. Kendall's $\tau_b$ measures agreement
between calibration-method rankings obtained on the in-domain source
and corresponding external-domain datasets. Positive values indicate
greater rank preservation, while negative values indicate rank
reversals. No test exhibits significantly positive rank preservation
at $p<0.05$.
}
\label{tab:rank_transfer_statistics}
\end{table*}

\section{Paired Statistical Comparison of MCM}
\label{sec:mcm_wilcoxon}

We further compare MCM against each calibration baseline using matched
backbone--modality configurations. For every configuration, we compute

\[
\Delta \mathrm{ECE}
=
\mathrm{ECE}_{\mathrm{baseline}}
-
\mathrm{ECE}_{\mathrm{MCM}},
\]

such that a positive value indicates lower ECE for MCM. We then apply
a paired Wilcoxon signed-rank test across the 12 matched
backbone--modality settings. Table~\ref{tab:mcm_wilcoxon} summarizes
the resulting mean and median differences, the number of configurations
in which MCM obtains lower or higher ECE, and the corresponding
Wilcoxon $p$-values.
\begin{table*}[t]
\centering
\small
\resizebox{0.88\textwidth}{!}{%
\begin{tabular}{l l r r c c c}
\toprule
\textbf{Setting}
& \textbf{Baseline}
& \textbf{Mean diff}
& \textbf{Median diff}
& \textbf{MCM wins /12}
& \textbf{MCM losses /12}
& \textbf{Wilcoxon $p$}
\\
\midrule
ID & Base          & 0.8175  & 1.055  & 10 & 2 & 0.0881 \\
ID & MDCA          & 0.7842  & 1.140  & 10 & 2 & 0.0881 \\
ID & LS            & 2.4592  & 3.185  & 10 & 2 & 0.0549 \\
ID & MBLS          & 0.8583  & 1.135  & 10 & 2 & 0.0881 \\
ID & ECCV\_ZS      & 7.9167  & 2.270  & 10 & 2 & 0.0212 \\
ID & ECCV\_Penalty & 13.8900 & 15.840 & 11 & 1 & 0.0005 \\
ID & TS            & 2.2217  & 2.915  & 10 & 2 & 0.0134 \\
\midrule
DS & Base          & 0.8175  & 0.985  & 8 & 4 & 0.2847 \\
DS & MDCA          & -0.5633 & 1.380  & 7 & 5 & 0.4548 \\
DS & LS            & -1.5650 & 0.605  & 8 & 4 & 0.4849 \\
DS & MBLS          & 0.6475  & 1.775  & 8 & 4 & 0.2119 \\
DS & ECCV\_ZS      & 0.7033  & 1.480  & 7 & 5 & 0.4849 \\
DS & ECCV\_Penalty & 1.1308  & -0.050 & 6 & 6 & 0.4548 \\
DS & TS            & 1.1100  & 2.795  & 9 & 3 & 0.1167 \\
\bottomrule
\end{tabular}%
}
\caption{
Paired configuration-level comparison of MCM against existing
calibration baselines across 12 matched backbone--modality settings.
The difference is defined as
$\Delta\mathrm{ECE}
=\mathrm{ECE}_{\mathrm{baseline}}
-\mathrm{ECE}_{\mathrm{MCM}}$,
so positive values favor MCM. In ID settings, MCM obtains lower ECE
in 10/12 configurations against most baselines and in 11/12
configurations against ECCV\_Penalty. Under DS, MCM remains competitive
but does not uniformly dominate the baselines.
}
\label{tab:mcm_wilcoxon}
\end{table*}
The paired analysis provides additional statistical support for the
MCM results. In ID settings, MCM obtains lower ECE in 10 of 12 matched
configurations against Base, MDCA, LS, MBLS, ECCV\_ZS, and TS, and in
11 of 12 configurations against ECCV\_Penalty. The improvements are
statistically significant against ECCV\_ZS ($p=0.0212$),
ECCV\_Penalty ($p=0.0005$), and TS ($p=0.0134$). Improvements relative
to Base, MDCA, LS, and MBLS are numerically consistent but do not reach
the $p<0.05$ threshold. Under DS, MCM wins in 6--9 of 12 configurations
depending on the baseline, but none of the paired comparisons are
statistically significant. These results support our main conclusion
that MCM provides consistent ID calibration gains while remaining
competitive, rather than uniformly dominant, under domain shift.

\section{Scale of the Experimental Evaluation}
\label{sec:experiment_scale}

To clarify the scale of MVC-Bench, we provide a detailed breakdown of the experimental evaluation used in the core calibration benchmark. Since our study evaluates calibration across multiple modalities, datasets, backbone families, calibration methods, and random seeds, the total number of experiments is obtained from the Cartesian product of these factors. We distinguish between three levels of counting: \textit{experimental configurations}, \textit{seed-level controlled runs}, and \textit{metric-level evaluations}.

An \textit{experimental configuration} denotes a unique combination of modality, dataset, backbone, and calibration method. A \textit{seed-level controlled run} refers to one execution of an experimental configuration under a specific random seed. Since each configuration is repeated over three random seeds, the number of seed-level runs is three times the number of experimental configurations. Finally, a \textit{metric-level evaluation} refers to the calibration metrics computed from each seed-level run. In the core calibration analysis, we evaluate each run using three binning-based calibration metrics: Expected Calibration Error (ECE), Maximum Calibration Error (MCE), and Adaptive Calibration Error (ACE).

\subsection{In-Domain Evaluation}

For the in-domain (ID) evaluation, we consider three medical imaging modalities: fundus imaging, chest X-ray, and histopathology. The fundus imaging setting includes eight backbones, four datasets, three random seeds, and eight calibration settings, including the vanilla Base method. This results in:
\[
8 \times 4 \times 3 \times 8 = 768
\]
seed-level controlled runs.

For chest X-ray evaluation, we use two backbones, two datasets, three random seeds, and eight calibration settings. This results in:
\[
2 \times 2 \times 3 \times 8 = 96
\]
seed-level controlled runs.

For histopathology, we use two backbones, three datasets, three random seeds, and eight calibration settings. This results in:
\[
2 \times 3 \times 3 \times 8 = 144
\]
seed-level controlled runs.

Therefore, the in-domain evaluation contains:
\[
768 + 96 + 144 = 1008
\]
seed-level controlled runs.

Equivalently, before accounting for random seeds, the ID evaluation contains:
\[
\begin{aligned}
&(8 \times 4 \times 8) + (2 \times 2 \times 8) + (2 \times 3 \times 8) \\
&= 256 + 32 + 48 = 336 .
\end{aligned}
\]
unique backbone--dataset--method configurations.

\subsection{Domain-Shift Evaluation}

For the domain shift (DS) evaluation, we evaluate calibration transfer from in-domain datasets to shifted datasets. In this setting, the zero-shot calibration setting is excluded, resulting in seven calibration methods. The fundus domain shift setting includes eight backbones, three shifted datasets, three random seeds, and seven calibration methods. This gives:
\[
8 \times 3 \times 3 \times 7 = 504
\]
seed-level controlled runs.

For chest X-ray domain shift evaluation, we use two backbones, one shifted dataset, three random seeds, and seven calibration methods. This gives:
\[
2 \times 1 \times 3 \times 7 = 42
\]
seed-level controlled runs.

For histopathology domain shift evaluation, we use two backbones, two shifted datasets, three random seeds, and seven calibration methods. This gives:
\[
2 \times 2 \times 3 \times 7 = 84
\]
seed-level controlled runs.

Therefore, the domain shift evaluation contains:
\[
504 + 42 + 84 = 630
\]
seed-level controlled runs.

Before accounting for random seeds, the DS evaluation contains:
\[
\begin{aligned}
&(8 \times 3 \times 7) + (2 \times 1 \times 7) + (2 \times 2 \times 7) \\
&= 168 + 14 + 28 = 210
\end{aligned}
\]
unique backbone--dataset--method configurations.

\subsection{Total Scale of the Core Calibration Benchmark}

Combining the ID and DS evaluations, the core MVC-Bench calibration benchmark contains:
\[
336 + 210 = 546
\]
unique backbone--dataset--method configurations.

Since each configuration is repeated over three random seeds, the total number of seed-level controlled runs is:
\[
546 \times 3 = 1638.
\]

Thus, the core calibration benchmark consists of 1,638 seed-level controlled runs. Since each run is evaluated using ECE, MCE, and ACE, the total number of binning-based calibration metric evaluations is:
\[
1638 \times 3 = 4914.
\]

In addition to these calibration metrics, accuracy is also computed for each seed-level run to analyze the relationship between predictive performance and calibration reliability. Therefore, the reported experimental scale supports both accuracy-based and calibration-based analysis across in-domain and domain shift settings.

\subsection{Summary of Experimental Scale}

Table~\ref{tab:experiment_scale} summarizes the experiment count across modalities and evaluation settings.

\begin{table*}[h]
\centering
\small
\begin{tabular}{l l c c c c}
\toprule
\textbf{Setting} & \textbf{Modality} & \textbf{Backbones} & \textbf{Datasets} & \textbf{Methods} & \textbf{Seed-level Runs} \\
\midrule
ID & Fundus & 8 & 4 & 8 & 768 \\
ID & Chest X-ray & 2 & 2 & 8 & 96 \\
ID & Histopathology & 2 & 3 & 8 & 144 \\
\midrule
DS & Fundus & 8 & 3 & 7 & 504 \\
DS & Chest X-ray & 2 & 1 & 7 & 42 \\
DS & Histopathology & 2 & 2 & 7 & 84 \\
\midrule
\multicolumn{5}{r}{\textbf{Total seed-level controlled runs}} & \textbf{1638} \\
\bottomrule
\end{tabular}
\caption{Breakdown of the core MVC-Bench experimental scale. ID denotes in-domain evaluation and DS denotes domain-shift evaluation. The zero-shot calibration setting is excluded from the DS count, resulting in seven calibration methods for DS evaluation.}
\label{tab:experiment_scale}
\end{table*}

\subsection{Scope of the Count}

The above count refers only to the core calibration benchmark used to evaluate in-domain and domain shift calibration across backbones, datasets, calibration methods, and seeds. It does not include analyses such as prompt-tuning comparisons, Brier score evaluation, class-wise ECE analysis, or other supplementary diagnostic experiments. Therefore, the reported count of 1,638 seed-level controlled runs should be interpreted as a conservative estimate of the core experimental scale of MVC-Bench.

\section{Other Metrics Analysis on Research Questions}\label{sec:Other Metrics Comparison}
To provide a more comprehensive calibration assessment, we further evaluate our research questions using binning-based metrics, including Adaptive Calibration Error (ACE)~\cite{nixon2019measuring} and Maximum Calibration Error (MCE)~\cite{naeini2015obtaining}, in addition to ECE.
\subsection{Robustness: How do domain shifts affect calibration}
\label{sec:supp_ds_metrics}
\begin{figure*}[h!]
\centering
\includegraphics[width=\textwidth,clip]{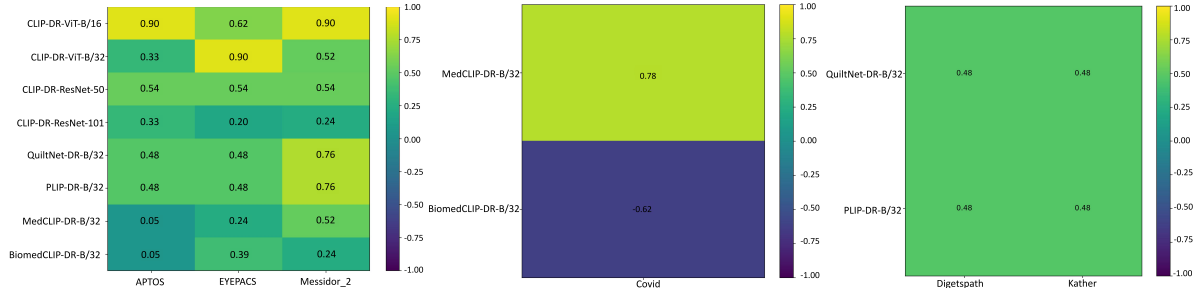} 
\vspace{-15pt}
\caption{Kendall rank correlation ($\tau$) between ACE-based method rankings on the In-domain(ID) source and each Domain shift(DS) dataset is shown for every backbone (rows = backbones, columns = DS sets). Cells are colored by $\tau$ (-$1$ to $1$) and annotated with values: higher/positive $\tau$ means rankings (and thus relative calibration quality) are preserved under shift, while near-zero/negative $\tau$ indicates rank reversals of the calibration under shift.}
\label{fig:dgace}
\vspace{-10pt}
\end{figure*}
\begin{figure*}[h!]
\centering
\includegraphics[width=\textwidth,clip]{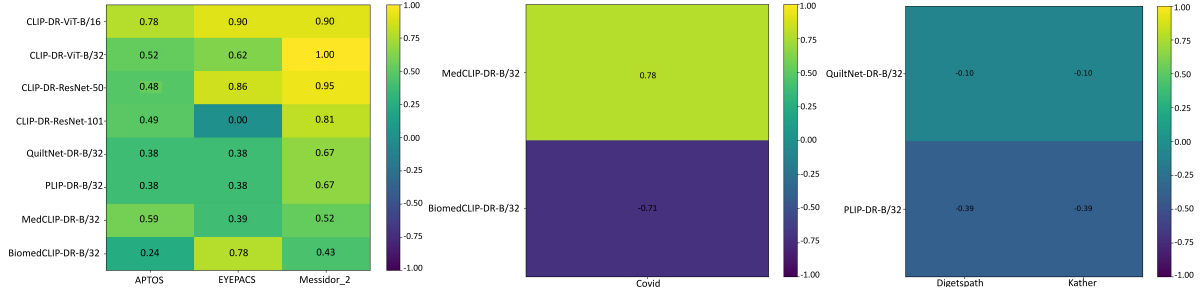} 
\vspace{-15pt}
\caption{Kendall rank correlation ($\tau$) between MCE-based method rankings on the In-domain(ID) source and each Domain shift(DS) dataset is shown for every backbone (rows = backbones, columns = DS sets). Cells are colored by $\tau$ (-$1$ to $1$) and annotated with values: higher/positive $\tau$ means rankings (and thus relative calibration quality) are preserved under shift, while near-zero/negative $\tau$ indicates rank reversals of the calibration under shift.}
\label{fig:dgmce}
\vspace{-10pt}
\end{figure*}

In Fig.\ref{fig:dgace} and Fig.\ref{fig:dgmce}, the ACE and MCE domain-shift analyses show that calibration transfer from in-domain to shifted datasets is metric-dependent, but domain shift still introduces clear instability. For ACE, the retinopathy setting shows several positive Kendall rank correlations, especially for CLIP-DR-ViT-B/16 and CLIP-DR-ViT-B/32, suggesting that adaptive-bin calibration rankings are partly preserved across APTOS, EyePACS, and Messidor-2. Histopathology also shows moderate consistency for both QuiltNet and PLIP, while the X-ray shift reveals a strong contrast: MedCLIP remains positively correlated, whereas BioMedCLIP shows negative transfer under COVID shift. For MCE, retinopathy again shows stronger rank preservation for several CLIP-based backbones, particularly on Messidor-2, indicating that worst-case calibration behaviour is sometimes stable across related fundus datasets. However, MCE exposes weaker or negative transfer in histopathology, especially for PLIP, and again shows poor transfer for BioMedCLIP under X-ray shift. Overall, compared with the ECE-based main analysis, ACE and MCE suggest that some calibration rankings can remain stable under closer domain shifts, but this stability is not universal across modalities, backbones, or metrics. This reinforces the takeaway that calibration should be evaluated under domain shift using multiple metrics, since average, adaptive-bin, and worst-case calibration errors may reveal different failure patterns. An aggregate statistical analysis over all 30 ID to DS rank transfer
tests is provided in Sec.~\ref{sec:ds_statistical_analysis} it shows
a median Kendall $\tau_b=-0.0476$, with 17/30 negative correlations
and no significantly positive rank-preservation case at $p<0.05$.

\subsection{Robustness: Which VLM-based backbones are better calibrated}
\label{sec:supp_backbone_metrics}
\begin{figure}[h!]
\centering
\includegraphics[width=1.0\linewidth]{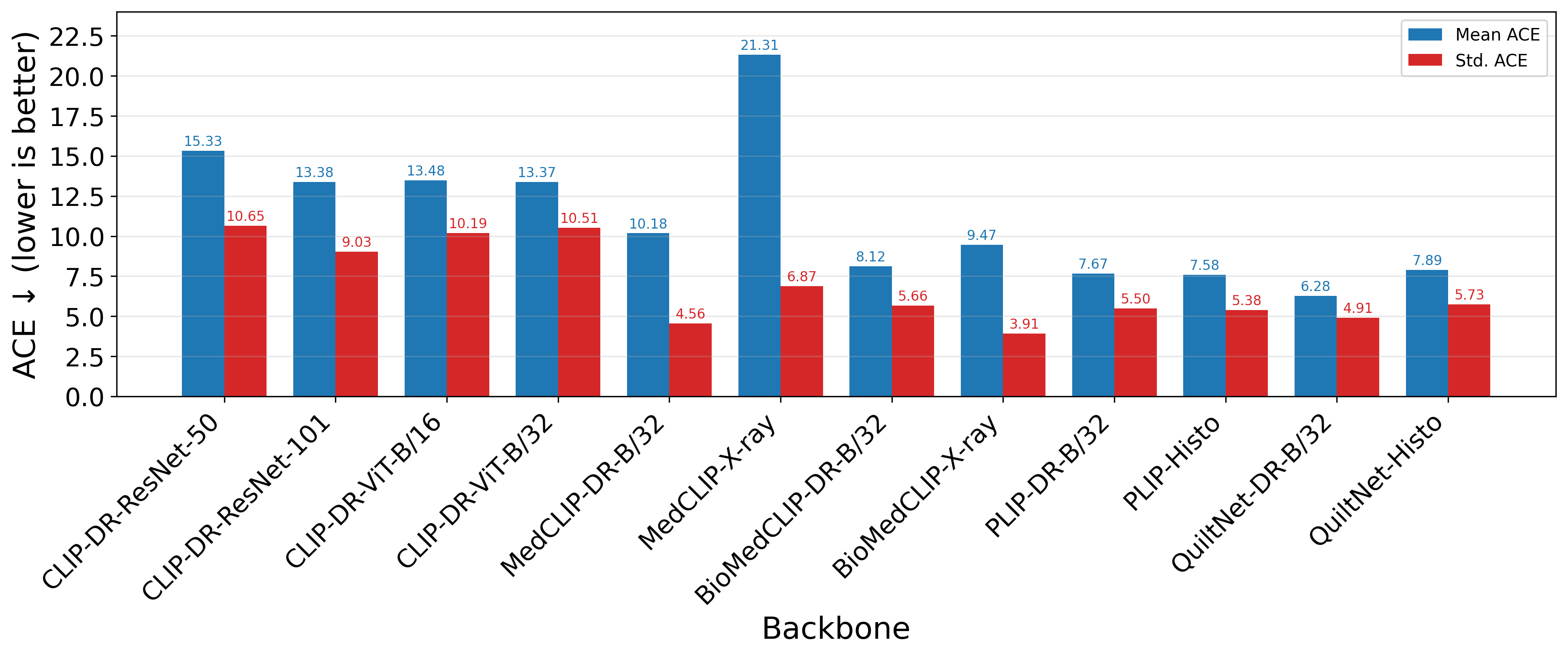}
\vspace{-10pt}
\caption{Comparison of mean and Std. of ACE for
various VLM-based backbones across different medical
modality inputs} 
\label{fig:ACE backbone}
\vspace{-5pt}
\end{figure}

\begin{figure}[h!]
\vspace{-8pt}
\centering
\includegraphics[width=1.0\linewidth]{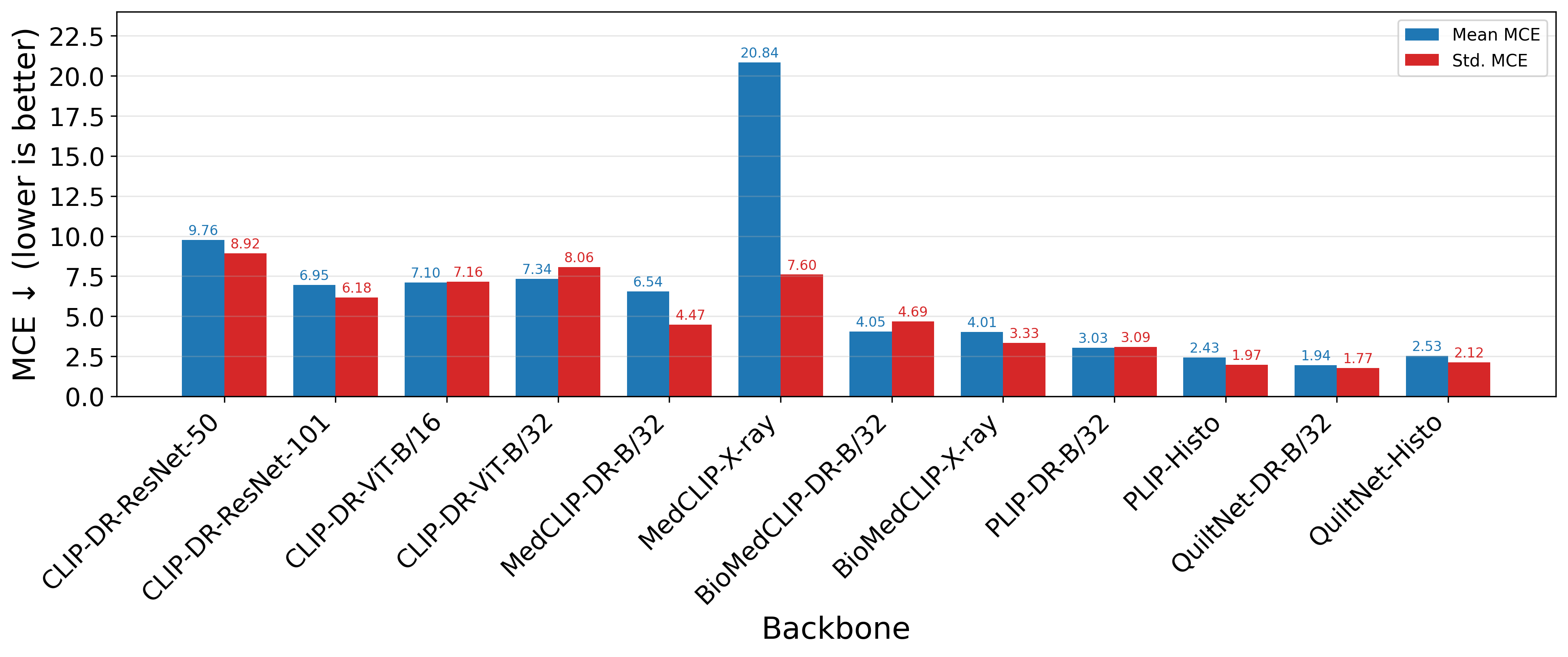}
\vspace{-10pt}
\caption{Comparison of mean and Std. of MCE for
various VLM-based backbones across different medical
modality inputs} 
\label{fig:MCE backbone}
\vspace{-5pt}
\end{figure}

The ACE and MCE analyses from Fig.\ref{fig:ACE backbone} and Fig.\ref{fig:MCE backbone} further confirm that backbone choice strongly affects calibration, consistent with the ECE-based finding in the main paper that calibration varies substantially across VLM and Medical-VLM backbones. Under MCE, the best worst-case calibration behaviour is observed for QuiltNet-DR-B/32, PLIP-Histo, PLIP-DR-B/32, and QuiltNet-Histo, all showing low mean MCE values below approximately 3\%, while BioMedCLIP-X-ray and BioMedCLIP-DR-B/32 also remain relatively stable. In contrast, MedCLIP-X-ray shows the poorest calibration, with a very high mean MCE of 20.84\%, indicating severe worst-bin calibration failures. A similar pattern is visible in ACE, where QuiltNet, PLIP, and BioMedCLIP variants generally achieve lower adaptive calibration error than generic CLIP backbones, while MedCLIP-X-ray again reports the highest ACE of 21.31\%. Generic CLIP variants show relatively high ACE values, especially CLIP-DR-ResNet-50, CLIP-DR-ResNet-101, and ViT-based CLIP models, suggesting that natural-image pretraining is less reliable for calibrated confidence in medical settings. Overall, ACE and MCE support the main ECE observation that domain-specific or medically aligned backbones are generally better calibrated than generic CLIP backbones, but they also reveal that some models, particularly MedCLIP-X-ray, can suffer from severe average and worst-case calibration instability.

\subsection{Effectiveness: How does calibration vary with accuracy for different calibration methods?}
\label{sec:supp_accuracy_metrics}
\begin{figure}[h!]
 \vspace{-10pt}
\centering
\includegraphics[width=1.0\linewidth]{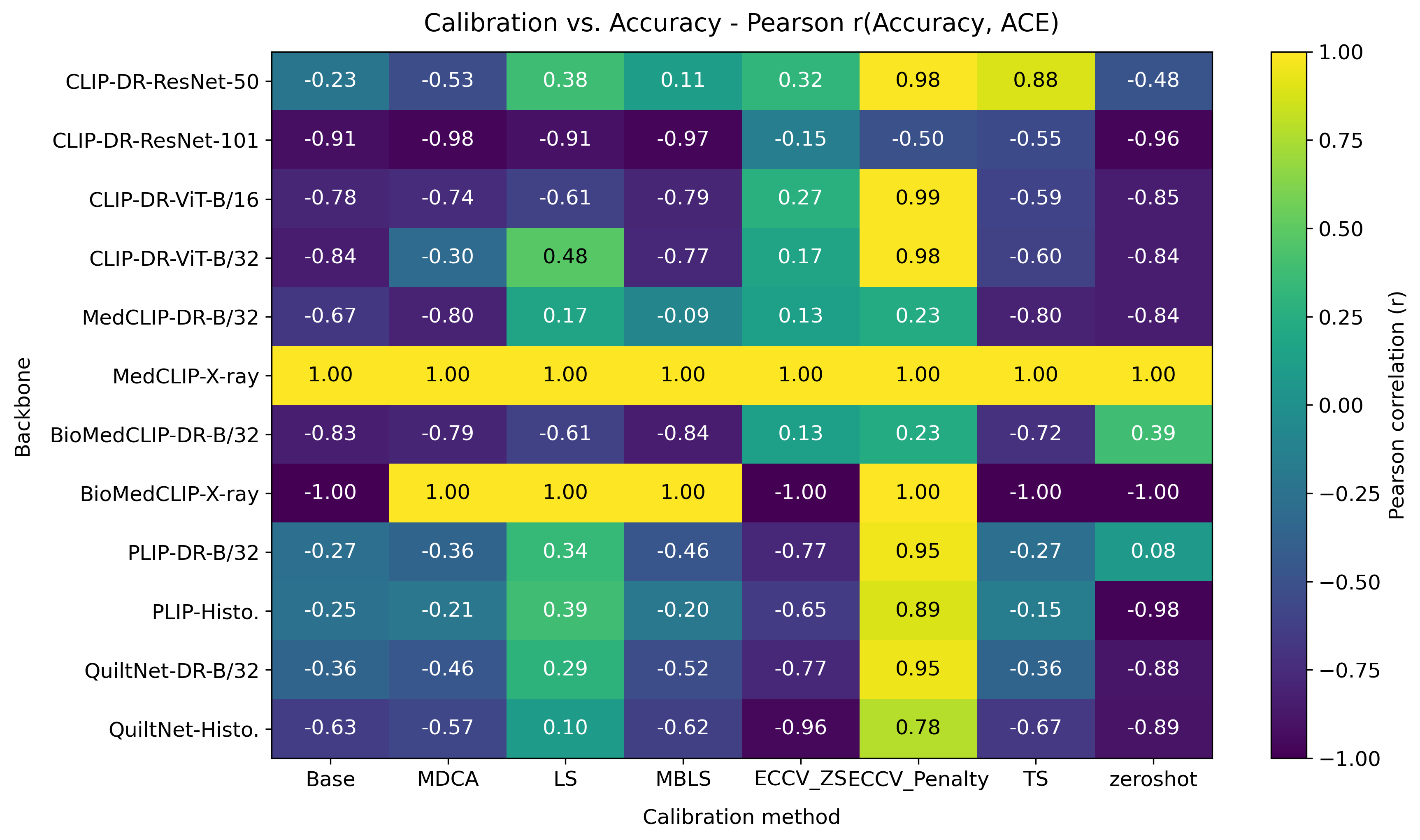}
\vspace{-10pt}
\caption{Pearson correlation matrix between accuracy and ACE evaluated across diverse calibration methods and backbones.}
\label{fig:pearsonace}
\vspace{-5pt}
\end{figure}

\begin{figure}[h!]
 \vspace{-10pt}
\centering
\includegraphics[width=1.0\linewidth]{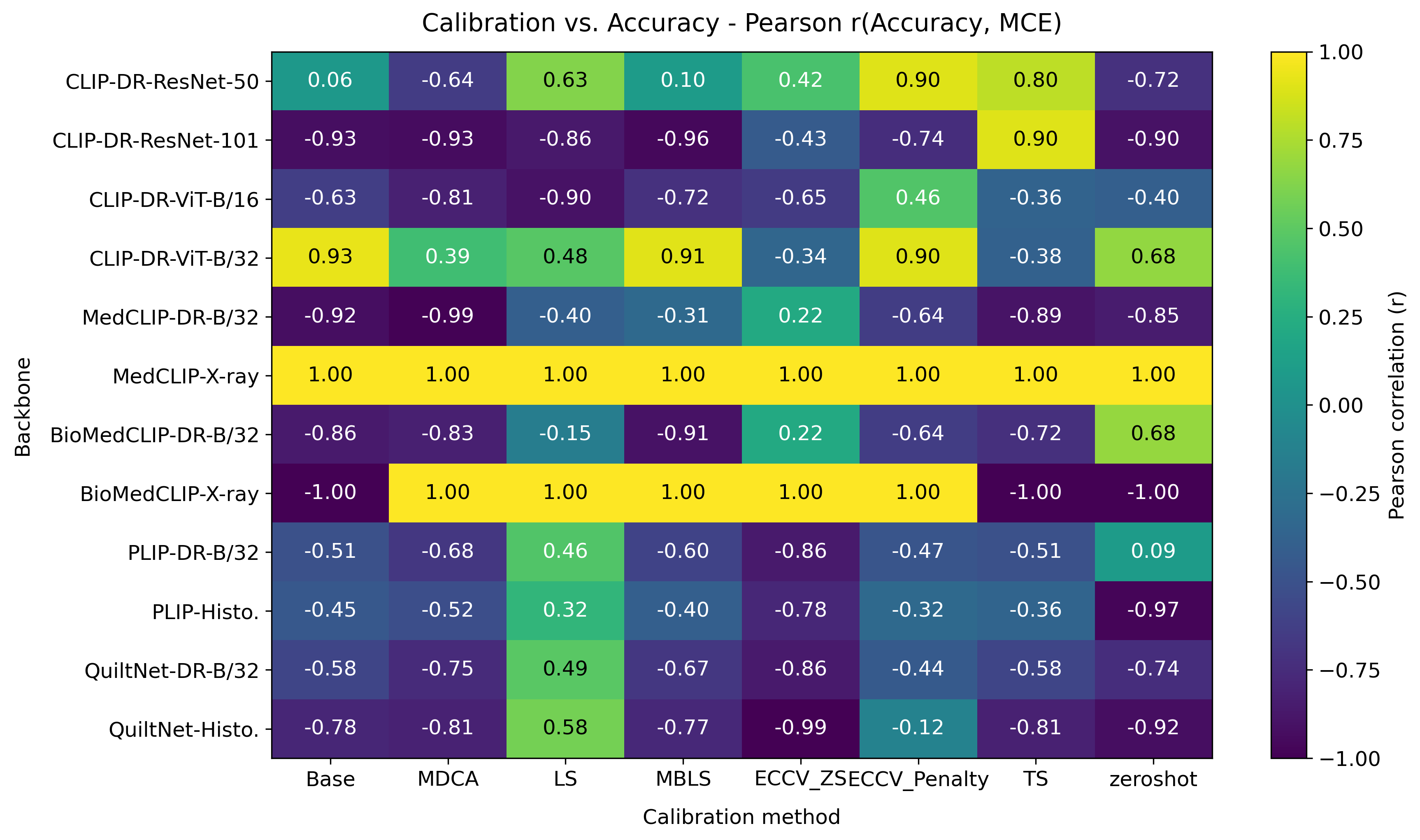}
\vspace{-10pt}
\caption{Pearson correlation matrix between accuracy and MCE evaluated across diverse calibration methods and backbones.}
\label{fig:pearsonmce}
\vspace{-5pt}
\end{figure}

Fig.\ref{fig:pearsonace} and Fig.\ref{fig:pearsonmce} show that ACE and MCE findings are Consistent with ECE-based findings. The ACE and MCE Pearson correlation analyses show that the relationship between accuracy and calibration is not universal, but strongly depends on the calibration method and backbone. In Fig.\ref{fig:pearson}, the ECE-based analysis showed that some methods, such as Base, MDCA, MBLS, and TS, often produced negative correlations, meaning that higher accuracy was generally associated with lower calibration error, while ECCV\_Penalty frequently showed positive correlations, indicating an accuracy–calibration trade-off. The ACE results follow a broadly similar pattern, where Base, MDCA, MBLS, and zeroshot mostly show negative correlations across many backbones, suggesting that accuracy improvement often aligns with better adaptive-bin calibration, whereas ECCV\_Penalty shows strong positive correlations for several CLIP-based backbones, meaning that higher accuracy can increase ACE and therefore worsen calibration. The MCE results also support the same conclusion, but with complementary observations in ECCV\_Penalty across some of the backbones, because the difference arises where ACE summarizes calibration behaviour across bins, while MCE is dominated by the single worst bin. Therefore, ACE and MCE reveal complementary aspects of calibration, ACE captures broader calibration degradation, while MCE highlights whether the most severe confidence error is amplified or reduced.  Overall, compared with the ECE-based findings, the ACE and MCE analyses confirm the same key takeaway that improving accuracy does not necessarily guarantee better calibration.

\subsection{ Effectiveness: Which calibration methods show consistent calibration performance}
\label{sec:supp_method_metrics}
 \begin{figure*}[h!]
 \centering
 \includegraphics[width=0.92\linewidth]{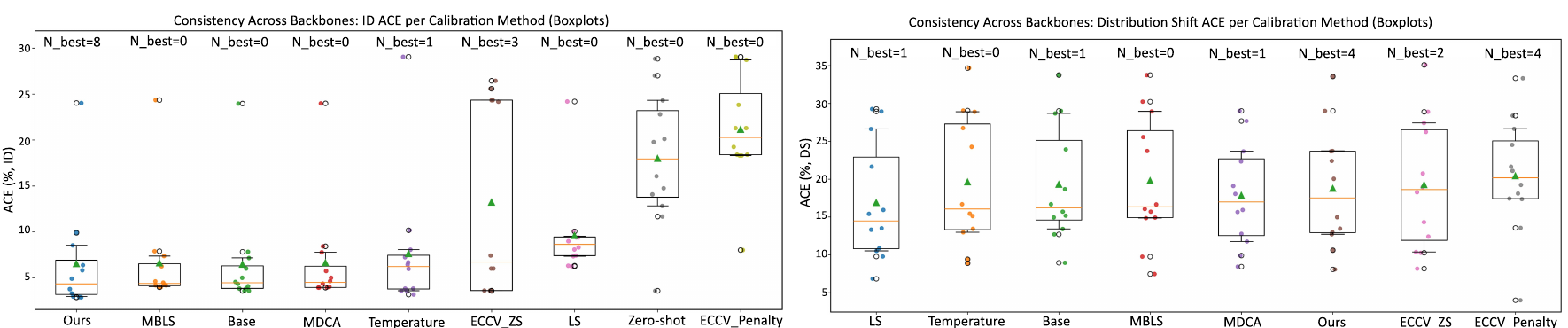}
 \caption{
ACE (\%) across backbones for each calibration method under ID (top) and DS (bottom) settings. Boxplots show the interquartile range (IQR), whiskers indicate the 10--90\% range, and individual backbone results are overlaid as colored markers. N$_{\text{best}}$ denotes the number of backbones for which a method achieves the lowest ACE. Lower medians and tighter distributions indicate more consistent calibration across architectures.}
 \label{fig:ace-method}
 \vspace{-10pt}
 \end{figure*}

Fig.~\ref{fig:ace-method} illustrates ACE results across backbones. In the ID setting, \textbf{Ours} achieves the highest $N_{\text{best}}$ ($=8$), indicating strong consistency, while other methods show limited improvement and higher variability. Under DS, performance deteriorates across all methods, though \textbf{Ours} and ECCV\_ZS obtain the highest $N_{\text{best}}$ counts.

\begin{figure*}[h!]
 \centering
 \includegraphics[width=0.92\linewidth]{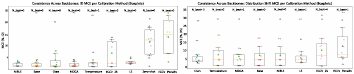}
 \caption{
 MCE (\%) across backbones for each calibration method under ID (top) and DS (bottom) settings. Boxplots show the interquartile range (IQR), whiskers indicate the 10--90\% range, and individual backbone results are overlaid as colored markers. N$_{\text{best}}$ denotes the number of backbones for which a method achieves the lowest MCE. Lower medians and tighter distributions indicate more consistent calibration across architectures.}
 \label{fig:mce-method}
 \vspace{-10pt}
 \end{figure*}

Fig.~\ref{fig:mce-method} shows MCE results, where \textbf{Ours} leads in ID ($N_{\text{best}} = 8$) and DS ($N_{\text{best}} = 5$), followed by ECCV\_ZS. 


Overall, the results are illustrating consistent pattern across all metrics for all the methods for both ID and DS. Furthermore, our method demonstrates superior ID calibration consistency across metrics and maintains competitive DS performance compared to existing methods.


\subsection{Effectiveness: How do different prompt-tuning methods interact with calibration?}
\label{sec:supp_prompt_metrics}
\begin{figure}[h!]
\centering
\includegraphics[width=1.0\linewidth]{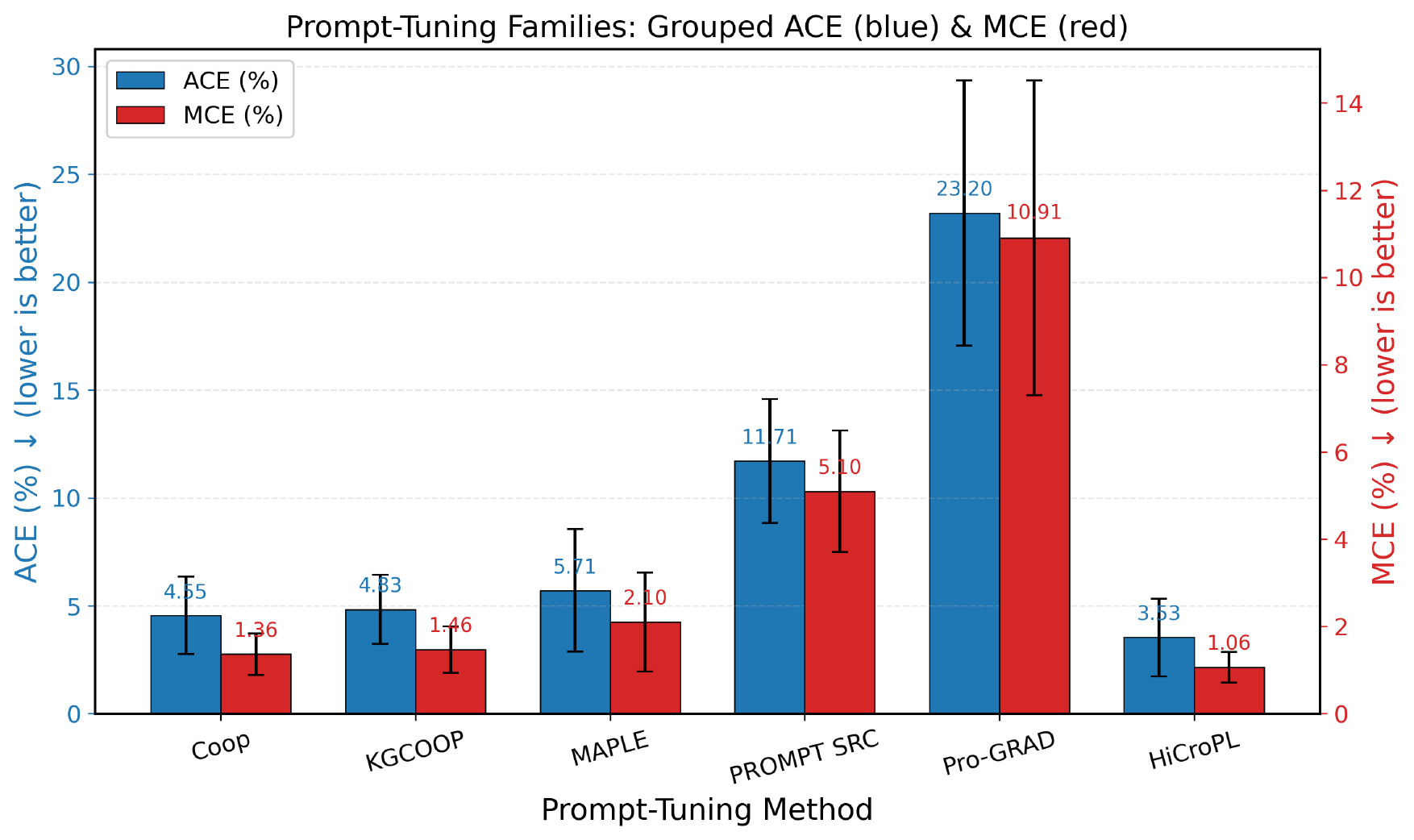}
\vspace{-10pt}
\caption{Comparison of ACE and MCE for different prompt-tuning methods.} 
\label{fig:ACEMCE prompt}
\vspace{-5pt}
\end{figure}

Fig.\ref{fig:ACEMCE prompt} shows that ACE and MCE have similar trend as how ECE based observation for how different prompt-tuning families behave on calibration. HiCroPL provides the most stable calibration, achieving the lowest ACE 3.53\% and MCE 1.06\%, followed by CoOp and KgCoOp, which also maintain relatively low average and worst-case calibration errors.   

\subsection{Stability: How does hard-prompt initialization affect calibration?}
\label{sec:supp_template_metrics}
\begin{figure}[h!]
\centering
\includegraphics[width=1.0\linewidth]{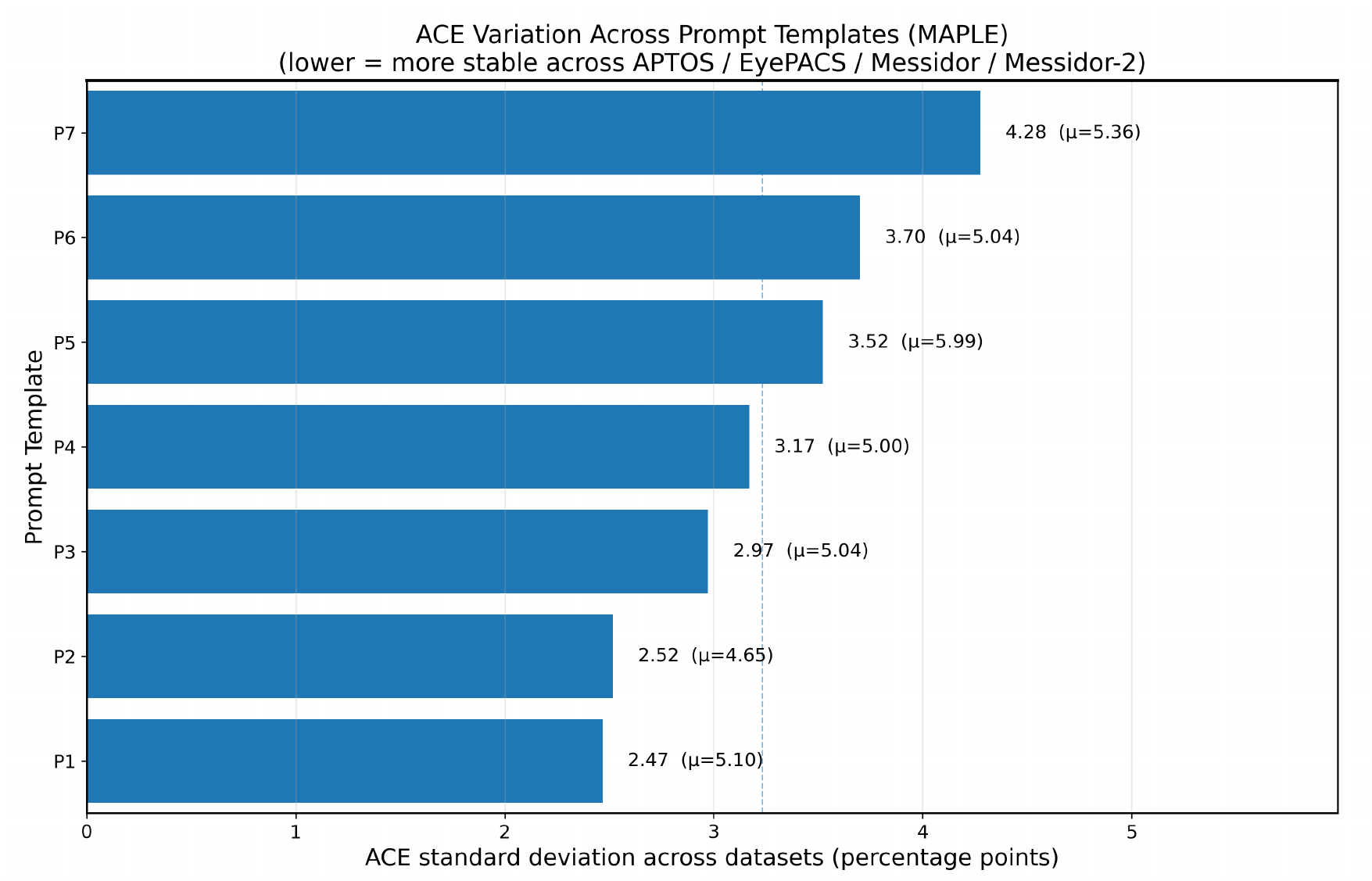}
\vspace{-10pt}
\caption{Comparison of Std. and mean of ACE ($\mu$) across datasets for seven different hard-prompt templates (P1-P7). Templates P1-P4 have low and stable ACE across datasets, while P7 has a high variance. This highlights that the prompting method is sensitive to hard-prompt initialization.} 
\label{fig:ACE hardprompt}
\vspace{-5pt}
\end{figure}

\begin{figure}[h!]
\centering
\includegraphics[width=1.0\linewidth]{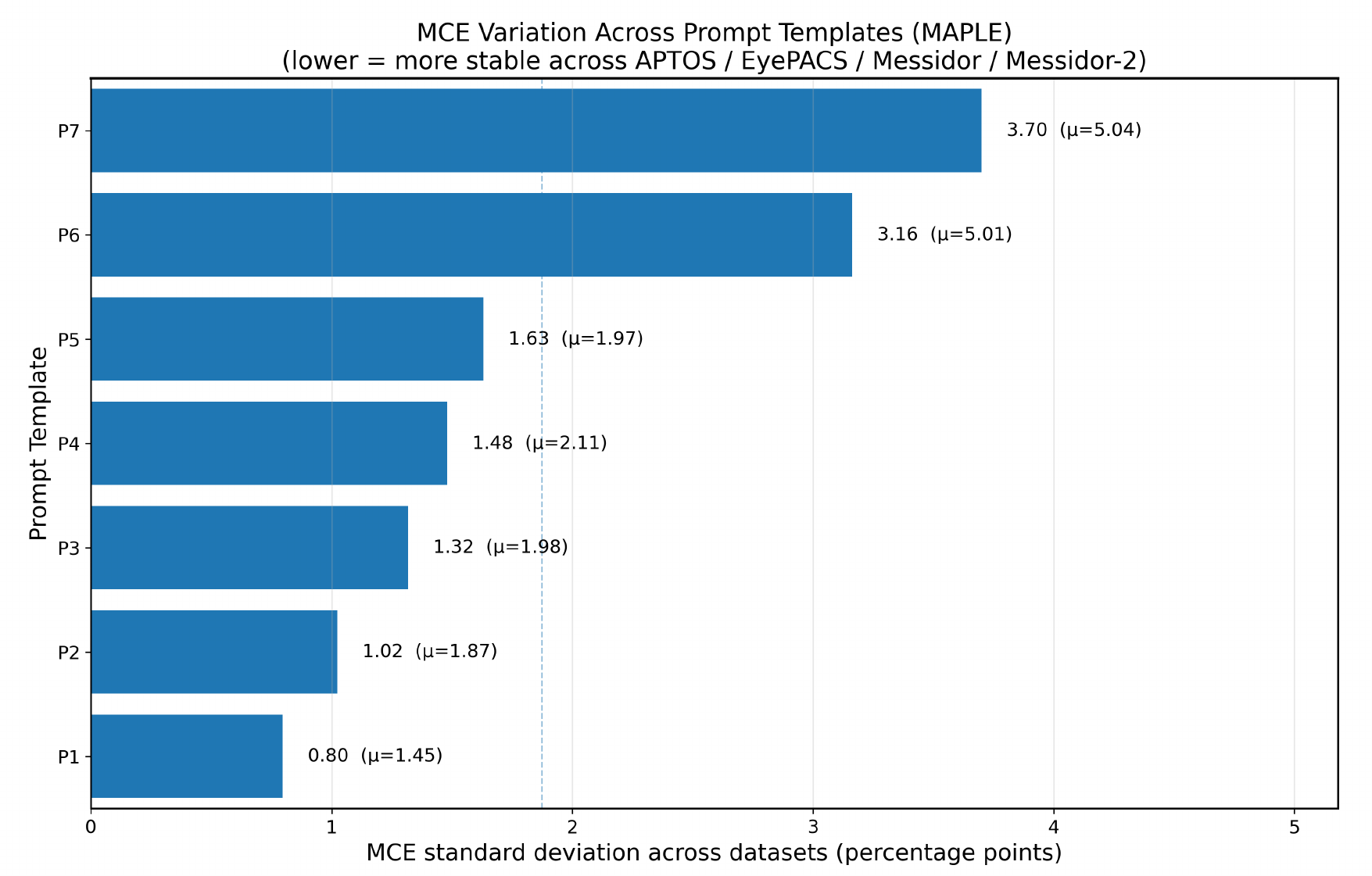}
\vspace{-10pt}
\caption{Comparison of Std. and mean of MCE ($\mu$) across datasets for seven different hard-prompt templates (P1-P7). Templates P1-P4 have low and stable MCE across datasets, while P7 has a high variance. This highlights that the prompting method is sensitive to hard-prompt initialization.} 
\label{fig:MCE hardprompt}
\vspace{-5pt}
\end{figure}

In Fig.\ref{fig:ACE hardprompt} and Fig.\ref{fig:MCE hardprompt}, the ACE and MCE analyses show that calibration is clearly sensitive to hard-prompt initialization. For MCE, early templates such as P1 and P2 are the most stable, with low standard deviations of 0.80 and 1.02, and low mean MCE values of 1.45 and 1.87, respectively. A similar trend is observed for ACE, where P1 and P2 also show the lowest variation, while later templates become increasingly unstable. In particular, P6 and P7 show much larger variations in both metrics, with P7 producing the highest MCE variation 3.70 and ACE variation 4.28, suggesting that poorly aligned or noisy prompt initializations can strongly destabilize calibration across datasets. Overall, both ACE and MCE support the ECE-based finding that prompt initialization is not a minor design choice, it directly affects both average calibration behaviour and worst-case confidence reliability.

\subsection{Stability: How do different seed initializations affect calibration performance?}
\label{sec:supp_seed_metrics}
\begin{figure}[h!]
\centering
\includegraphics[width=1.0\linewidth]{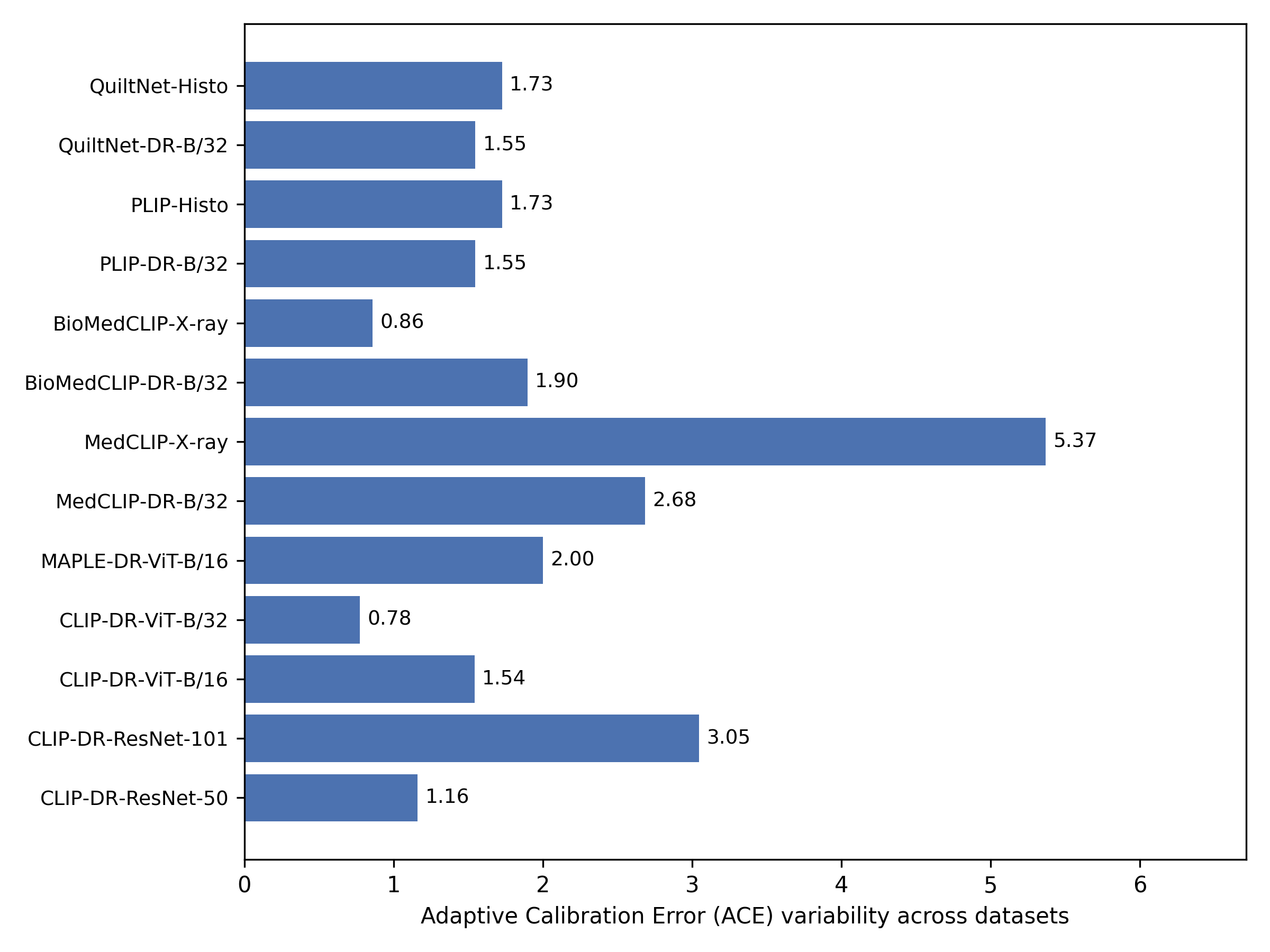}
\vspace{-10pt}
\caption{Comparison of seed variability, i.e., Std. of ACE, for various backbones across three random seeds.} 
\label{fig:ACE seed}
\vspace{-5pt}
\end{figure}

\begin{figure}[h!]
\centering
\includegraphics[width=1.0\linewidth]{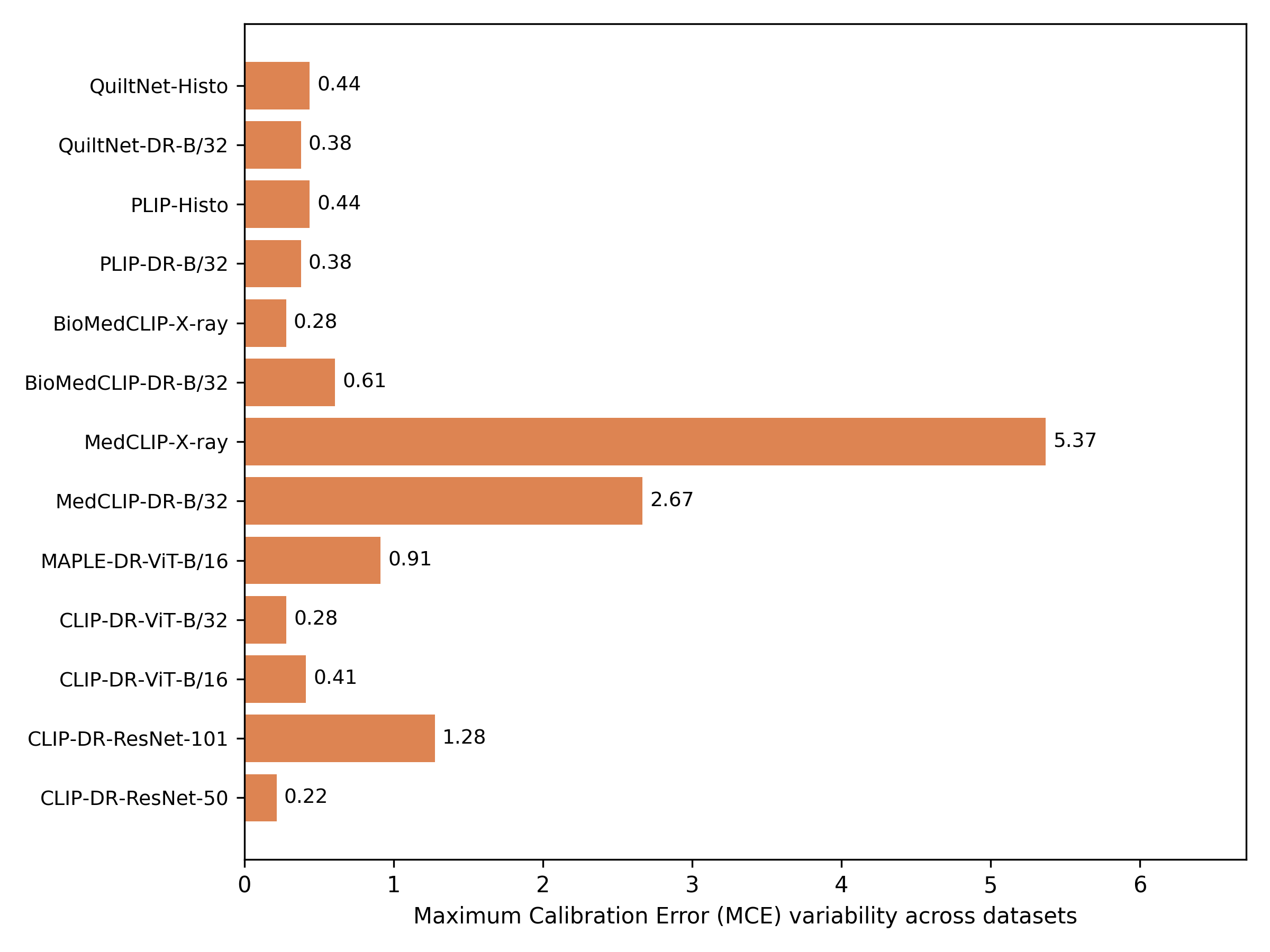}
\vspace{-10pt}
\caption{Comparison of seed variability, i.e., Std. of MCE, for various backbones across three random seeds.} 
\label{fig:MCE seed}
\vspace{-5pt}
\end{figure}

In Fig.\ref{fig:ACE seed} and Fig.\ref{fig:MCE seed}, the ACE and MCE analyses show that calibration stability differs substantially across backbones when evaluated beyond ECE. Most backbones exhibit relatively low MCE variability, such as CLIP-DR-ResNet-50 0.22, BioMedCLIP-X-ray 0.28, CLIP-DR-ViT-B/32 0.28, PLIP-DR-B/32 0.38, and QuiltNet-DR-B/32 0.38, indicating that their worst-case calibration error remains comparatively stable. A similar trend is observed in ACE, where BioMedCLIP-X-ray 0.86 and CLIP-DR-ViT-B/32 0.78 show strong stability, suggesting that their adaptive-bin calibration behaviour is less sensitive. However, MedCLIP-based models show much higher variability across all the metrics, especially MedCLIP-X-ray, which records the largest ACE and MCE variability 5.37, followed by MedCLIP-DR-B/32 with ACE 2.68 and MCE 2.67. This indicates that some backbones may appear acceptable under average calibration metrics but remain unstable when calibration is evaluated across datasets or initialization conditions. Overall, ACE and MCE support the ECE-based finding that calibration performance is sensitive to seed initialization and backbone choice, with domain-aligned models such as BioMedCLIP, PLIP, and CLIP-DR-ViT-B/32 generally showing more stable behaviour across ECE, ACE, and MCE than MedCLIP, while worst-case calibration failures are most visible through all the metrics.

\section{Brier Score Analyses}
\label{sec:Brier Score Analyses}
\textbf{Brier Score (BS)}~\cite{Brier1950VERIFICATIONOF}
is a bin-free proper scoring rule that measures the mean squared difference
between the predicted probability distribution and the true one-hot encoded
class label. Unlike ECE-based metrics, it does not depend on confidence bins and
therefore provides a complementary assessment of probabilistic prediction
quality. For a multi-class classification problem, it is defined as:
\[
\mathrm{BS}
=
\frac{1}{N}
\sum_{i=1}^{N}
\sum_{c=1}^{C}
\left(p_{ic} - y_{ic}\right)^2,
\]
where \(N\) is the number of samples, \(C\) is the number of classes,
\(p_{ic}\) is the predicted probability assigned to class \(c\) for sample
\(i\), and \(y_{ic}\in\{0,1\}\) is the corresponding one-hot ground-truth label.
A lower Brier Score indicates better probabilistic prediction quality.

\begin{figure}[h!]
\centering
\includegraphics[width=1.0\linewidth]{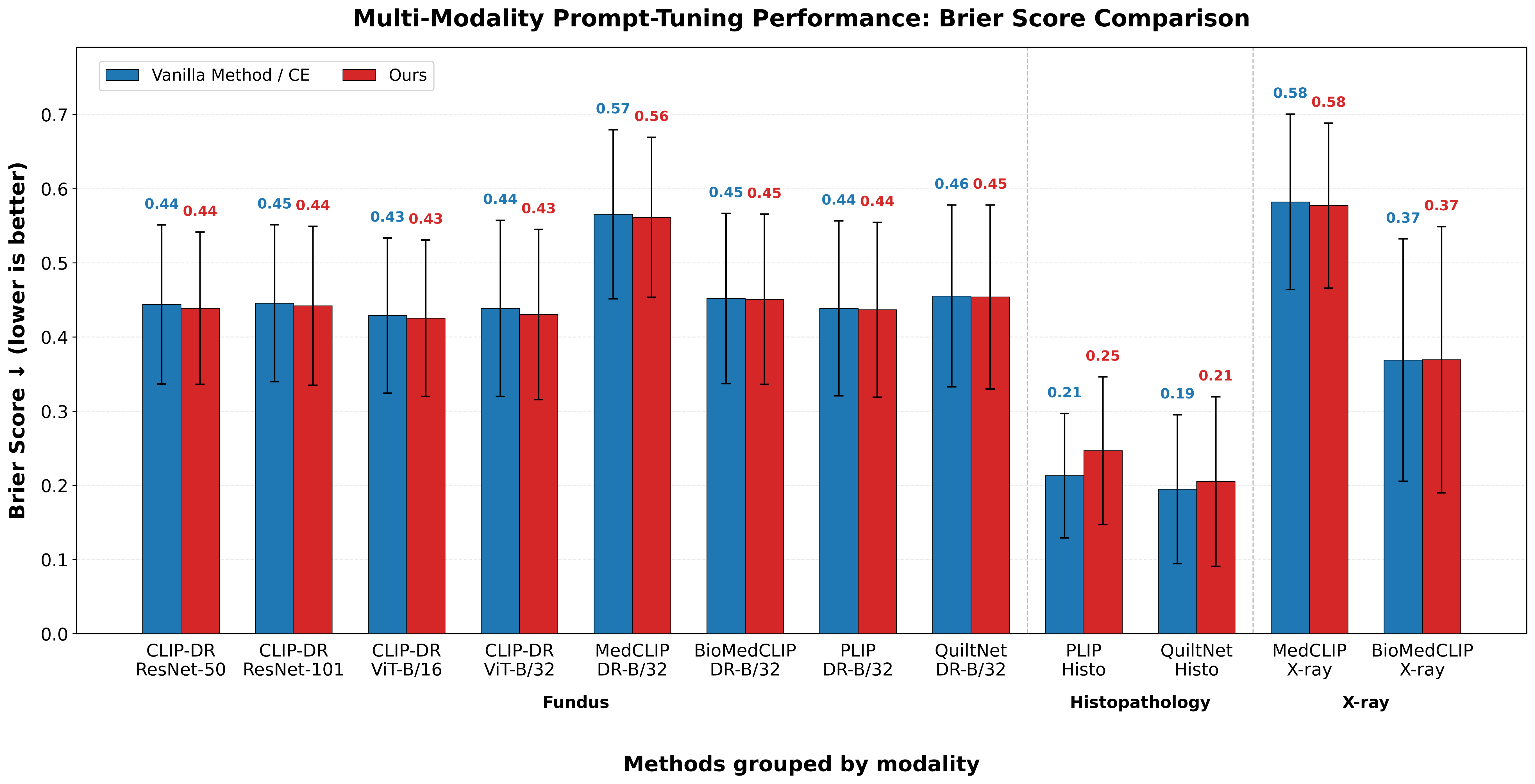}
\vspace{-10pt}
\caption{Brier Score Analyses across Backbones and Modalities.} 
\label{fig:brier score ana}
\vspace{-5pt}
\end{figure}

In Fig.\ref {fig:brier score ana}, the Brier score analysis shows that integrating our loss function generally preserves or slightly improves probabilistic prediction quality compared with the vanilla CE baseline across most backbones, especially in the fundus modality. For fundus models, our method gives small but consistent reductions in Brier score for several backbones, such as CLIP-DR-ResNet-101, CLIP-DR-ViT-B/32, MedCLIP-DR-B/32, and QuiltNet-DR-B/32, while maintaining similar performance for CLIP-DR-ResNet-50, CLIP-DR-ViT-B/16, BioMedCLIP-DR-B/32, and PLIP-DR-B/32. This suggests that the proposed loss improves calibration-related probability quality without substantially harming predictive behaviour. In the X-ray modality, the Brier scores remain almost unchanged for both MedCLIP-X-ray and BioMedCLIP-X-ray, indicating that our loss preserves the baseline probabilistic performance but does not produce a strong improvement. However, in histopathology, the Brier score slightly increases for PLIP-Histo and QuiltNet-Histo, suggesting that the benefit of the proposed loss is less consistent in this modality. Overall, the Brier score results support the main calibration findings by showing that our loss is mostly stable across modalities, with the strongest benefit in fundus datasets, while also highlighting that modality-specific behaviour should be considered when applying calibration-oriented training objectives.

\section{Weighted Subset ECE Analyses}
\label{sec:Weighted Subset ECE Analyses}
\textbf{Weighted Subset ECE (WSECE)~\cite{Obadinma2021ClasswiseCA}.}
Weighted Subset ECE measures calibration separately for each class and
then combines the class-level calibration errors using the number of samples
in each class as weights. Therefore, classes with more test samples contribute
more to the final score, while smaller classes still remain explicitly
evaluated. This metric is useful for imbalanced datasets because it shows
whether the model is well calibrated across different classes, rather than
only reporting one overall ECE value.

For class \(c\), let \(\mathrm{ECE}_c\) be the ECE computed using only the
samples that belong to class \(c\). If \(N_c\) is the number of samples in
class \(c\), and \(N\) is the total number of samples, WSECE is defined as
\[
\mathrm{WSECE}
=
\sum_{c=1}^{K}
\frac{N_c}{N}
\mathrm{ECE}_c ,
\]
where \(K\) is the number of classes. A lower WSECE indicates better
class-wise calibration.

\begin{figure}[h!]
\centering
\includegraphics[width=1.0\linewidth]{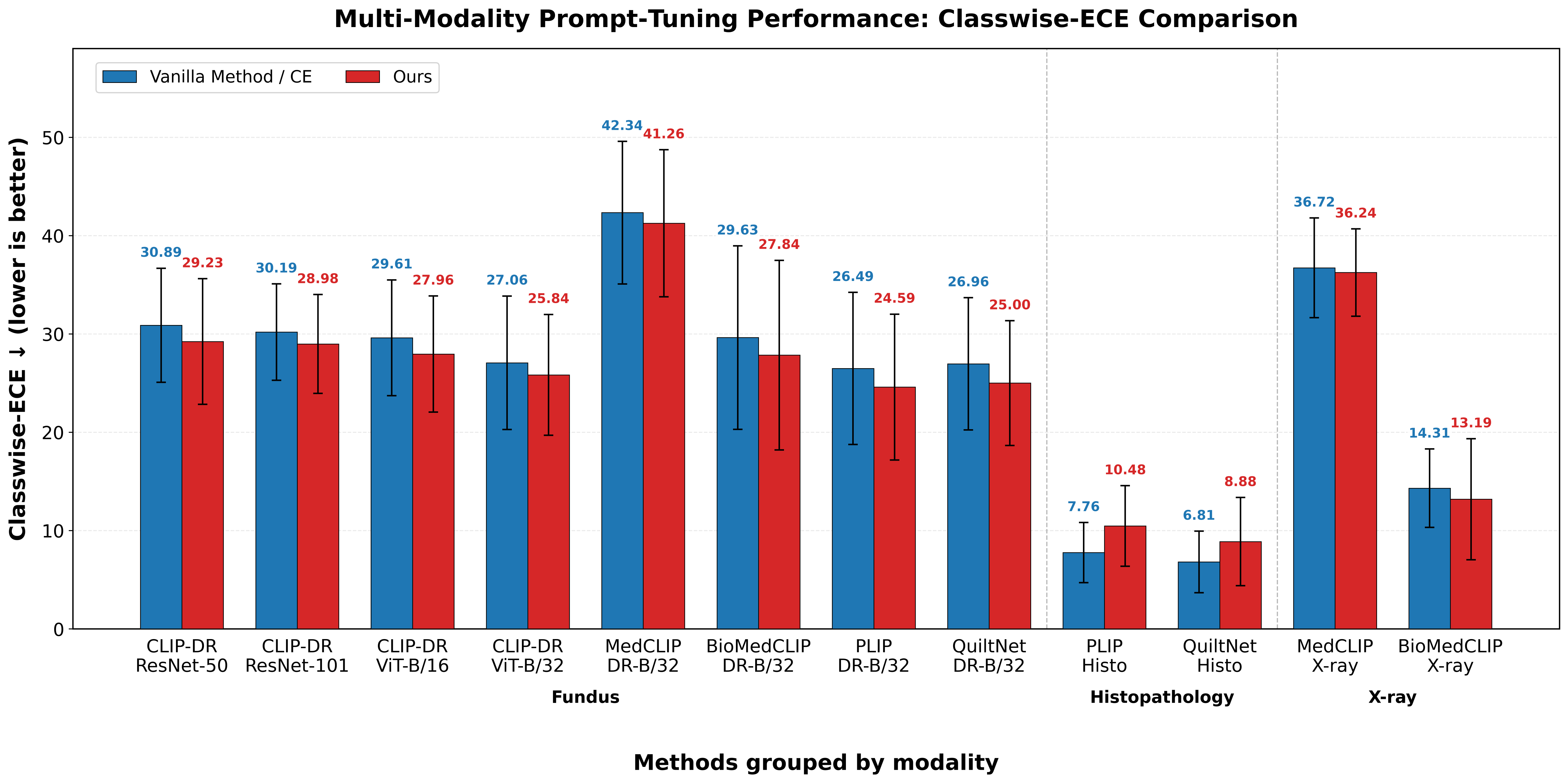}
\vspace{-10pt}
\caption{Weighted Subset ECE Analyses across Backbones and Modalities.} 
\label{fig:Weighted Subset ECE ana}
\vspace{-5pt}
\end{figure}

In Fig.\ref{fig:Weighted Subset ECE ana}, the WSECE analysis shows that integrating our loss function generally improves class-wise calibration compared with the vanilla CE baseline, especially in the fundus and X-ray modalities. In fundus datasets, class imbalance is high, for example, the "No DR" category represents up to 74\% of the EyePACS dataset. Our method consistently reduces WSECE across all backbones, including CLIP-DR-ResNet-50, CLIP-DR-ResNet-101, CLIP-DR-ViT-B/16, CLIP-DR-ViT-B/32, MedCLIP-DR-B/32, BioMedCLIP-DR-B/32, PLIP-DR-B/32, and QuiltNet-DR-B/32, indicating that the proposed loss helps align confidence more reliably across class-specific distributions. In X-ray, the improvement is smaller but still visible, particularly for BioMedCLIP-X-ray, where WSECE decreases from 14.31 to 13.19, while MedCLIP-X-ray shows only a slight reduction. However, in histopathology, our method increases WSECE for both PLIP-Histo and QuiltNet-Histo, suggesting that the proposed loss is less effective for class-wise calibration in this modality, possibly due to stronger visual heterogeneity or class-level ambiguity. Overall, the WSECE results support the broader calibration findings by showing that our loss improves class-wise calibration in most fundus and X-ray settings, while also highlighting that calibration gains are modality-dependent and should be evaluated separately across class-imbalanced medical datasets.

\section{Results}\label{sec:Results}

We evaluate three medical imaging modalities, diabetic retinopathy (APTOS~\cite{aptos2019}, EyePACS~\cite{eyepacs2015}, Messidor, Messidor-2~\cite{decenciere2014feedback}), histopathology (DigestPath~\cite{da2022digestpath}, Kather~\cite{kather2019predicting}, PaNuke~\cite{gamper2019pannuke}), and chest X-ray (COVIDX~\cite{rahman2021exploring}, RSNA18~\cite{Stein2018}) across 12 modality-specific configurations (CLIP-ResNet/ViT~\cite{radford2021clip}, MedCLIP~\cite{wang2022medclip}, BioMedCLIP~\cite{zhang2023biomedclip}, PLIP~\cite{huang2023plip}, QuiltNet~\cite{ikezogwo2023quilt1m}). We report accuracy and calibration with Expected Calibration Error (ECE), Maximum Calibration Error (MCE), and Adaptive Calibration Error (ACE). All results are averaged across three seeds. Complete results are provided in Tab.~\ref{tab:idallmet} and Tab.~\ref{tab:dsallmet}.
\subsection{In-domain (ID) performance}

\textbf{Diabetic retinopathy (DR)} MCM improves calibration in CLIP backbones:
\begin{itemize}
\item CLIP-DR-ResNet-50 (avg over 4 datasets): Acc. (\(67.87 \rightarrow 67.71\)), ECE \(4.40 \rightarrow 2.96\), MCE \(1.49 \rightarrow 1.11\), ACE \(5.00 \rightarrow 3.26\)
\item CLIP-DR-ResNet-101: Acc. \(68.29 \rightarrow 68.19\), ECE \(5.82 \rightarrow 4.82\), MCE \(2.04 \rightarrow 1.62\), ACE \(6.01 \rightarrow 4.90\)
\item CLIP-DR-ViT-B/16: Acc. \(69.18 \rightarrow 69.05\), ECE \(4.44 \rightarrow 3.43\), MCE \(1.37 \rightarrow 1.18\), ACE \(4.56 \rightarrow 3.77\)
\item CLIP-DR-ViT-B/32: Acc. \(68.13 \rightarrow 67.79\), ECE \(3.89 \rightarrow 2.77\), MCE \(1.35 \rightarrow 0.90\), ACE \(4.35 \rightarrow 2.89\)
\end{itemize}
Medical vision-language models show the same trend, e.g., MedCLIP-DR-B/32 ECE \(6.69 \rightarrow 4.94\) without compromising accuracy.

\textbf{Chest X-ray}
\begin{itemize}
\item MedCLIP-X-ray (avg over 2 datasets): Acc. \(65.77 \rightarrow 65.83\), ECE \(23.97 \rightarrow 20.03\), MCE/ACE also drop.
\item BioMedCLIP-X-ray: Acc. \(73.56 \rightarrow 73.70\), ECE \(7.78 \rightarrow 5.71\).
\end{itemize}

\textbf{Histopathology}
Calibration gains are not consistent:
\begin{itemize}
\item PLIP-Histo (avg over 3 datasets): Acc. (\(85.43 \rightarrow 85.30\)) but ECE increases (\(7.00 \rightarrow 9.92\)).
\item QuiltNet-Histo (avg): Acc. (\(86.73 \rightarrow 86.87\)) with higher ECE (\(6.21 \rightarrow 8.55\)).
\end{itemize}

\subsection{Domain-shift (out-domain) performance}

We compare domain-shifted (\texttt{DS}) variants.
\begin{itemize}
\item CLIP-DR-ViT-B/16-DS: Acc. \(32.90 \rightarrow 33.36\), ECE \(33.69 \rightarrow 20.08\)
\item CLIP-DR-ViT-B/32-DS: Acc. \(43.58 \rightarrow 46.59\), ECE \(15.83 \rightarrow 12.44\).
\item MedCLIP-DR-B/32-DS: Acc. \(52.69 \rightarrow 59.83\), ECE \(23.95 \rightarrow 23.54\).
\item BioMedCLIP-X-ray-DS: Acc. \(74.56 \rightarrow 78.84\), ECE \(9.00 \rightarrow 8.02\).
\end{itemize}
Histology backbones can degrade under shift:
\begin{itemize}
\item PLIP-DR-B/32-DS: Acc. \(55.40 \rightarrow 47.16\), ECE \(12.79 \rightarrow 13.56\).
\item QuiltNet-DR-B/32-DS: Acc. \(40.08 \rightarrow 27.92\), ECE \(15.42 \rightarrow 23.72\).
\end{itemize}


MCM provides consistent calibration gains on DR and CXR in ID and remains competitive in DS. 


\begin{table*}[t!]
\centering
\resizebox{0.96\textwidth}{!}{%
\begin{tabular}{@{}l|c|c|*{8}{d}D@{}}
\toprule
\textbf{Backbone} & \textbf{Dataset} & \textbf{Metric} &
\multicolumn{1}{c}{\rotatebox[origin=c]{45}{\parbox{15mm}{\centering \textbf{Base}}}} &
\multicolumn{1}{c}{\rotatebox[origin=c]{45}{\parbox{15mm}{\centering \textbf{MDCA}}}} &
\multicolumn{1}{c}{\rotatebox[origin=c]{45}{\parbox{15mm}{\centering \textbf{LS}}}} &
\multicolumn{1}{c}{\rotatebox[origin=c]{45}{\parbox{15mm}{\centering \textbf{MBLS}}}} &
\multicolumn{1}{c}{\rotatebox[origin=c]{45}{\parbox{15mm}{\centering \textbf{ECCV\_ ZS}}}} &
\multicolumn{1}{c}{\rotatebox[origin=c]{45}{\parbox{15mm}{\centering \textbf{ECCV\_ Penalty}}}} &
\multicolumn{1}{c}{\rotatebox[origin=c]{45}{\parbox{15mm}{\centering \textbf{TS}}}} & \multicolumn{1}{c}{\rotatebox[origin=c]{45}{\parbox{15mm}{\centering \textbf{zeroshot}}}} &
\multicolumn{1}{>{\columncolor{pink!20}}c@{}}{\rotatebox[origin=c]{45}{\parbox{15mm}{\centering \textbf{Ours (MCM)}}}} \\

\midrule

\multirow{20}{*}{CLIP-DR-ResNet-50}
& \multirow{4}{*}{APTOS}
& Acc. & 76.69 & 76.19 & 76.64 & 76.82 & 75.34 & 66.72 & 78.59 & 8.62 & 76.59 \\
& & ECE & 6.31 & 5.11 & 9.87 & 7.11 & 40.82 & 28.60 & 3.44 & 26.47 & 1.99 \\
& & MCE & 1.49 & 1.37 & 2.37 & 2.01 & 30.20 & 25.01 & 25.01 & 18.84 & 0.48 \\
& & ACE & 6.26 & 5.15 & 9.81 & 7.00 & 39.31 & 33.31 & 33.31 & 26.47 & 2.65 \\
\cline{2-12}
& \multirow{4}{*}{EyePACS} 
& Acc. & 74.48 & 74.06 & 74.02 & 74.12 & 74.40 & 73.34 & 74.57 & 7.39 & 74.09 \\
& & ECE & 2.16 & 2.06 & 3.58 & 2.00 & 12.89 & 38.52 & 7.22 & 20.41 & 3.60 \\
& & MCE & 1.66 & 0.79 & 1.99 & 0.67 & 3.06 & 26.81 & 26.81 & 13.70 & 1.84 \\
& & ACE & 3.24 & 2.39 & 3.84 & 2.22 & 13.08 & 38.52 & 38.52 & 20.41 & 3.60 \\
\cline{2-12}
& \multirow{4}{*}{Messidor} 
& Acc. & 56.12 & 54.44 & 57.19 & 57.06 & 75.34 & 44.70 & 55.69 & 20.58 & 56.20 \\
& & ECE & 4.88 & 4.43 & 5.06 & 4.48 & 39.31 & 14.65 & 6.42 & 18.83 & 4.06 \\
& & MCE & 1.66 & 1.68 & 1.97 & 1.53 & 13.62 & 13.96 & 13.96 & 9.80 & 1.52 \\
& & ACE & 5.79 & 5.79 & 5.36 & 4.97 & 24.93 & 14.70 & 14.70 & 18.83 & 4.37 \\
\cline{2-12}
& \multirow{4}{*}{Messidor-2} 
& Acc. & 64.20 & 63.45 & 63.95 & 64.07 & 63.28 & 58.31 & 63.97 & 13.46 & 63.95 \\
& & ECE & 4.26 & 3.84 & 5.97 & 3.06 & 20.00 & 29.68 & 6.05 & 14.76 & 2.19 \\
& & MCE & 1.13 & 1.17 & 1.71 & 0.86 & 6.73 & 25.42 & 25.42 & 9.77 & 0.58 \\
& & ACE & 4.69 & 4.10 & 6.24 & 3.66 & 20.00 & 29.68 & 29.68 & 14.76 & 2.41 \\
\cline{2-12}
& \multirow{4}{*}{Average} 
& Acc. & 67.8725 & 67.035 & 67.95 & 68.0175 & 72.09 & 60.7675 & 68.205 & 12.5125 & 67.7075 \\
& & ECE & 4.4025 & 3.86 & 6.12 & 4.1625 & 28.255 & 27.8625 & 5.7825 & 20.1175 & 2.96 \\
& & MCE & 1.485 & 1.2525 & 2.01 & 1.2675 & 13.4025 & 22.80 & 22.80 & 13.0275 & 1.105 \\
& & ACE & 4.995 & 4.3575 & 6.3125 & 4.4625 & 24.33 & 29.0525 & 29.0525 & 20.1175 & 3.2575 \\

\midrule

\multirow{20}{*}{CLIP-DR-ResNet-101}
& \multirow{4}{*}{APTOS}
& Acc. & 75.75 & 75.12 & 75.02 & 75.78 & 74.81 & 57.18 & 75.75 & 8.03 & 75.43 \\
& & ECE & 3.86 & 3.80 & 7.18 & 3.76 & 38.90 & 24.53 & 5.49 & 33.49 & 3.40 \\
& & MCE & 1.12 & 1.12 & 1.64 & 1.11 & 18.21 & 14.78 & 3.30 & 19.20 & 1.04 \\
& & ACE & 3.95 & 4.11 & 7.32 & 3.77 & 38.90 & 25.64 & 5.77 & 33.46 & 3.54 \\
\cline{2-12}
& \multirow{4}{*}{EyePACS} 
& Acc. & 73.92 & 73.93 & 73.88 & 73.95 & 74.14 & 73.47 & 73.93 & 2.59 & 73.95 \\
& & ECE & 2.89 & 3.08 & 4.81 & 2.98 & 9.15 & 34.81 & 7.96 & 32.94 & 2.92 \\
& & MCE & 0.84 & 0.93 & 2.83 & 0.87 & 1.98 & 2.85 & 2.85 & 22.17 & 0.86 \\
& & ACE & 3.10 & 3.14 & 5.15 & 3.06 & 9.16 & 7.86 & 7.86 & 32.94 & 2.92 \\
\cline{2-12}
& \multirow{4}{*}{Messidor} 
& Acc. & 59.47 & 57.67 & 59.31 & 59.50 & 58.21 & 45.50 & 59.47 & 20.67 & 59.08 \\
& & ECE & 10.67 & 9.09 & 10.90 & 10.67 & 29.36 & 16.91 & 6.87 & 11.31 & 8.97 \\
& & MCE & 4.10 & 3.87 & 4.14 & 4.04 & 19.00 & 16.49 & 2.02 & 11.30 & 3.35 \\
& & ACE & 10.95 & 9.03 & 10.85 & 10.92 & 29.36 & 16.99 & 7.58 & 11.31 & 9.13 \\
\cline{2-12}
& \multirow{4}{*}{Messidor-2} 
& Acc. & 64.03 & 63.68 & 64.45 & 64.11 & 63.59 & 58.31 & 64.01 & 2.01 & 64.28 \\
& & ECE & 5.86 & 6.58 & 8.84 & 7.01 & 24.95 & 26.53 & 6.31 & 37.70 & 4.00 \\
& & MCE & 2.11 & 1.87 & 2.91 & 2.37 & 11.04 & 21.58 & 1.69 & 18.16 & 1.24 \\
& & ACE & 6.03 & 6.68 & 9.01 & 7.25 & 24.95 & 26.53 & 7.76 & 37.70 & 4.01 \\
\cline{2-12}
& \multirow{4}{*}{Average} 
& Acc. & 68.2925 & 67.60 & 68.165 & 68.335 & 67.6875 & 58.615 & 68.29 & 8.325 & 68.185 \\
& & ECE & 5.82 & 5.6375 & 7.9325 & 6.105 & 25.59 & 25.695 & 6.6575 & 28.86 & 4.8225 \\
& & MCE & 2.0425 & 1.9475 & 2.88 & 2.0975 & 12.5575 & 13.925 & 2.465 & 17.7075 & 1.6225 \\
& & ACE & 6.0075 & 5.74 & 8.0825 & 6.25 & 25.5925 & 19.255 & 7.2425 & 28.8525 & 4.90 \\

\midrule

\multirow{20}{*}{CLIP-DR-ViT-B/16}
& \multirow{4}{*}{APTOS}
& Acc. & 77.90 & 77.45 & 77.84 & 77.94 & 76.14 & 56.78 & 77.90 & 7.95 & 77.50 \\
& & ECE & 3.99 & 4.87 & 7.88 & 4.09 & 38.96 & 24.87 & 4.44 & 29.78 & 1.85 \\
& & MCE & 1.46 & 1.46 & 1.81 & 1.38 & 12.54 & 13.95 & 1.86 & 17.20 & 0.61 \\
& & ACE & 3.99 & 4.84 & 7.90 & 4.08 & 38.96 & 26.23 & 4.74 & 29.78 & 2.03 \\
\cline{2-12}
& \multirow{4}{*}{EyePACS}
& Acc. & 74.64 & 74.65 & 74.53 & 74.64 & 74.85 & 72.71 & 74.64 & 3.21 & 74.53 \\
& & ECE & 2.25 & 2.31 & 4.36 & 2.25 & 19.06 & 41.84 & 7.53 & 24.38 & 3.25 \\
& & MCE & 0.68 & 0.78 & 2.23 & 0.72 & 5.45 & 21.84 & 3.24 & 19.22 & 1.08 \\
& & ACE & 2.30 & 2.29 & 4.29 & 2.33 & 18.80 & 41.84 & 7.47 & 24.38 & 3.25 \\
\cline{2-12}
& \multirow{4}{*}{Messidor}
& Acc. & 59.50 & 57.17 & 59.61 & 59.53 & 58.31 & 43.00 & 59.50 & 21.00 & 59.72 \\
& & ECE & 5.29 & 6.02 & 9.15 & 5.31 & 26.87 & 14.80 & 6.95 & 14.72 & 5.09 \\
& & MCE & 1.78 & 2.17 & 2.53 & 1.82 & 17.43 & 12.69 & 2.91 & 14.55 & 1.73 \\
& & ACE & 5.69 & 6.55 & 9.18 & 5.81 & 26.87 & 14.91 & 7.08 & 14.72 & 5.89 \\
\cline{2-12}
& \multirow{4}{*}{Messidor-2}
& Acc. & 64.66 & 64.51 & 64.96 & 64.62 & 64.95 & 58.26 & 64.66 & 2.18 & 64.43 \\
& & ECE & 6.24 & 6.10 & 8.24 & 6.13 & 21.25 & 32.13 & 7.36 & 28.41 & 3.53 \\
& & MCE & 1.54 & 2.05 & 2.31 & 1.55 & 9.51 & 31.87 & 2.51 & 14.18 & 1.28 \\
& & ACE & 6.24 & 6.33 & 8.37 & 6.26 & 21.17 & 31.94 & 7.46 & 28.41 & 3.91 \\
\cline{2-12}
& \multirow{4}{*}{Average}
& Acc. & 69.175 & 68.445 & 69.235 & 69.1825 & 68.5625 & 57.6875 & 69.175 & 8.585 & 69.045 \\
& & ECE & 4.4425 & 4.825 & 7.4075 & 4.445 & 26.535 & 28.41 & 6.57 & 24.3225 & 3.43 \\
& & MCE & 1.365 & 1.615 & 2.22 & 1.3675 & 11.2325 & 20.0875 & 2.63 & 16.2875 & 1.175 \\
& & ACE & 4.555 & 5.0025 & 7.435 & 4.62 & 26.45 & 28.73 & 6.6875 & 24.3225 & 3.77 \\

\bottomrule

\end{tabular}%
}
\end{table*}

\begin{table*}[t!]
\centering
\resizebox{\textwidth}{!}{%
\begin{tabular}{@{}l|c|c|*{8}{d}D@{}}
\toprule
\textbf{Backbone} & \textbf{Dataset} & \textbf{Metric} &
\multicolumn{1}{c}{\rotatebox[origin=c]{45}{\parbox{15mm}{\centering \textbf{Base}}}} &
\multicolumn{1}{c}{\rotatebox[origin=c]{45}{\parbox{15mm}{\centering \textbf{MDCA}}}} &
\multicolumn{1}{c}{\rotatebox[origin=c]{45}{\parbox{15mm}{\centering \textbf{LS}}}} &
\multicolumn{1}{c}{\rotatebox[origin=c]{45}{\parbox{15mm}{\centering \textbf{MBLS}}}} &
\multicolumn{1}{c}{\rotatebox[origin=c]{45}{\parbox{15mm}{\centering \textbf{ECCV\_ ZS}}}} &
\multicolumn{1}{c}{\rotatebox[origin=c]{45}{\parbox{15mm}{\centering \textbf{ECCV\_ Penalty}}}} &
\multicolumn{1}{c}{\rotatebox[origin=c]{45}{\parbox{15mm}{\centering \textbf{TS}}}} & \multicolumn{1}{c}{\rotatebox[origin=c]{45}{\parbox{15mm}{\centering \textbf{zeroshot}}}} &
\multicolumn{1}{>{\columncolor{pink!20}}c@{}}{\rotatebox[origin=c]{45}{\parbox{15mm}{\centering \textbf{Ours (MCM)}}}} \\

\midrule

\multirow{20}{*}{CLIP-DR-ViT-B/32}
& \multirow{4}{*}{APTOS}
& Acc. & 78.28 & 77.98 & 78.53 & 78.26 & 75.34 & 66.72 & 78.59 & 7.86 & 77.52 \\
& & ECE & 4.07 & 4.91 & 8.97 & 4.15 & 39.31 & 28.60 & 3.44 & 31.08 & 2.53 \\
& & MCE & 1.52 & 1.74 & 1.78 & 1.51 & 13.62 & 25.01 & 0.78 & 17.28 & 0.81 \\
& & ACE & 4.00 & 4.89 & 8.93 & 4.10 & 39.31 & 33.31 & 3.54 & 31.08 & 2.53 \\
\cline{2-12}
& \multirow{4}{*}{EyePACS} 
& Acc. & 74.58 & 74.49 & 74.52 & 74.55 & 74.40 & 73.34 & 74.57 & 8.86 & 74.54 \\
& & ECE & 3.29 & 3.05 & 6.24 & 3.06 & 12.89 & 38.52 & 7.22 & 23.22 & 2.39 \\
& & MCE & 1.66 & 1.55 & 3.26 & 1.52 & 3.06 & 26.81 & 2.99 & 18.52 & 0.72 \\
& & ACE & 3.24 & 2.97 & 6.22 & 3.04 & 13.08 & 38.52 & 7.20 & 24.91 & 2.41 \\
\cline{2-12}
& \multirow{4}{*}{Messidor} 
& Acc. & 55.47 & 53.22 & 55.45 & 55.39 & 56.17 & 44.70 & 55.69 & 20.58 & 55.47 \\
& & ECE & 3.11 & 3.82 & 5.01 & 3.07 & 24.93 & 14.65 & 6.42 & 21.65 & 3.58 \\
& & MCE & 0.91 & 1.61 & 1.62 & 0.93 & 13.88 & 13.96 & 2.32 & 18.74 & 1.20 \\
& & ACE & 5.15 & 4.56 & 6.21 & 5.04 & 24.93 & 14.70 & 7.11 & 21.65 & 3.96 \\
\cline{2-12}
& \multirow{4}{*}{Messidor-2} 
& Acc. & 64.20 & 63.90 & 63.51 & 64.10 & 63.28 & 58.31 & 63.97 & 7.63 & 63.61 \\
& & ECE & 5.09 & 6.11 & 8.04 & 5.06 & 20.00 & 29.68 & 6.05 & 27.48 & 2.57 \\
& & MCE & 1.31 & 1.53 & 1.81 & 1.39 & 6.73 & 25.42 & 2.13 & 17.94 & 0.85 \\
& & ACE & 4.99 & 6.09 & 8.03 & 4.90 & 20.00 & 29.68 & 6.05 & 30.38 & 2.66 \\
\cline{2-12}
& \multirow{4}{*}{Average} 
& Acc. & 68.1325 & 67.3975 & 68.0025 & 68.075 & 67.2975 & 60.7675 & 68.205 & 11.2325 & 67.785 \\
& & ECE & 3.89 & 4.4725 & 7.065 & 3.835 & 24.2825 & 27.8625 & 5.7825 & 25.8575 & 2.7675 \\
& & MCE & 1.35 & 1.6075 & 2.1175 & 1.3375 & 9.3225 & 22.80 & 2.055 & 18.12 & 0.895 \\
& & ACE & 4.345 & 4.6275 & 7.3475 & 4.27 & 24.33 & 29.0525 & 5.975 & 27.005 & 2.89 \\

\midrule

\multirow{20}{*}{MedCLIP-DR-B/32}
& \multirow{4}{*}{APTOS}
& Acc. & 62.27 & 62.14 & 63.16 & 63.96 & 62.27 & 49.43 & 62.27 & 11.25 & 61.73 \\
& & ECE & 10.04 & 9.99 & 14.14 & 10.68 & 9.98 & 21.19 & 7.48 & 8.82 & 7.64 \\
& & MCE & 3.27 & 3.16 & 4.46 & 3.22 & 3.16 & 16.74 & 4.29 & 8.82 & 2.8 \\
& & ACE & 10.04 & 10.00 & 14.13 & 10.68 & 9.98 & 22.19 & 6.77 & 8.82 & 7.626666667 \\
\cline{2-12}
& \multirow{4}{*}{EyePACS}
& Acc. & 73.66 & 73.66 & 73.66 & 73.66 & 73.65 & 73.66 & 73.66 & 3.54 & 73.66 \\
& & ECE & 2.84 & 2.09 & 7.65 & 2.89 & 3.40 & 20.87 & 7.85 & 16.56 & 3.17 \\
& & MCE & 1.36 & 1.12 & 4.10 & 1.36 & 1.85 & 14.62 & 4.74 & 16.56 & 3.46 \\
& & ACE & 3.54 & 3.47 & 7.64 & 3.66 & 3.67 & 20.87 & 7.84 & 16.00 & 5.183333333 \\
\cline{2-12}
& \multirow{4}{*}{Messidor}
& Acc. & 45.03 & 45.05 & 46.14 & 45.39 & 45.17 & 44.33 & 45.03 & 14.33 & 45.27 \\
& & ECE & 9.42 & 9.19 & 6.16 & 5.04 & 3.25 & 19.09 & 12.55 & 10.72 & 1.93 \\
& & MCE & 8.37 & 7.54 & 4.82 & 2.27 & 1.47 & 19.09 & 11.27 & 10.72 & 0.986666667 \\
& & ACE & 9.41 & 10.01 & 7.10 & 5.28 & 3.37 & 19.09 & 13.72 & 10.72 & 3.706666667 \\
\cline{2-12}
& \multirow{4}{*}{Messidor-2}
& Acc. & 58.31 & 58.31 & 58.31 & 58.31 & 58.31 & 57.78 & 58.31 & 8.94 & 58.31 \\
& & ECE & 4.45 & 6.96 & 10.34 & 9.75 & 5.90 & 22.96 & 12.41 & 11.13 & 7.03 \\
& & MCE & 2.61 & 3.82 & 6.41 & 3.73 & 3.17 & 12.28 & 7.77 & 11.13 & 3.776666667 \\
& & ACE & 5.62 & 7.60 & 11.31 & 9.81 & 6.98 & 23.00 & 12.41 & 11.13 & 9.03 \\
\cline{2-12}
& \multirow{4}{*}{Average}
& Acc. & 59.8167 & 59.7917 & 60.3167 & 60.3300 & 59.8500 & 56.3000 & 59.8167 & 9.5150 & 59.7433 \\
& & ECE & 6.6883 & 7.0575 & 9.5725 & 7.0917 & 5.6317 & 21.0275 & 10.0708 & 11.8075 & 4.9425 \\
& & MCE & 3.9017 & 3.9133 & 4.9467 & 2.6458 & 2.4142 & 15.6825 & 7.0183 & 11.8075 & 2.755833334 \\
& & ACE & 7.1508 & 7.7708 & 10.0442 & 7.3575 & 5.9992 & 21.2892 & 10.1850 & 11.6675 & 6.386666667 \\

\midrule

\multirow{12}{*}{MedCLIP-X-ray}
& \multirow{4}{*}{COVIDX}
& Acc. & 79.45 & 79.45666667 & 79.56333333 & 79.69333333 & 79.52666667 & 79.02 & 79.4 & 78.77 & 79.5 \\
& & ECE & 29.33666667 & 29.34333333 & 29.45 & 29.58 & 29.41333333 & 28.90666667 & 29.30333333 & 28.67 & 24.39 \\
& & MCE & 29.33666667 & 29.34333333 & 29.45 & 29.58 & 29.41333333 & 28.90666667 & 1.256666667 & 28.67 & 24.39 \\
& & ACE & 29.33666667 & 29.34333333 & 29.45 & 29.58 & 29.41333333 & 28.90666667 & 4.983333333 & 28.67 & 29.39 \\
\cline{2-12}
& \multirow{4}{*}{RSNA18}
& Acc. & 52.08 & 52.11 & 52.43333333 & 52.6 & 52.4 & 52.21 & 52.02 & 47.6 & 52.15 \\
& & ECE & 18.60333333 & 18.63333333 & 18.95333333 & 19.11333333 & 18.92 & 18.73333333 & 18.56333333 & 14.05 & 15.67333333 \\
& & MCE & 18.60333333 & 18.63333333 & 18.95333333 & 19.11333333 & 18.92 & 18.73333333 & 0.4933333333 & 14.05 & 15.67333333 \\
& & ACE & 18.60333333 & 18.63333333 & 18.95333333 & 19.11333333 & 18.92 & 18.76 & 1.383333333 & 16.94 & 18.67333333 \\
\cline{2-12}
& \multirow{4}{*}{Average}
& Acc. & 65.765 & 65.78333334 & 65.99833333 & 66.14666667 & 65.96333334 & 65.615 & 65.71 & 63.185 & 65.825 \\
& & ECE & 23.97 & 23.98833333 & 24.20166667 & 24.34666667 & 24.16666667 & 23.82 & 23.93333333 & 21.36 & 20.03166667 \\
& & MCE & 23.97 & 23.98833333 & 24.20166667 & 24.34666667 & 24.16666667 & 23.82 & 0.8750000002 & 21.36 & 20.03166667 \\
& & ACE & 23.97 & 23.98833333 & 24.20166667 & 24.34666667 & 24.16666667 & 23.82 & 3.183333333 & 22.805 & 24.03166667 \\

\bottomrule

\end{tabular}%
}
\end{table*}

\begin{table*}[t!]
\centering
\resizebox{\textwidth}{!}{%
\begin{tabular}{@{}l|c|c|*{8}{d}D@{}}
\toprule
\textbf{Backbone} & \textbf{Dataset} & \textbf{Metric} &
\multicolumn{1}{c}{\rotatebox[origin=c]{45}{\parbox{15mm}{\centering \textbf{Base}}}} &
\multicolumn{1}{c}{\rotatebox[origin=c]{45}{\parbox{15mm}{\centering \textbf{MDCA}}}} &
\multicolumn{1}{c}{\rotatebox[origin=c]{45}{\parbox{15mm}{\centering \textbf{LS}}}} &
\multicolumn{1}{c}{\rotatebox[origin=c]{45}{\parbox{15mm}{\centering \textbf{MBLS}}}} &
\multicolumn{1}{c}{\rotatebox[origin=c]{45}{\parbox{15mm}{\centering \textbf{ECCV\_ ZS}}}} &
\multicolumn{1}{c}{\rotatebox[origin=c]{45}{\parbox{15mm}{\centering \textbf{ECCV\_ Penalty}}}} &
\multicolumn{1}{c}{\rotatebox[origin=c]{45}{\parbox{15mm}{\centering \textbf{TS}}}} & \multicolumn{1}{c}{\rotatebox[origin=c]{45}{\parbox{15mm}{\centering \textbf{zeroshot}}}} &
\multicolumn{1}{>{\columncolor{pink!20}}c@{}}{\rotatebox[origin=c]{45}{\parbox{15mm}{\centering \textbf{Ours (MCM)}}}} \\

\midrule
\multirow{20}{*}{BioMedCLIP-DR-B/32}
& \multirow{4}{*}{APTOS}
& Acc. & 77.24666667 & 77.01666667 & 76.88 & 76.91666667 & 62.27 & 49.42666667 & 77.24666667 & 54.59 & 76.71 \\
& & ECE & 2.49 & 2.273333333 & 5.623333333 & 2.473333333 & 9.98 & 21.19 & 2.876666667 & 18.68 & 3.14 \\
& & MCE & 0.8033333333 & 0.67 & 1.326666667 & 0.7666666667 & 3.156666667 & 16.74333333 & 0.59 & 9.01 & 1.226666667 \\
& & ACE & 2.65 & 2.51 & 5.6 & 2.596666667 & 9.976666667 & 22.19333333 & 2.45 & 18.98 & 3.396666667 \\
\cline{2-12}
& \multirow{4}{*}{EyePACS}
& Acc. & 74.12666667 & 74.12666667 & 74.06666667 & 74.13 & 73.65333333 & 73.66 & 74.12666667 & 70.51 & 3.14 \\
& & ECE & 1.476666667 & 1.673333333 & 72.21 & 1.496666667 & 3.396666667 & 20.87 & 7.856666667 & 16.19 & 1.96 \\
& & MCE & 0.56 & 0.55 & 2.613333333 & 0.5633333333 & 1.853333333 & 14.62333333 & 2.806666667 & 9.31 & 0.013333333 \\
& & ACE & 1.826666667 & 1.85 & 3.616666667 & 1.826666667 & 3.67 & 20.87 & 7.73 & 16.16 & 2.01\\
\cline{2-12}
& \multirow{4}{*}{Messidor}
& Acc. & 55.53 & 54.97333333 & 55.30666667 & 55.41666667 & 45.16666667 & 44.33333333 & 55.53 & 45.67 & 55.36 \\
& & ECE & 5.126666667 & 4.246666667 & 5.98 & 4.616666667 & 3.246666667 & 19.08666667 & 9.04 & 8.9 & 2.92 \\
& & MCE & 1.733333333 & 1.38 & 1.826666667 & 1.566666667 & 1.473333333 & 19.08666667 & 3.446666667 & 3.43 & 1.073333333 \\
& & ACE & 5.356666667 & 5.116666667 & 6.73 & 5.376666667 & 3.366666667 & 19.09 & 8.836666667 & 9.22 & 2.97 \\
\cline{2-12}
& \multirow{4}{*}{Messidor-2}
& Acc. & 62.96 & 63.13 & 63.19 & 63.20666667 & 58.31 & 57.78 & 62.96 & 57.57 & 62.92 \\
& & ECE & 6.426666667 & 6.193333333 & 8.883333333 & 6.143333333 & 5.903333333 & 22.96333333 & 6.59 & 7.37 & 3.26 \\
& & MCE & 2.006666667 & 1.676666667 & 2.62 & 1.59 & 3.173333333 & 12.27666667 & 2.236666667 & 4.15 & 1.193333333 \\
& & ACE & 6.456666667 & 6.586666667 & 9.086666667 & 6.29 & 6.983333333 & 23.00333333 & 6.773333333 & 6.94 & 3.47 \\
\cline{2-12}
& \multirow{4}{*}{Average}
& Acc. & 67.46583333 & 67.31166667 & 67.36083333 & 67.4175 & 59.85 & 56.3 & 67.46583333 & 57.085 & 66.80 \\
& & ECE & 3.88 & 3.596666667 & 6.030833333 & 3.6825 & 5.631666667 & 21.0275 & 6.590833333 & 12.785 & 2.82 \\
& & MCE & 1.275833333 & 1.069166667 & 2.096666667 & 1.121666667 & 2.414166667 & 15.6825 & 2.27 & 6.475 & 0.8766666665 \\
& & ACE & 4.0725 & 4.015833333 & 6.258333333 & 4.0225 & 5.999166667 & 21.28916667 & 6.4475 & 12.825 & 2.96 \\

\midrule
\multirow{12}{*}{BioMedCLIP-X-ray}
& \multirow{4}{*}{COVIDX}
& Acc. & 83.32 & 78.27 & 82.82666667 & 83.38666667 & 82.27666667 & 81.45666667 & 83.35666667 & 84.37 & 83.63666667 \\
& & ECE & 6.966666667 & 8.84 & 11.67 & 8.646666667 & 6.72 & 8.953333333 & 6.99 & 10.7 & 3.496666667 \\
& & MCE & 2.42 & 3.85 & 4.56 & 3.34 & 2.51 & 3.6 & 0.9 & 8.88 & 1.24 \\
& & ACE & 6.97 & 8.93 & 11.69 & 8.64 & 6.72 & 9.03 & 7.09 & 10.07 & 3.66 \\
\cline{2-12}
& \multirow{4}{*}{RSNA18}
& Acc. & 63.80333333 & 63.97333333 & 63.36 & 63.58 & 63.60666667 & 61.67333333 & 63.81333333 & 49.71 & 63.75666667 \\
& & ECE & 8.596666667 & 7.313333333 & 4.78 & 7.386666667 & 8.15 & 6.723333333 & 8.656666667 & 29.49 & 7.923333333 \\
& & MCE & 2.98 & 2.56 & 1.44 & 2.38 & 2.39 & 2.33 & 3.3 & 16.67 & 2.44 \\
& & ACE & 8.69 & 7.93 & 4.93 & 7.17 & 8.14 & 7.01 & 9.04 & 29.49 & 7.98 \\
\cline{2-12}
& \multirow{4}{*}{Average}
& Acc. & 73.56166667 & 71.12166667 & 73.09333334 & 73.48333334 & 72.94166667 & 71.565 & 73.585 & 67.04 & 73.69666667 \\
& & ECE & 7.781666667 & 8.076666667 & 8.225 & 8.016666667 & 7.435 & 7.838333333 & 7.823333333 & 20.095 & 5.71 \\
& & MCE & 2.7 & 3.205 & 3 & 2.86 & 2.45 & 2.965 & 2.1 & 12.775 & 1.84 \\
& & ACE & 7.83 & 8.43 & 8.31 & 7.905 & 7.43 & 8.02 & 8.06 & 19.78 & 5.82 \\

\midrule
\multirow{20}{*}{PLIP-DR-B/32}
& \multirow{4}{*}{APTOS}
& Acc. & 78.66333333 & 78.76333333 & 78.03333333 & 78.67 & 77.51666667 & 74.78666667 & 78.66666667 & 78.77 & 78.45666667 \\
& & ECE & 4.983333333 & 5.26 & 12.19333333 & 5.14 & 3.603333333 & 20.99666667 & 1.933333333 & 11.12 & 0.9566666667 \\
& & MCE & 1.256666667 & 1.236666667 & 2.423333333 & 1.303333333 & 0.76 & 4.236666667 & 1.256666667 & 6.85 & 0.2366666667 \\
& & ACE & 4.983333333 & 5.236666667 & 12.18666667 & 5.053333333 & 3.523333333 & 20.99666667 & 4.983333333 & 10.71 & 0.99 \\
\cline{2-12}
& \multirow{4}{*}{EyePACS}
& Acc. & 74.43 & 74.44333333 & 74.28333333 & 74.48 & 74.33 & 74.01666667 & 74.43 & 57.41 & 74.45 \\
& & ECE & 1.443333333 & 1.16 & 7.393333333 & 1.45 & 1.306666667 & 19.37 & 7.563333333 & 29.22 & 3.703333333 \\
& & MCE & 0.4933333333 & 0.3866666667 & 3.376666667 & 0.52 & 0.3266666667 & 7.686666667 & 0.4933333333 & 21.04 & 1.206666667 \\
& & ACE & 1.383333333 & 1.146666667 & 7.35 & 1.426666667 & 1.316666667 & 19.34666667 & 1.383333333 & 29.22 & 3.7 \\
\cline{2-12}
& \multirow{4}{*}{Messidor}
& Acc. & 55.63666667 & 55.83333333 & 55.47333333 & 55.58333333 & 56.02666667 & 54.36 & 55.63666667 & 45.58 & 55.25 \\
& & ECE & 4.186666667 & 4.263333333 & 8.993333333 & 5.136666667 & 5.696666667 & 15.0 & 6.363333333 & 2.8 & 4.046666667 \\
& & MCE & 1.52 & 1.743333333 & 2.386666667 & 1.64 & 1.666666667 & 7.74 & 1.52 & 2.62 & 1.05 \\
& & ACE & 5.113333333 & 5.41 & 9.033333333 & 5.45 & 5.923333333 & 14.86 & 5.113333333 & 4.39 & 5.476666667	\\
\cline{2-12}
& \multirow{4}{*}{Messidor-2}
& Acc. & 63.45666667 & 63.42 & 63.74333333 & 63.64333333 & 63.28333333 & 62.27333333 & 63.45666667 & 47.25 & 63.70333333 \\
& & ECE & 2.92 & 3.67 & 8.993333333 & 4.31 & 3.52 & 18.00666667 & 6.236666667 & 14.69 & 2.276666667 \\
& & MCE & 1.01 & 1.353333333 & 2.14 & 1.216666667 & 0.93 & 6.066666667 & 1.01 & 8.61 & 0.9733333333 \\
& & ACE & 2.91 & 3.946666667 & 9.033333333 & 4.78 & 3.66 & 18.00666667 & 2.91 & 14.69 & 3.043333333 \\
\cline{2-12}
& \multirow{4}{*}{Average}
& Acc. & 68.04666667 & 68.115 & 67.88333333 & 68.09416667 & 67.78916667 & 66.35916667 & 68.0475 & 57.2525 & 67.965 \\
& & ECE & 3.383333333 & 3.588333333 & 9.393333332 & 4.009166667 & 3.531666667 & 18.34333334 & 5.524166667 & 14.4575 & 2.745833333 \\
& & MCE & 1.07 & 1.18 & 2.581666667 & 1.17 & 0.9208333334 & 6.4325 & 1.07 & 9.78 & 0.8666666668 \\
& & ACE & 3.5975 & 3.935 & 9.400833334 & 4.1775 & 3.605833333 & 18.3025 & 3.5975 & 14.7525 & 3.3025 \\

\bottomrule

\end{tabular}%
}
\end{table*}

\begin{table*}[t!]
\centering
\resizebox{\textwidth}{!}{%
\begin{tabular}{@{}l|c|c|*{8}{d}D@{}}
\toprule
\textbf{Backbone} & \textbf{Dataset} & \textbf{Metric} &
\multicolumn{1}{c}{\rotatebox[origin=c]{45}{\parbox{15mm}{\centering \textbf{Base}}}} &
\multicolumn{1}{c}{\rotatebox[origin=c]{45}{\parbox{15mm}{\centering \textbf{MDCA}}}} &
\multicolumn{1}{c}{\rotatebox[origin=c]{45}{\parbox{15mm}{\centering \textbf{LS}}}} &
\multicolumn{1}{c}{\rotatebox[origin=c]{45}{\parbox{15mm}{\centering \textbf{MBLS}}}} &
\multicolumn{1}{c}{\rotatebox[origin=c]{45}{\parbox{15mm}{\centering \textbf{ECCV\_ ZS}}}} &
\multicolumn{1}{c}{\rotatebox[origin=c]{45}{\parbox{15mm}{\centering \textbf{ECCV\_ Penalty}}}} &
\multicolumn{1}{c}{\rotatebox[origin=c]{45}{\parbox{15mm}{\centering \textbf{TS}}}} & \multicolumn{1}{c}{\rotatebox[origin=c]{45}{\parbox{15mm}{\centering \textbf{zeroshot}}}} &
\multicolumn{1}{>{\columncolor{pink!20}}c@{}}{\rotatebox[origin=c]{45}{\parbox{15mm}{\centering \textbf{Ours (MCM)}}}} \\

\midrule
\multirow{16}{*}{PLIP-Histo.}
& \multirow{4}{*}{DigestPath}
& Acc. & 90.73666667 & 91.12 & 89.29666667 & 91.08666667 & 91.47333333 & 83.32333333 & 92.33333333 & 80.69 & 90.86333333 \\
& & ECE & 4.923333333 & 3.62 & 4.073333333 & 4.616666667 & 4.433333333 & 14.66666667 & 3.866666667 & 6.23 & 6.796666667 \\
& & MCE & 1.256666667 & 1.236666667 & 2.423333333 & 1.303333333 & 0.76 & 4.236666667 & 1.256666667 & 1.83 & 5.72 \\
& & ACE & 4.983333333 & 5.236666667 & 12.18666667 & 5.053333333 & 3.523333333 & 20.99666667 & 4.983333333 & 6.23 & 6.79 \\
\cline{2-12}
& \multirow{4}{*}{Kather}
& Acc. & 86.99666667 & 86.51333333 & 86.71 & 85.90333333 & 87.32666667 & 87.69 & 87.20666667 & 57.74 & 87.24 \\
& & ECE & 4.106666667 & 5.03 & 7.86 & 3.76 & 2.91 & 5.563333333 & 4.056666667 & 16.41 & 7.56 \\
& & MCE & 0.4933333333 & 0.3866666667 & 3.376666667 & 0.52 & 0.3266666667 & 7.686666667 & 0.4933333333 & 5.34 & 5.31 \\
& & ACE & 1.383333333 & 1.146666667 & 7.35 & 1.426666667 & 1.316666667 & 19.34666667 & 1.383333333 & 16.38 & 7.54 \\
\cline{2-12}
& \multirow{4}{*}{PaNuke}
& Acc. & 78.56666667 & 77.84333333 & 78.20666667 & 78.35 & 79.99666667 & 67.86333333 & 79.43 & 56.74 & 77.78666667 \\
& & ECE & 11.97666667 & 9.753333333 & 2.95 & 10.18333333 & 10.90666667 & 4.826666667 & 11.21 & 18.82 & 15.39 \\
& & MCE & 1.52 & 1.743333333 & 2.386666667 & 1.64 & 1.666666667 & 7.74 & 1.52 & 6.66 & 11.95 \\
& & ACE & 5.113333333 & 5.41 & 9.033333333 & 5.45 & 5.923333333 & 14.86 & 5.113333333 & 19.64 & 15.39 \\
\cline{2-12}
& \multirow{4}{*}{Average}
& Acc. & 85.43333334 & 85.15888889 & 84.73777778 & 85.11333333 & 86.26555556 & 79.62555555 & 86.32333333 & 65.05666667 & 85.29666667 \\
& & ECE & 7.002222223 & 6.134444444 & 5.685833333 & 6.186666666 & 6.083333334 & 8.352222223 & 6.377777778 & 13.82 & 9.915555556 \\
& & MCE & 1.09 & 1.122222222 & 2.890833334 & 1.154444444 & 0.9177777779 & 6.554444445 & 1.09 & 4.61 & 7.66 \\
& & ACE & 3.826666666 & 3.931111111 & 8.980000001 & 3.976666667 & 3.587777778 & 18.40111111 & 3.826666666 & 14.08333333 & 9.906666667 \\

\midrule
\multirow{20}{*}{QuiltNet-DR-B/32}
& \multirow{4}{*}{APTOS}
& Acc. & 76.97 & 76.12333333 & 76.99666667 & 76.72 & 77.51666667 & 73.04 & 76.97 & 10.03 & 75.81333333 \\
& & ECE & 2.753333333 & 2.08 & 4.426666667 & 2.226666667 & 3.603333333 & 16.75 & 3.723333333 & 23.63666667 & 2.713333333 \\
& & MCE & 1.256666667 & 1.236666667 & 2.423333333 & 1.303333333 & 0.76 & 4.236666667 & 1.256666667 & 1.256666667 & 0.8366666667 \\
& & ACE & 4.983333333 & 5.236666667 & 12.18666667 & 5.053333333 & 3.523333333 & 20.99666667 & 4.983333333 & 4.983333333 & 2.746666667 \\
\cline{2-12}
& \multirow{4}{*}{EyePACS}
& Acc. & 74.43 & 74.40333333 & 74.37666667 & 74.32666667 & 74.33 & 73.9 & 74.43 & 22.22 & 74.36666667 \\
& & ECE & 3.146666667 & 3.406666667 & 4.733333333 & 3.456666667 & 1.306666667 & 19.91666667 & 7.516666667 & 11.37 & 1.21 \\
& & MCE & 0.4933333333 & 0.3866666667 & 3.376666667 & 0.52 & 0.3266666667 & 7.686666667 & 0.4933333333 & 0.4933333333 & 0.3866666667 \\
& & ACE & 1.383333333 & 1.146666667 & 7.35 & 1.426666667 & 1.316666667 & 19.34666667 & 1.383333333 & 1.383333333 & 1.25 \\
\cline{2-12}
& \multirow{4}{*}{Messidor}
& Acc. & 52.30666667 & 53.0 & 53.36 & 51.58666667 & 56.02666667 & 51.19666667 & 52.30666667 & 12.58 & 53.08333333 \\
& & ECE & 4.006666667 & 5.763333333 & 4.66 & 4.366666667 & 5.696666667 & 13.31 & 6.206666667 & 26.7 & 2.556666667 \\
& & MCE & 1.52 & 1.743333333 & 2.386666667 & 1.64 & 1.666666667 & 7.74 & 1.52 & 1.52 & 0.9733333333 \\
& & ACE & 5.113333333 & 5.41 & 9.033333333 & 5.45 & 5.923333333 & 14.86 & 5.113333333 & 5.113333333 & 3.89 \\
\cline{2-12}
& \multirow{4}{*}{Messidor-2}
& Acc. & 62.44333333 & 62.44 & 62.9 & 63.22666667 & 63.28333333 & 61.75333333 & 62.44333333 & 24.03 & 62.71333333 \\
& & ECE & 3.893333333 & 3.88 & 5.126666667 & 3.79 & 3.52 & 15.54666667 & 8.726666667 & 7.85 & 3.103333333 \\
& & MCE & 1.01 & 1.353333333 & 2.14 & 1.216666667 & 0.93 & 6.066666667 & 1.01 & 1.01 & 1.373333333 \\
& & ACE & 2.91 & 3.946666667 & 9.033333333 & 4.78 & 3.66 & 18.00666667 & 2.91 & 2.91 & 3.57 \\
\cline{2-12}
& \multirow{4}{*}{Average}
& Acc. & 66.5375 & 66.49166667 & 66.90833334 & 66.465 & 67.78916667 & 64.9725 & 66.5375 & 17.215 & 66.49416667 \\
& & ECE & 3.45 & 3.7825 & 4.736666667 & 3.46 & 3.531666667 & 16.38083334 & 6.543333334 & 17.38916667 & 2.395833333 \\
& & MCE & 1.07 & 1.18 & 2.581666667 & 1.17 & 0.9208333334 & 6.4325 & 1.07 & 1.07 & 0.8924999999 \\
& & ACE & 3.5975 & 3.935 & 9.400833334 & 4.1775 & 3.605833333 & 18.3025 & 3.5975 & 3.5975 & 2.864166667 \\

\midrule
\multirow{16}{*}{QuiltNet-Histo.}
& \multirow{4}{*}{DigestPath}
& Acc. & 89.76333333 & 90.07 & 89.68333333 & 89.8 & 90.39666667 & 83.70666667 & 89.22333333 & 53.43 & 89.62 \\
& & ECE & 5.27 & 4.086666667 & 2.416666667 & 5.353333333 & 4.416666667 & 13.58 & 5.313333333 & 19.85 & 7.81 \\
& & MCE & 1.256666667 & 1.236666667 & 2.423333333 & 1.303333333 & 0.76 & 4.236666667 & 1.256666667 & 7.31 & 6.57 \\
& & ACE & 4.983333333 & 5.236666667 & 12.18666667 & 5.053333333 & 3.523333333 & 20.99666667 & 4.983333333 & 20.23 & 7.81 \\
\cline{2-12}
& \multirow{4}{*}{Kather}
& Acc. & 91.55666667 & 90.75333333 & 90.61333333 & 90.49 & 93.23333333 & 91.4 & 91.84 & 60.32 & 91.62666667 \\
& & ECE & 1.51 & 1.913333333 & 2.286666667 & 1.74 & 1.406666667 & 10.95 & 1.296666667 & 4.24 & 4.383333333 \\
& & MCE & 0.4933333333 & 0.3866666667 & 3.376666667 & 0.52 & 0.3266666667 & 7.686666667 & 0.4933333333 & 1.2 & 2.873333333 \\
& & ACE & 1.383333333 & 1.146666667 & 7.35 & 1.426666667 & 1.316666667 & 19.34666667 & 1.383333333 & 3.84 & 4.36 \\
\cline{2-12}
& \multirow{4}{*}{PaNuke}
& Acc. & 78.85666667 & 78.46 & 79.18666667 & 78.98 & 79.98666667 & 72.62 & 78.11 & 55.43 & 79.37 \\
& & ECE & 11.86333333 & 10.44 & 7.98 & 11.04 & 10.42666667 & 8.946666667 & 11.74333333 & 24.15 & 13.45333333 \\
& & MCE & 1.52 & 1.743333333 & 2.386666667 & 1.64 & 1.666666667 & 7.74 & 1.52 & 8.12 & 10.13 \\
& & ACE & 5.113333333 & 5.41 & 9.033333333 & 5.45 & 5.923333333 & 14.86 & 5.113333333 & 24.15 & 13.41666667 \\
\cline{2-12}
& \multirow{4}{*}{Average}
& Acc. & 86.72555556 & 86.42777778 & 86.49444444 & 86.42333333 & 87.87222222 & 82.57555556 & 86.39111111 & 56.39333333 & 86.87222222 \\
& & ECE & 6.214444443 & 5.48 & 4.227777778 & 6.044444444 & 5.416666668 & 11.15888889 & 6.117777778 & 16.08 & 8.548888888 \\
& & MCE & 1.09 & 1.122222222 & 2.728888889 & 1.154444444 & 0.9177777779 & 6.554444445 & 1.09 & 5.543333333 & 6.524444444 \\
& & ACE & 3.826666666 & 3.931111111 & 9.523333334 & 3.976666667 & 3.587777778 & 18.40111111 & 3.826666666 & 16.07333333 & 8.52888889 \\

\bottomrule

\end{tabular}%
}
\caption{Accuracy and ECE of each calibration method in in-domain (ID) across 12 modality-specific configurations and 7 calibration methods covering three modalities: Diabetic retinopathy grading, histopathology, and chest-X-ray.}
\label{tab:idallmet}
\end{table*}

\begin{table*}[t!]
\centering
\resizebox{\textwidth}{!}{%
\begin{tabular}{@{}l|c|c|*{7}{d}D@{}}
\toprule
\textbf{Backbone} & \textbf{Dataset} & \textbf{Metric} &
\multicolumn{1}{c}{\rotatebox[origin=c]{45}{\parbox{15mm}{\centering \textbf{Base}}}} &
\multicolumn{1}{c}{\rotatebox[origin=c]{45}{\parbox{15mm}{\centering \textbf{MDCA}}}} &
\multicolumn{1}{c}{\rotatebox[origin=c]{45}{\parbox{15mm}{\centering \textbf{LS}}}} &
\multicolumn{1}{c}{\rotatebox[origin=c]{45}{\parbox{15mm}{\centering \textbf{MBLS}}}} &
\multicolumn{1}{c}{\rotatebox[origin=c]{45}{\parbox{15mm}{\centering \textbf{ECCV\_ ZS}}}} &
\multicolumn{1}{c}{\rotatebox[origin=c]{45}{\parbox{15mm}{\centering \textbf{ECCV\_ Penalty}}}} &
\multicolumn{1}{c}{\rotatebox[origin=c]{45}{\parbox{15mm}{\centering \textbf{TS}}}} & 
\multicolumn{1}{>{\columncolor{pink!20}}c@{}}{\rotatebox[origin=c]{45}{\parbox{15mm}{\centering \textbf{Ours (MCM)}}}} \\

\midrule
\multirow{16}{*}{CLIP-DR-ResNet-50-DS}
& \multirow{4}{*}{APTOS}
& Acc. & 43.75 & 48.01 & 50.54 & 43.80 & 20.71 & 43.83 & 44.10 & 43.68 \\
& & ECE & 24.13 & 15.37 & 19.67 & 24.25 & 12.22 & 17.57 & 20.38 & 17.77 \\
& & MCE & 6.07 & 4.58 & 8.21 & 6.05 & 7.43 & 17.26 & 6.88 & 7.23 \\
& & ACE & 24.01 & 15.33 & 19.74 & 24.03 & 13.02 & 18.49 & 20.46 & 17.69 \\
\cline{2-11}
& \multirow{4}{*}{EyePACS}
& Acc. & 66.33 & 66.48 & 69.84 & 66.34 & 44.33 & 55.79 & 68.16 & 65.52 \\
& & ECE & 10.94 & 13.97 & 10.83 & 10.95 & 25.52 & 31.43 & 10.07 & 7.49 \\
& & MCE & 2.99 & 4.19 & 3.26 & 2.99 & 19.14 & 31.29 & 2.50 & 2.20 \\
& & ACE & 10.71 & 13.68 & 10.89 & 10.70 & 25.37 & 31.89 & 10.01 & 7.50 \\
\cline{2-11}
& \multirow{4}{*}{Messidor-2}
& Acc. & 46.79 & 50.29 & 52.96 & 46.69 & 18.90 & 48.95 & 49.87 & 44.72 \\
& & ECE & 10.15 & 10.18 & 8.41 & 9.98 & 18.57 & 23.22 & 9.52 & 7.89 \\
& & MCE & 3.82 & 3.71 & 3.21 & 3.81 & 12.57 & 23.01 & 3.13 & 2.85 \\
& & ACE & 10.17 & 9.32 & 8.03 & 10.03 & 18.48 & 23.22 & 9.89 & 8.62 \\
\cline{2-11}
& \multirow{4}{*}{Average}
& Acc. & 52.29 & 54.92666667 & 57.78 & 52.27666667 & 27.98 & 49.52333333 & 54.04333333 & 51.30666667 \\
& & ECE & 15.07333333 & 13.17333333 & 12.97 & 15.06 & 18.77 & 24.07333333 & 13.32333333 & 10.26 \\
& & MCE & 4.29333333 & 4.16 & 4.89 & 4.28333333 & 13.04666667 & 23.85333333 & 4.17 & 4.09333333 \\
& & ACE & 14.96333333 & 12.77666667 & 10.89 & 14.92 & 18.95666667 & 24.53333333 & 13.45333333 & 10.6075 \\

\midrule
\multirow{16}{*}{CLIP-DR-ResNet-101-DS}
& \multirow{4}{*}{APTOS}
& Acc. & 35.38 & 49.90 & 36.88 & 35.45 & 51.13 & 18.08 & 34.20 & 41.95 \\
& & ECE & 21.38 & 27.42 & 17.84 & 21.24 & 26.89 & 15.56 & 32.71 & 19.32 \\
& & MCE & 6.25 & 7.12 & 4.96 & 6.32 & 23.43 & 11.09 & 10.77 & 4.86 \\
& & ACE & 21.48 & 28.20 & 17.92 & 21.55 & 27.81 & 16.61 & 32.66 & 19.51 \\
\cline{2-11}
& \multirow{4}{*}{EyePACS}
& Acc. & 50.16 & 69.27 & 64.66 & 50.25 & 69.71 & 18.47 & 35.70 & 61.60 \\
& & ECE & 16.53 & 23.59 & 15.13 & 16.60 & 43.77 & 11.73 & 26.14 & 13.52 \\
& & MCE & 7.57 & 8.83 & 5.26 & 7.59 & 37.90 & 11.25 & 13.28 & 4.84 \\
& & ACE & 15.98 & 22.81 & 15.35 & 16.07 & 43.77 & 18.68 & 26.02 & 13.17 \\
\cline{2-11}
& \multirow{4}{*}{Messidor-2}
& Acc. & 40.20 & 57.72 & 46.46 & 40.27 & 58.05 & 9.12 & 35.11 & 51.64 \\
& & ECE & 13.07 & 17.10 & 12.52 & 13.32 & 33.72 & 16.72 & 22.07 & 12.54 \\
& & MCE & 5.98 & 7.39 & 3.81 & 5.90 & 32.64 & 14.89 & 11.87 & 4.96 \\
& & ACE & 12.60 & 16.04 & 12.98 & 12.36 & 33.72 & 22.46 & 21.61 & 12.22 \\
\cline{2-11}
& \multirow{4}{*}{Average}
& Acc. & 41.91333333 & 58.96333333 & 49.33333333 & 41.99 & 59.63 & 15.22333333 & 35.00333333 & 51.73 \\
& & ECE & 16.99333333 & 22.70333333 & 15.16333333 & 17.05333333 & 34.79333333 & 14.67 & 26.97333333 & 15.12 \\
& & MCE & 6.60 & 7.78 & 4.67 & 6.60333333 & 31.32333333 & 12.41 & 11.97333333 & 4.88666667 \\
& & ACE & 16.68666667 & 22.35 & 15.41 & 16.66 & 35.10 & 19.25 & 26.76333333 & 14.96666667 \\

\midrule
\multirow{16}{*}{CLIP-DR-ViT-B/16-DS}
& \multirow{4}{*}{APTOS}
& Acc. & 28.69 & 34.50 & 26.82 & 28.82 & 31.30 & 47.17 & 30.16 & 29.74 \\
& & ECE & 41.70 & 22.61 & 30.89 & 41.80 & 8.17 & 23.21 & 29.15 & 30.48 \\
& & MCE & 12.65 & 7.69 & 8.20 & 12.65 & 7.21 & 22.98 & 7.32 & 6.28 \\
& & ACE & 41.56 & 22.96 & 31.06 & 41.64 & 11.50 & 23.36 & 29.98 & 30.48 \\
\cline{2-11}
& \multirow{4}{*}{EyePACS}
& Acc. & 40.95 & 45.32 & 45.77 & 41.02 & 53.78 & 68.11 & 26.71 & 41.91 \\
& & ECE & 32.71 & 15.60 & 14.94 & 32.77 & 25.69 & 43.89 & 22.74 & 12.89 \\
& & MCE & 12.93 & 6.28 & 7.22 & 12.94 & 17.74 & 43.84 & 11.20 & 6.79 \\
& & ACE & 32.49 & 15.25 & 15.11 & 32.55 & 27.30 & 43.89 & 23.11 & 12.72 \\
\cline{2-11}
& \multirow{4}{*}{Messidor-2}
& Acc. & 29.07 & 34.81 & 29.80 & 29.07 & 30.47 & 57.72 & 25.21 & 28.42 \\
& & ECE & 26.65 & 15.61 & 19.79 & 26.59 & 13.53 & 32.77 & 19.18 & 16.87 \\
& & MCE & 11.87 & 8.64 & 12.51 & 11.81 & 12.08 & 32.30 & 9.99 & 9.22 \\
& & ACE & 27.23 & 15.91 & 18.80 & 27.10 & 16.05 & 32.77 & 19.71 & 16.96 \\
\cline{2-11}
& \multirow{4}{*}{Average}
& Acc. & 32.90333333 & 38.21 & 34.13 & 32.97 & 38.51666667 & 57.66666667 & 27.36 & 33.35666667 \\
& & ECE & 33.68666667 & 17.94 & 21.87333333 & 33.72 & 15.79666667 & 33.29 & 23.69 & 20.08 \\
& & MCE & 12.48333333 & 7.53666667 & 9.31 & 12.46666667 & 12.34333333 & 33.04 & 9.50333333 & 7.43 \\
& & ACE & 33.76 & 18.04 & 21.65666667 & 33.76333333 & 18.28333333 & 33.34 & 24.26666667 & 20.05333333 \\

\bottomrule

\end{tabular}%
}
\end{table*}

\begin{table*}[t!]
\centering
\resizebox{\textwidth}{!}{%
\begin{tabular}{@{}l|c|c|*{7}{d}D@{}}
\toprule
\textbf{Backbone} & \textbf{Dataset} & \textbf{Metric} &
\multicolumn{1}{c}{\rotatebox[origin=c]{45}{\parbox{15mm}{\centering \textbf{Base}}}} &
\multicolumn{1}{c}{\rotatebox[origin=c]{45}{\parbox{15mm}{\centering \textbf{MDCA}}}} &
\multicolumn{1}{c}{\rotatebox[origin=c]{45}{\parbox{15mm}{\centering \textbf{LS}}}} &
\multicolumn{1}{c}{\rotatebox[origin=c]{45}{\parbox{15mm}{\centering \textbf{MBLS}}}} &
\multicolumn{1}{c}{\rotatebox[origin=c]{45}{\parbox{15mm}{\centering \textbf{ECCV\_ ZS}}}} &
\multicolumn{1}{c}{\rotatebox[origin=c]{45}{\parbox{15mm}{\centering \textbf{ECCV\_ Penalty}}}} &
\multicolumn{1}{c}{\rotatebox[origin=c]{45}{\parbox{15mm}{\centering \textbf{TS}}}} &
\multicolumn{1}{>{\columncolor{pink!20}}c@{}}{\rotatebox[origin=c]{45}{\parbox{15mm}{\centering \textbf{Ours (MCM)}}}} \\

\midrule
\multirow{16}{*}{CLIP-DR-ViT-B/32-DS}
& \multirow{4}{*}{APTOS}
& Acc. & 40.88 & 44.18 & 28.73 & 40.88 & 26.70 & 25.76 & 36.44 & 41.29 \\
& & ECE & 23.27 & 22.18 & 16.39 & 23.27 & 17.19 & 17.13 & 20.60 & 23.10 \\
& & MCE & 5.59 & 5.42 & 7.72 & 5.59 & 7.73 & 8.38 & 5.03 & 6.72 \\
& & ACE & 23.12 & 22.64 & 16.42 & 23.12 & 16.73 & 21.06 & 20.60 & 23.01 \\
\cline{2-11}
& \multirow{4}{*}{EyePACS}
& Acc. & 51.50 & 58.45 & 49.68 & 51.50 & 37.91 & 22.99 & 52.39 & 59.97 \\
& & ECE & 11.67 & 15.79 & 8.15 & 11.67 & 11.95 & 20.01 & 9.98 & 8.37 \\
& & MCE & 4.40 & 7.86 & 3.82 & 4.40 & 5.29 & 10.52 & 2.99 & 2.52 \\
& & ACE & 11.38 & 15.56 & 8.17 & 11.38 & 11.75 & 20.90 & 10.10 & 8.66 \\
\cline{2-11}
& \multirow{4}{*}{Messidor-2}
& Acc. & 38.36 & 38.28 & 26.76 & 38.36 & 22.48 & 27.93 & 35.51 & 38.50 \\
& & ECE & 12.55 & 7.75 & 14.19 & 12.55 & 14.46 & 21.29 & 15.24 & 5.84 \\
& & MCE & 3.17 & 2.25 & 5.62 & 3.17 & 6.20 & 7.08 & 6.23 & 0.59 \\
& & ACE & 12.65 & 8.82 & 15.37 & 12.65 & 14.51 & 21.37 & 15.61 & 6.42 \\
\cline{2-11}
& \multirow{4}{*}{Average}
& Acc. & 43.58 & 46.97 & 35.05666667 & 43.58 & 29.03 & 25.56 & 41.44666667 & 46.58666667 \\
& & ECE & 15.83 & 15.24 & 12.91 & 15.83 & 14.53333333 & 19.47666667 & 15.27333333 & 12.43666667 \\
& & MCE & 4.38666667 & 5.17666667 & 5.72 & 4.38666667 & 6.40666667 & 8.66 & 4.75 & 3.27666667 \\
& & ACE & 15.71666667 & 15.67333333 & 13.32 & 15.71666667 & 14.33 & 21.11 & 15.43666667 & 12.69666667 \\

\midrule
\multirow{20}{*}{MedCLIP-DR-B/32-DS}
& \multirow{4}{*}{APTOS}
& Acc. & 47.79 & 40.03333333 & 48.56333333 & 49.10 & 48.91 & 34.97333333 & 49.05333333 & 48.75 \\
& & ECE & 18.92333333 & 8.21 & 15.99 & 14.13 & 14.17333333 & 16.08333333 & 18.06666667 & 12.14 \\
& & MCE & 16.60 & 6.35 & 10.29 & 12.22 & 10.50333333 & 16.08333333 & 9.33666667 & 8.416666667 \\ 
& & ACE & 18.90 & 11.88666667 & 16.12 & 14.68 & 14.37 & 17.71666667 & 18.12 & 12.57 \\
\cline{2-11}
& \multirow{4}{*}{EyePACS}
& Acc. & 58.90 & 53.76333333 & 72.60666667 & 68.24333333 & 73.14 & 30.89666667 & 73.28666667 & 73.04 \\
& & ECE & 30.47333333 & 27.10333333 & 39.25333333 & 9.72666667 & 40.88 & 18.75 & 41.64333333 & 37.27 \\
& & MCE & 19.73666667 & 17.21 & 25.78666667 & 24.11 & 26.87333333 & 18.75 & 31.25 & 25.19666667 \\
& & ACE & 30.47333333 & 28.81 & 39.25333333 & 38.30 & 40.88 & 18.98333333 & 41.72 & 37.27 \\
\cline{2-11}

& \multirow{4}{*}{Messidor-2}
& Acc. & 51.38 & 46.63333333 & 57.51333333 & 57.96666667 & 57.66666667 & 30.65666667 & 57.97 & 57.70 \\
& & ECE & 22.44 & 15.69 & 24.59333333 & 23.73333333 & 23.63666667 & 14.57 & 26.46333333 & 21.20 \\
& & MCE & 16.11 & 10.26333333 & 14.33666667 & 15.18333333 & 14.61333333 & 14.57 & 17.30333333 & 14.43333333 \\
& & ACE & 22.44 & 16.60666667 & 24.55 & 23.73333333 & 23.49333333 & 15.37333333 & 26.92 & 21.20333333 \\
\cline{2-11}
& \multirow{4}{*}{Average}
& Acc. & 52.69 & 46.81 & 59.56 & 58.44 & 59.91 & 32.18 & 60.10 & 59.83 \\
& & ECE & 23.95 & 17.00 & 26.61 & 15.86 & 26.23 & 16.47 & 28.72 & 23.54 \\
& & MCE & 17.48 & 11.27 & 16.80 & 17.17 & 17.33 & 16.47 & 19.30 & 16.02 \\
& & ACE & 23.94 & 19.10 & 26.64 & 25.57 & 26.25 & 17.36 & 28.92 & 23.68 \\

\midrule
\multirow{8}{*}{MedCLIP-X-ray-DS}
& \multirow{4}{*}{COVIDX}
& Acc. & 79.15 & 79.15 & 79.07666667 & 79.06 & 79.02666667 & 78.49666667 & 79.16333333 & 79.14666667 \\
& & ECE & 29.03333333 & 29.03333333 & 28.96666667 & 28.94666667 & 28.91333333 & 28.38666667 & 29.06666667 & 28.03666667 \\
& & MCE & 29.03333333 & 29.03333333 & 28.96666667 & 28.94666667 & 28.91333333 & 28.38666667 & 29.06666667 & 28.03666667 \\
& & ACE & 29.03333333 & 29.03333333 & 28.96666667 & 28.94666667 & 28.91333333 & 28.38666667 & 29.06666667 & 29.03666667 \\
\cline{2-11}
& \multirow{4}{*}{Average}
& Acc. & 79.15 & 79.15 & 79.07666667 & 79.06 & 79.02666667 & 78.49666667 & 79.16333333 & 79.14666667 \\
& & ECE & 29.03333333 & 29.03333333 & 28.96666667 & 28.94666667 & 28.91333333 & 28.38666667 & 29.06666667 & 28.03666667 \\
& & MCE & 29.03333333 & 29.03333333 & 28.96666667 & 28.94666667 & 28.91333333 & 28.38666667 & 29.06666667 & 28.03666667 \\
& & ACE & 29.03333333 & 29.03333333 & 28.96666667 & 28.94666667 & 28.91333333 & 28.38666667 & 29.06666667 & 29.03666667 \\

\bottomrule

\end{tabular}%
}
\end{table*}

\begin{table*}[t!]
\centering
\resizebox{\textwidth}{!}{%
\begin{tabular}{@{}l|c|c|*{7}{d}D@{}}
\toprule
\textbf{Backbone} & \textbf{Dataset} & \textbf{Metric} &
\multicolumn{1}{c}{\rotatebox[origin=c]{45}{\parbox{15mm}{\centering \textbf{Base}}}} &
\multicolumn{1}{c}{\rotatebox[origin=c]{45}{\parbox{15mm}{\centering \textbf{MDCA}}}} &
\multicolumn{1}{c}{\rotatebox[origin=c]{45}{\parbox{15mm}{\centering \textbf{LS}}}} &
\multicolumn{1}{c}{\rotatebox[origin=c]{45}{\parbox{15mm}{\centering \textbf{MBLS}}}} &
\multicolumn{1}{c}{\rotatebox[origin=c]{45}{\parbox{15mm}{\centering \textbf{ECCV\_ ZS}}}} &
\multicolumn{1}{c}{\rotatebox[origin=c]{45}{\parbox{15mm}{\centering \textbf{ECCV\_ Penalty}}}} &
\multicolumn{1}{c}{\rotatebox[origin=c]{45}{\parbox{15mm}{\centering \textbf{TS}}}} &
\multicolumn{1}{>{\columncolor{pink!20}}c@{}}{\rotatebox[origin=c]{45}{\parbox{15mm}{\centering \textbf{Ours (MCM)}}}} \\

\midrule
\multirow{20}{*}{BioMedCLIP-DR-B/32-DS}
& \multirow{4}{*}{APTOS}
& Acc. & 50.20 & 51.76666667 & 50.73666667 & 50.58 & 50.65333333 & 52.80333333 & 51.02 & 50.12 \\
& & ECE & 23.70333333 & 26.15333333 & 20.65333333 & 21.85 & 30.02 & 15.19 & 25.28333333 & 22.19 \\
& & MCE & 7.02 & 6.58666667 & 7.89666667 & 6.91 & 10.80 & 6.45 & 6.78333333 & 6.563333333 \\
& & ACE & 24.07333333 & 26.86333333 & 20.74666667 & 21.92 & 30.26333333 & 15.75666667 & 26.01 & 22.34666667 \\
\cline{2-11}
& \multirow{4}{*}{EyePACS}
& Acc. & 68.81666667 & 69.49666667 & 68.33333333 & 68.24333333 & 69.08333333 & 70.98333333 & 69.19 & 68.58 \\
& & ECE & 8.95666667 & 9.34 & 12.53333333 & 9.72666667 & 15.68333333 & 13.41333333 & 9.08 & 6.55 \\
& & MCE & 2.25666667 & 2.36666667 & 3.13 & 2.38333333 & 4.25333333 & 3.75666667 & 4.17333333 & 1.673333333 \\
& & ACE & 8.96333333 & 9.32666667 & 12.52666667 & 9.72333333 & 15.42666667 & 13.41333333 & 7.41 & 6.543333333 \\
\cline{2-11}

& \multirow{4}{*}{Messidor-2}
& Acc. & 59.34666667 & 59.46 & 58.64 & 58.73333333 & 57.60666667 & 59.50 & 59.55666667 & 59.07 \\
& & ECE & 12.52 & 11.94666667 & 14.44333333 & 13.01666667 & 17.37666667 & 11.50666667 & 12.73666667 & 10.08 \\
& & MCE & 3.79333333 & 4.11 & 4.83333333 & 3.91666667 & 7.11666667 & 4.81666667 & 2.54333333 & 3.073333333 \\
& & ACE & 12.48666667 & 11.64333333 & 14.50333333 & 12.95 & 16.59 & 11.50666667 & 11.98 & 10.09 \\
\cline{2-11}
& \multirow{4}{*}{Average}
& Acc. & 59.45 & 60.24 & 59.24 & 59.19 & 59.11 & 61.10 & 59.92 & 59.26 \\
& & ECE & 15.06 & 15.81 & 15.88 & 14.86 & 21.03 & 13.37 & 15.70 & 12.94 \\
& & MCE & 4.36 & 4.35 & 5.29 & 4.40 & 7.39 & 5.01 & 4.50 & 3.77 \\
& & ACE & 15.17 & 15.94 & 15.93 & 14.86 & 20.76 & 13.56 & 15.13 & 12.99 \\

\midrule
\multirow{8}{*}{BioMedCLIP-X-ray-DS}
& \multirow{4}{*}{COVIDX}
& Acc. & 74.55666667 & 73.39666667 & 75.55333333 & 77.44666667 & 73.63 & 77.15 & 74.09666667 & 78.84333333 \\
& & ECE & 8.996666667 & 9.793333333 & 6.823333333 & 7.313333333 & 10.38 & 3.833333333 & 8.696666667 & 8.023333333 \\
& & MCE & 4.41 & 4.36 & 3.61 & 4.71 & 4.91 & 1.61 & 4.2 & 5.74 \\
& & ACE & 8.96 & 9.91 & 6.83 & 7.47 & 10.27 & 3.99 & 8.87 & 8.083333333 \\
\cline{2-11}
& \multirow{4}{*}{Average}
& Acc. & 74.55666667 & 73.39666667 & 75.55333333 & 77.44666667 & 73.63 & 77.15 & 74.09666667 & 78.84333333 \\
& & ECE & 8.996666667 & 9.793333333 & 6.823333333 & 7.313333333 & 10.38 & 3.833333333 & 8.696666667 & 8.023333333 \\
& & MCE & 4.41 & 4.36 & 3.61 & 4.71 & 4.91 & 1.61 & 4.2 & 5.74 \\
& & ACE & 8.96 & 9.91 & 6.83 & 7.47 & 10.27 & 3.99 & 8.87 & 8.083333333 \\

\midrule
\multirow{16}{*}{PLIP-B/32-DS}
& \multirow{4}{*}{APTOS}
& Acc. & 39.01 & 41.09 & 41.26 & 28.77333333 & 39.45 & 37.14666667 & 39.00666667 & 35.19 \\
& & ECE & 17.74666667 & 17.01666667 & 13.11666667 & 23.03666667 & 14.10666667 & 9.86333333 & 16.15666667 & 21.78 \\
& & MCE & 1.25666667 & 1.23666667 & 2.42333333 & 1.30333333 & 0.76 & 4.23666667 & 1.25666667 & 4.923333333 \\
& & ACE & 17.87 & 17.18 & 12.18666667 & 22.91 & 14.21 & 9.95 & 15.64 & 21.96333333 \\
\cline{2-11}
& \multirow{4}{*}{EyePACS}
& Acc. & 70.34 & 70.73333333 & 66.95 & 69.95666667 & 69.88333333 & 71.12333333 & 69.58666667 & 61.84 \\
& & ECE & 10.96 & 9.69333333 & 16.85666667 & 13.50 & 8.88 & 25.38333333 & 13.09333333 & 10.69 \\
& & MCE & 0.49333333 & 0.38666667 & 3.37666667 & 0.52 & 0.32666667 & 7.68666667 & 0.49333333 & 3.993333333 \\
& & ACE & 10.38333333 & 8.14666667 & 17.35 & 12.42666667 & 8.31666667 & 24.34666667 & 12.38333333 & 10.5 \\
\cline{2-11}
& \multirow{4}{*}{Messidor-2}
& Acc. & 56.84333333 & 53.47666667 & 52.31 & 53.72666667 & 53.48 & 46.88666667 & 54.95 & 44.45 \\
& & ECE & 9.65 & 9.97333333 & 12.94 & 13.07 & 9.66333333 & 21.18666667 & 11.10333333 & 8.20 \\
& & MCE & 1.01 & 1.35333333 & 2.14 & 1.21666667 & 0.93 & 6.06666667 & 1.01 & 3.573333333 \\
& & ACE & 9.91 & 9.94666667 & 11.03333333 & 12.78 & 8.56 & 20.00666667 & 10.91 & 7.933333333 \\
\cline{2-11}
& \multirow{4}{*}{Average}
& Acc. & 55.39777778 & 55.10 & 53.50666667 & 50.81888889 & 54.27111111 & 51.71888889 & 54.51444444 & 47.16 \\
& & ECE & 12.78555556 & 12.22777778 & 14.30444445 & 16.53555556 & 10.88333333 & 18.81111111 & 13.45111111 & 13.55666667 \\
& & MCE & 0.92000000 & 0.99222222 & 2.64666667 & 1.01333333 & 0.67222222 & 5.99666667 & 0.92000000 & 4.163333333 \\
& & ACE & 12.72 & 11.76 & 13.52 & 16.04 & 10.36 & 18.10 & 12.97 & 13.46555555 \\

\bottomrule

\end{tabular}%
}
\end{table*}

\begin{table*}[t!]
\centering
\resizebox{\textwidth}{!}{%
\begin{tabular}{@{}l|c|c|*{7}{d}D@{}}
\toprule
\textbf{Backbone} & \textbf{Dataset} & \textbf{Metric} &
\multicolumn{1}{c}{\rotatebox[origin=c]{45}{\parbox{15mm}{\centering \textbf{Base}}}} &
\multicolumn{1}{c}{\rotatebox[origin=c]{45}{\parbox{15mm}{\centering \textbf{MDCA}}}} &
\multicolumn{1}{c}{\rotatebox[origin=c]{45}{\parbox{15mm}{\centering \textbf{LS}}}} &
\multicolumn{1}{c}{\rotatebox[origin=c]{45}{\parbox{15mm}{\centering \textbf{MBLS}}}} &
\multicolumn{1}{c}{\rotatebox[origin=c]{45}{\parbox{15mm}{\centering \textbf{ECCV\_ ZS}}}} &
\multicolumn{1}{c}{\rotatebox[origin=c]{45}{\parbox{15mm}{\centering \textbf{ECCV\_ Penalty}}}} &
\multicolumn{1}{c}{\rotatebox[origin=c]{45}{\parbox{15mm}{\centering \textbf{TS}}}} &
\multicolumn{1}{>{\columncolor{pink!20}}c@{}}{\rotatebox[origin=c]{45}{\parbox{15mm}{\centering \textbf{Ours (MCM)}}}} \\

\midrule
\multirow{12}{*}{PLIP-Histo.-DS}
& \multirow{4}{*}{DigestPath}
& Acc. & 65.14666667 & 46.46 & 61.52333333 & 50.25333333 & 74.22 & 39.36 & 64.95333333 & 56.56333333 \\
& & ECE & 24.36333333 & 39.39333333 & 15.87333333 & 35.58666667 & 16.71666667 & 33.58 & 24.06333333 & 33.76666667 \\
& & MCE & 1.25666667 & 1.23666667 & 2.42333333 & 1.30333333 & 0.76 & 4.23666667 & 1.25666667 & 2.58 \\
& & ACE & 24.9833333 & 35.23666667 & 12.18666667 & 35.05333333 & 15.52333333 & 32.99666667 & 24.98333333 & 33.76666667 \\
\cline{2-11}
& \multirow{4}{*}{Kather}
& Acc. & 40.51333333 & 26.23666667 & 27.38 & 26.88666667 & 56.48666667 & 21.15 & 51.50 & 51.20 \\
& & ECE & 12.79 & 13.11333333 & 5.08666667 & 12.53333333 & 9.77 & 9.37 & 7.23 & 11.09666667 \\
& & MCE & 0.49333333 & 0.38666667 & 3.37666667 & 0.52 & 0.32666667 & 7.68666667 & 0.49333333 & 1.946666667 \\
& & ACE & 12.38333333 & 12.14666667 & 7.35 & 12.42666667 & 9.31666667 & 10.34666667 & 8.38333333 & 11.07333333 \\
\cline{2-11}
& \multirow{4}{*}{Average}
& Acc. & 52.83 & 36.34833334 & 44.45166667 & 38.57 & 65.35333334 & 30.255 & 58.22666667 & 53.88166667 \\
& & ECE & 18.57666667 & 26.25333333 & 10.48 & 24.06 & 13.24333334 & 21.475 & 15.64666667 & 22.43166667 \\
& & MCE & 0.87500000 & 0.81166667 & 2.90000000 & 0.91166667 & 0.54333333 & 5.96166667 & 0.87500000 & 2.26 \\
& & ACE & 18.68 & 23.69 & 9.77 & 23.74 & 12.42 & 21.67 & 16.68 & 22.42 \\

\midrule
\multirow{16}{*}{QuiltNet-DR-B/32-DS}
& \multirow{4}{*}{APTOS}
& Acc. & 32.36 & 45.50333333 & 42.62 & 39.79666667 & 47.17666667 & 48.21666667 & 42.51 & 27.86333333 \\
& & ECE & 19.28333333 & 10.25 & 9.73 & 12.67 & 10.22666667 & 11.88666667 & 8.67 & 22.03333333 \\
& & MCE & 1.25666667 & 1.23666667 & 2.42333333 & 1.30333333 & 0.76 & 4.23666667 & 1.25666667 & 2.42 \\
& & ACE & 14.98333333 & 8.23666667 & 12.18666667 & 11.05333333 & 9.52333333 & 14.99666667 & 6.98333333 & 22.13 \\
\cline{2-11}
& \multirow{4}{*}{EyePACS}
& Acc. & 55.53333333 & 56.39 & 61.05 & 66.04 & 62.39666667 & 58.74666667 & 61.90333333 & 34.33333333 \\
& & ECE & 8.80 & 11.51 & 9.27333333 & 11.77666667 & 8.74333333 & 18.87 & 10.80666667 & 23.30 \\
& & MCE & 0.49333333 & 0.38666667 & 3.37666667 & 0.52 & 0.32666667 & 7.68666667 & 0.49333333 & 3.996666667 \\
& & ACE & 7.38333333 & 10.14666667 & 9.35 & 9.42666667 & 8.31666667 & 19.34666667 & 11.38333333 & 23.28 \\
\cline{2-11}
& \multirow{4}{*}{Messidor-2}
& Acc. & 32.35666667 & 40.25 & 45.95 & 41.93333333 & 41.93666667 & 42.79333333 & 38.64666667 & 21.56 \\
& & ECE & 18.18666667 & 7.85333333 & 11.12 & 8.99333333 & 7.09333333 & 19.38333333 & 10.14666667 & 25.81666667 \\
& & MCE & 1.01 & 1.35333333 & 2.14 & 1.21666667 & 0.93 & 6.06666667 & 1.01 & 2.01 \\
& & ACE & 17.91 & 6.94666667 & 10.03333333 & 8.78 & 6.66 & 18.00666667 & 9.91 & 25.77 \\
\cline{2-11}
& \multirow{4}{*}{Average}
& Acc. & 40.08333333 & 47.38111111 & 49.87333333 & 49.25666667 & 50.50333334 & 49.91888889 & 47.68666667 & 27.91888889 \\
& & ECE & 15.42333333 & 9.87111111 & 10.04111111 & 11.14666667 & 8.68777778 & 16.71333333 & 9.87444444 & 23.71666667 \\
& & MCE & 0.92000000 & 0.99222222 & 2.64666667 & 1.01333333 & 0.67222222 & 5.99666667 & 0.92000000 & 2.81 \\
& & ACE & 13.42 & 8.44333333 & 10.52 & 9.75333333 & 8.16 & 17.45 & 9.42 & 23.72666667 \\

\midrule
\multirow{12}{*}{QuiltNet-Histo.-DS}
& \multirow{4}{*}{DigestPath}
& Acc. & 47.39666667 & 39.66333333 & 42.96 & 36.96666667 & 43.63 & 31.01666667 & 31.42666667 & 49.01333333 \\
& & ECE & 41.30 & 46.37666667 & 42.33333333 & 51.88 & 44.47666667 & 46.04 & 60.38666667 & 42.92666667 \\
& & MCE & 1.25666667 & 1.23666667 & 2.42333333 & 1.30333333 & 0.76 & 4.23666667 & 1.25666667 & 1.11 \\
& & ACE & 41.98333333 & 45.23666667 & 42.18666667 & 50.05333333 & 43.52333333 & 42.99666667 & 58.98333333 & 42.92666667 \\
\cline{2-11}
& \multirow{4}{*}{Kather}
& Acc. & 45.52666667 & 50.58666667 & 37.08333333 & 47.95 & 46.23333333 & 28.12666667 & 47.77333333 & 44.51 \\
& & ECE & 15.01666667 & 9.54333333 & 15.58666667 & 10.37666667 & 13.40666667 & 7.46333333 & 12.93666667 & 24.27 \\
& & MCE & 0.49333333 & 0.38666667 & 3.37666667 & 0.52 & 0.32666667 & 7.68666667 & 0.49333333 & 5.43 \\
& & ACE & 15.38333333 & 10.14666667 & 16.35 & 10.42666667 & 11.31666667 & 10.34666667 & 10.38333333 & 24.18666667 \\
\cline{2-11}
& \multirow{4}{*}{Average}
& Acc. & 46.46166667 & 45.125 & 40.02166667 & 42.45833334 & 44.93166667 & 29.57166667 & 39.60000000 & 46.76166667 \\
& & ECE & 28.15 & 27.96 & 28.96 & 31.12833334 & 28.94166667 & 26.75166667 & 36.66166667 & 33.59833334 \\
& & MCE & 0.87500000 & 0.81166667 & 2.90000000 & 0.91166667 & 0.54333333 & 5.96166667 & 0.87500000 & 3.27 \\
& & ACE & 28.68 & 27.69 & 29.27 & 30.24 & 27.42000000 & 26.67 & 34.68 & 33.55666667 \\

\bottomrule

\end{tabular}%
}
\caption{Accuracy and ECE of each calibration method in domain-shift (out-domain) across 12 modality-specific configurations and 7 calibration methods covering three modalities: Diabetic retinopathy grading, histopathology, and chest-X-ray.}
\label{tab:dsallmet}
\end{table*}
\section{Declaration of LLM usage}\label{sec:decllm}
During the preparation of this manuscript, ChatGPT was used only for language polishing, grammar checking, and improving sentence clarity and readability. The tool was not used to generate experimental results, design the methodology, conduct analysis, create citations, or formulate the scientific claims of the paper. All technical content, interpretations, figures, tables, and references were reviewed, verified, and finalized by the authors.

\clearpage


\end{document}